\documentclass{article}
\usepackage{tikz}
\usepackage{natbib}
\usepackage[utf8]{inputenc}  
\usepackage{fontenc}
\usepackage{hyperref}
\usepackage{amsmath}
\usepackage{amsfonts}
\usepackage{algorithmic}
\usepackage{lscape}
\usepackage{float}
\usepackage{a4wide}
\usepackage{dsfont}
\usepackage{bm}
\usepackage[]{algorithm2e}

\usetikzlibrary{shapes,arrows}

\title{Parallel Predictive Entropy Search for Multi-objective Bayesian Optimization with Constraints}

\author{
Eduardo C. Garrido-Merchán \\
Universidad Aut\'onoma de Madrid\\
\texttt{eduardo.garrido@uam.es} \\
\and 
Daniel Hern\'andez-Lobato\\
Universidad Aut\'onoma de Madrid\\
\texttt{daniel.hernandez@uam.es} 
}

\date{}

\begin{document}
	
\maketitle

\begin{abstract}
Real-world problems often involve the optimization of several
objectives under multiple constraints. An example is
the hyper-parameter tuning problem of machine learning algorithms. 
In particular, the minimization of the estimation of the generalization error of a 
deep neural network and at the same time the minimization of its prediction time. 
We may also consider as a constraint that the deep neural network 
must be implemented in a chip with an area below some size. Here, both 
the objectives and the constraint are black boxes, \textit{i.e}., functions 
whose analytical expressions are unknown and are expensive to evaluate. 
Bayesian optimization (BO) methodologies have given state-of-the-art results for 
the optimization of black-boxes. Nevertheless, most BO methods are sequential and
evaluate the objectives and the constraints at just one input location, iteratively.
Sometimes, however, we may have resources to evaluate several configurations in parallel.
Notwithstanding, no parallel BO method has been proposed to deal with the optimization 
of multiple objectives under several constraints. If the expensive evaluations can 
be carried out in parallel (as when a cluster of computers is available), sequential 
evaluations result in a waste of resources. This article 
introduces PPESMOC, Parallel Predictive Entropy Search for Multi-objective Bayesian 
Optimization with Constraints, an information-based batch method for 
the simultaneous optimization of multiple expensive-to-evaluate black-box functions 
under the presence of several constraints. Iteratively, PPESMOC selects a batch of 
input locations at which to evaluate the black-boxes so as to maximally reduce 
the entropy of the Pareto set of the optimization problem. To our 
knowledge, this is the first batch method for constrained multi-objective BO. We 
present empirical evidence in the form of synthetic, benchmark and 
real-world experiments that illustrate the effectiveness of PPESMOC.
\end{abstract}
	
\section{Introduction}

Black-boxes are defined as functions whose analytical expression is unknown. Hence, 
we can not compute their gradients. Moreover, 
these are expensive to evaluate functions, either in terms of computational time 
or other resources. Additionally, the evaluations of these functions
are noisy. That is, the evaluation of these functions is contaminated by noise. 
More precisely, an example of a black-box is the hyper-parameter tuning of
a machine learning algorithm \citep{feurer2019hyperparameter}. In this problem, the 
estimation of the generalization error carried out by a machine learning algorithm 
over a dataset is minimized. Furthermore, many real-world scenarios involve the 
simultaneous optimization of a set of objectives subject to several constraints being 
simultaneously validated. An example of such an scenario is tuning the control system of 
a four-legged robot. We may be interested in finding the optimal control parameters to 
minimize the robot’s energy consumption and maximize locomotion speed \citep{ariizumi2014expensive}, under 
the constraint that the amount of weight placed on a leg of the robot does not exceed 
a specific value, or similarly, that the maximum angle between the legs 
of the robot is below some other value for safety reasons. Measuring the objectives and the 
constraints in this case may involve an expensive computer simulation or doing some 
actual experiment with the robot. There is no analytical expression to describe the output 
of that process, which can take a significant amount of time. 
Moreover, the result of the experiment can be different each time, depending, \emph{e.g.}, 
on the environmental conditions.

Bayesian optimization (BO) has been empirically shown to be a good class of methods 
to deal with the optimization problems described in the previous paragraph 
\citep{mockus1978application}. In particular, several constrained multi-objective 
BO methods have obtained good results in that setting \citep{feliot2017bayesian,garrido2019predictive}.
These methods iteratively suggest, in an intelligent way, a point at which to evaluate the
objectives and the constraints to solve the optimization problem in a small number of steps.
This process is repeated until enough points have been evaluated. Then, a recommendation is made 
as the potential solution of the problem. However, a limitation of BO methods for constrained multi-objective optimization 
is that they choose a single location at a time at which to evaluate the black-boxes 
\citep{garrido2019predictive}. Assume that a cluster of computers or some other resource is 
available to perform the evaluation of the black-boxes at several points in 
parallel. If only a single point is evaluated each time, this results in a
a waste of resources and leads to sub-optimal optimization results. The problem described can be
solved by using BO methods that suggest not only a single point at which to evaluate the 
black-boxes, but a batch or collection of points of adjustable size 
\citep{azimi2012hybrid,bergstra2011algorithms,gonzalez2016batch,shah2015parallel}.

To the best of our knowledge, no BO method has been proposed to deal with the optimization of 
multiple objectives under several constraints, when the evaluations can be done in parallel. 
Only sequential methods have been introduced to tackle 
the constrained multi-objective scenario \citep{feliot2017bayesian,garrido2019predictive}. Therefore, 
the literature about BO is missing important methods to address BO problems with the characteristics 
described. In this work, we propose a BO method called PPESMOC, Parallel Predictive Entropy Search for 
Multi-objective Bayesian Optimization with Constraints, that can precisely address the problems described.
Specifically, such a method can suggest, at each iteration, a batch of points at which to evaluate all the 
black-boxes in parallel. The method proposed is based on an extension of the PESMOC (Predictive Entropy 
Search for Multi-objective Bayesian Optimization with Constraints) method \citep{garrido2019predictive} 
and an extension of parallel single-objective and un-constrained BO \citep{shah2015parallel}. 

A batch or parallel BO method usually consists of a probabilistic surrogate model that provides a predictive distribution of 
the black-box function and an acquisition function whose maximum indicates the next batch of points to be 
evaluated. In particular, the acquisition function of the BO method that we consider in 
this work receives as an input a batch of candidate input locations at which to perform the evaluation
of the black-boxes in parallel and estimates the expected reduction in the entropy of the solution of 
the optimization problem. The difference with PESMOC is that here, the acquisition function evaluates a 
batch of points instead of a single point. However, as we will see in the following sections, considering the 
suggestion of a batch of points is challenging and leads to several problems that need to be tackled by a novel methodology, 
which is precisely what the PPESMOC acquisition function represents.

Summing up, in this research work we present the first method for multi-objective BO with constraints that allows
for parallel evaluations. We believe that this is an important contribution for the BO community. Furthermore,
we provide an exhaustive and detailed empirical evaluation of such a method in several optimization problems.
We also compare results with some simple base-lines to address such a problem, showing that our approach
sometimes leads to better results. Importantly, however, PPESMOC scales better with the batch size than these
simple methods. This allows to consider bigger batch sizes.

The rest of the paper is organized as follows: Section \ref{section2:PBO} describes in detail 
the fundamentals of BO and, in particular, constrained multi-objective parallel or batch BO. 
Then, Section \ref{section3:PPESMOC} describes the proposed approach, PPESMOC, and how to 
compute an approximate acquisition function that suggests a batch of points based on the expected 
reduction of the entropy of the Pareto set. After that, Section \ref{section4:rw} includes related 
work that describes batch BO methods and BO methods that can address multi-objective problems with 
constraints. Then, Section \ref{section5:exps} contains extensive experiments to evaluate the performance of 
PPESMOC in synthetic, benchmark and real-world optimization problems. We have compared results 
with a base-line which chooses the points to evaluate at random and with a simple method that 
applies iteratively  a sequential BO method for constrained multi-objective BO (as many times as 
the batch size). This last method introduces virtual observations (fantasies) to avoid choosing many times 
the same point to be evaluated.  The results show that PPESMOC performs better than a random search strategy 
and similarly or better than sequential base-lines. The advantage of PPESMOC is, 
however, that its cost scales much better with respect to the batch size than the sequential base-lines. 
Finally, Section \ref{seq-conclusions} enumerates the conclusions of this article.

\section{Constrained Multi-Objective Parallel Bayesian Optimization}
\label{section2:PBO}

After introducing the parallel constrained multi-objective BO scenario, this section formalizes 
the described problem and the BO algorithm adapted to this scenario. Recall that in 
Section \ref{section3:PPESMOC} we will describe our proposed approach, PPESMOC, so here we will 
only focus on general parallel constrained multi-objective BO concepts that need to 
be understood before. 

As it was described in the previous section, the purpose of BO is to retrieve the 
extremum $\mathbf{x}^\star$ of a black-box function 
$f(\mathbf{x})$ where $\mathbf{x} \in \mathcal{X}$ and $\mathcal{X}$ is the input 
space where $f(\mathbf{x})$ can be evaluated. Formally, we seek to obtain $\mathbf{x}^\star$ such that,
\begin{equation}
\mathbf{x}^\star = \arg\min_{\mathbf{x} \in \mathcal{X}}f(\mathbf{x})\,,
\end{equation}
assuming minimization. However, in this work, we consider the problem of simultaneously minimizing
$K$ functions $f_1(\mathbf{x}),...,f_K(\mathbf{x})$ which we define as objectives,
subject to the non-negativity of $C$ constraints $c_1(\textbf{x}),....,c_C(\textbf{x}) $,
over some bounded domain $\mathcal{X} \in \mathds{R}^d$, where $d$ is the dimensionality
of the input space. The problem considered is:
\begin{align}
\underset{\mathbf{x} \in \mathcal{X}}{\text{min}} & \quad f_1(\mathbf{x}), \ldots, f_K(\mathbf{x}) &
\text{s.t.} \quad \quad c_1(\mathbf{x}) \geq 0, \ldots, c_C(\mathbf{x}) \geq 0\,.
\label{eq:optimization}
\end{align}
We say that a point $\mathbf{x} \in \mathcal{X}$ is feasible if $c_j(\mathbf{x})\geq 0$, $\forall j$,
that is, it satisfies all the constraints. This leads to the concept of feasible space $\mathcal{F}
\subset \mathcal{X}$, that is the set of points that are feasible. In this scenario, only the solutions
contained in $\mathcal{F}$ are considered valid.

Focusing on the multi-objective optimization part of the problem, most of the times it is impossible
to optimize all the objective functions at the same time, as they may
be conflicting. For example, in the control system of the robot described before, most probably
maximizing locomotion speed will lead to an increase in the energy consumption. 
In spite of this, it is still possible to find a set of optimal points $\mathcal{X}^{\star}$  known as the \textit{Pareto set}
\citep{siarry2003multiobjective}. More formally, we define that the point $\mathbf{x}$
dominates the point $\mathbf{x}'$ if $f_k (\mathbf{x})\leq f_k (\mathbf{x}')$ $\forall k$,
with at least one inequality being strict. Then, the Pareto set is the subset of
non-dominated points in $\mathcal{F}$, the feasible space, which is equivalent to this
expression $\forall \mathbf{x}^{\star} \in \mathcal{X}^{\star} \subset \mathcal{F} \,,
\forall \mathbf{x} \in \mathcal{F}\, \exists\, k \in {1,...,K}$ such that $f_k(\mathbf{x}^{\star}) < f_k(\mathbf{x})$.
The Pareto set is considered to be optimal because for each point in that set
one cannot improve in one of the objectives without deteriorating some
other objective. Given $\mathcal{X}^\star$, a final user may then choose a point
from this set according to their preferences, $\emph{e.g.}$, locomotion speed vs. energy consumption.

To solve efficiently the previous problems, \emph{i.e.}, find the Pareto set in $\mathcal{F}$ with
a small number of evaluations, BO methods perform a sequence of steps iteratively. The number of 
iterations is delimited by the budget that an user can afford to evaluate the black-boxes, which is assumed 
to be costly. For example, the user may only be able to test $50$ configurations of a machine 
learning algorithm. In this case, the number of BO iterations would be $50$. At each iteration 
$t$, a point $\mathbf{x}_t$ is suggested. The point $\mathbf{x}_t$ is suggested by following 
the next steps: first, BO methods fit a probabilistic model, typically, a Gaussian process (GP) 
to the observed data of each black-box function (objective or constraint). The uncertainty about the 
potential values of these functions given by the predictive distribution of the GPs is then used to build 
an acquisition function $\alpha(\mathbf{x})$ whose maximum determines the next point to be evaluated. 
An acquisition function $\alpha(\mathbf{x})$ represents an 
exploration-exploitation trade-off. Specifically, the acquisition function $\alpha(\mathbf{x})$ favors exploration 
in the sense that it is high in areas where no point $\mathbf{x}$ has been evaluated before. These 
areas may contain values near the optimum value. It also needs to perform exploitation, \textit{i.e.}, 
the acquisition function $\alpha(\mathbf{x})$ favors the evaluation of points that are likely to 
lead to good evaluations. Concretely, we expect that the new evaluations are near the current best observed value, as 
we assume that the function is smooth. Importantly, a constrained multi-objective acquisition function takes 
into account the predictive distribution of every black-box \citep{garrido2019predictive}.

Considering the sequential scenario, the maximum of the acquisition function 
$\alpha(\mathbf{x})$ indicates the most promising location at
which to evaluate next the objectives $f_1(\mathbf{x}),...,f_K(\mathbf{x})$ and 
the constraints $c_1(\textbf{x}),....,c_C(\textbf{x})$ to solve the
optimization problem at each iteration $t$. Namely,
\begin{equation}
\mathbf{x}_t  = \arg\max_{\mathbf{x} \in \mathcal{X}}\alpha_t(\mathbf{x})\,,
\end{equation}
The process of evaluating the black-boxes, updating the probabilistic models, 
and generating the acquisition function is repeated iteratively.
After enough observations have been collected like this,
the probabilistic models can be optimized
to provide an estimate of the Pareto set of the original
problem. Importantly, the acquisition function $\alpha(\mathbf{x})$ only depends
on the uncertainty provided by the probabilistic models and
not on the actual objectives or constraints. This means that
it can be evaluated and optimized very quickly to identify
the next evaluation point. 

\begin{figure}[htbp]
\centering{
        \begin{tabular}{rl}
                \includegraphics[width=\textwidth]{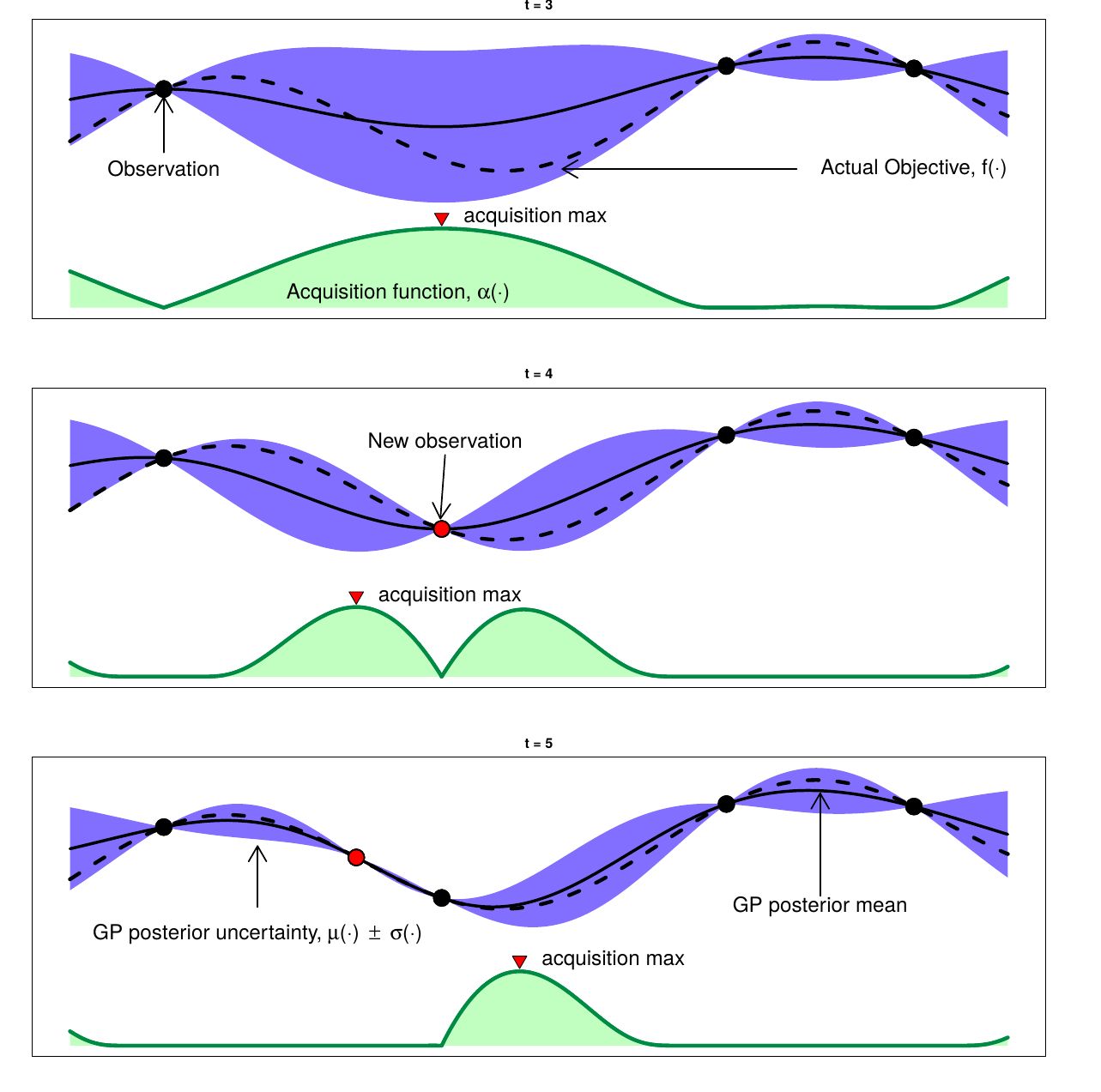} \\
        \end{tabular}
}
\caption{BO acquisition function $\alpha(\mathbf{x})$ and GP predictive distribution $p(f(\mathbf{x})|\mathcal{D})$ on a toy 1D noiseless problem. The figures show a GP estimation of the objective $f(\mathbf{x})$ over three iterations. The acquisition function $\alpha(\mathbf{x})$ is shown in the lower part of the plot. The acquisition $\alpha(\mathbf{x})$ is high where the GP predicts a low value of the objective $f(\mathbf{x})$ and where the uncertainty about its prediction is high. Those regions in which it is unlikely to find the global minimum $\mathbf{x}^\star$ of $f(\mathbf{x})$ have low acquisition values, and will not be explored.}
\label{fig:bo_process}
\end{figure}

By carefully choosing the points at
which to evaluate the objectives and the constraints, BO
methods find a good estimate of the solution of the
optimization problem in a small number of
evaluations \citep{brochu2010tutorial,shahriari2015taking}. We can see in 
Figure \ref{fig:bo_process} an unconstrained single-objective example of the steps performed 
by a BO method. In this figure, the acquisition function $\alpha(\mathbf{x})$ is plotted on 
green, the GP posterior distribution on blue, observations are represented as black 
dots and the latest observation is represented as a red dot. The ground truth is 
plotted as a dashed black line, the GP posterior mean as a continuous black line 
and the GP posterior uncertainty of its prediction  as a blue area, surrounding the mean. 
The GP posterior uncertainty represents one standard deviation of the predictive 
distribution around the mean. We can see how the GP predictive distribution 
$p(f(\mathbf{x})|\mathcal{D})$ and the acquisition function $\alpha(\mathbf{x})$ shapes change 
at each iteration $t$.

\begin{figure}[ht]
\begin{center}
\includegraphics[width=\textwidth]{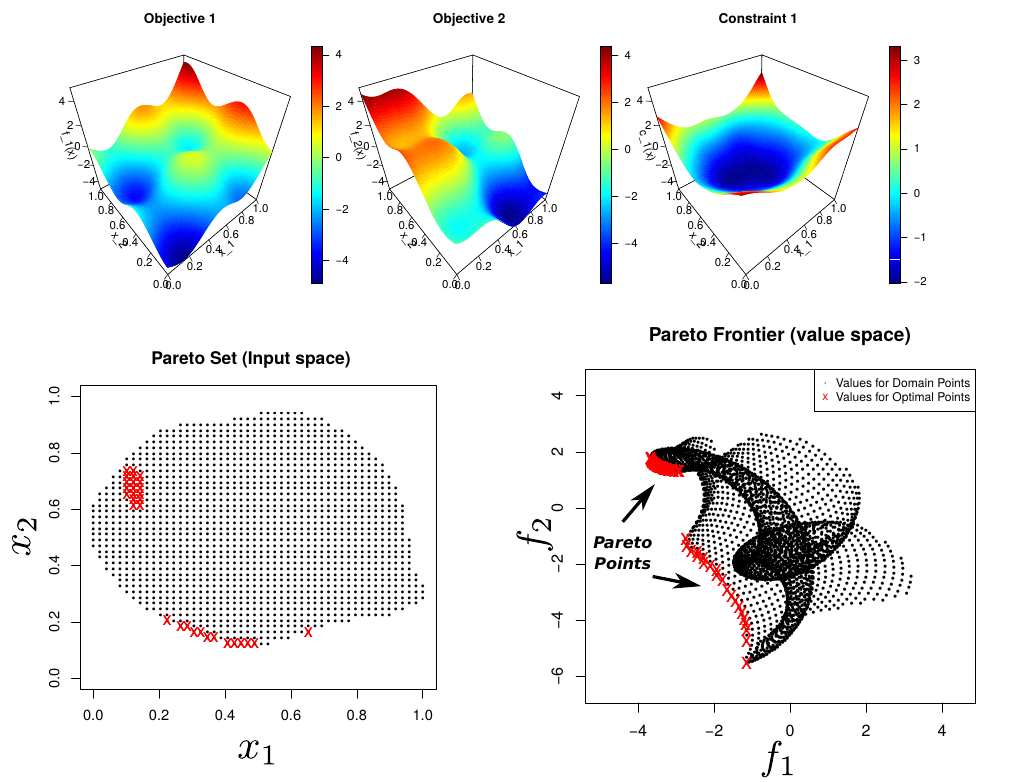}
\caption{Constrained multi-objective optimization problem with two objectives
and one constraint. (Top) Shapes of the objectives and constraints of the constrained
multi-objective optimization problem. (Bottom, left) Feasible space represented by black
dots and feasible Pareto set represented by red dots. (Bottom, right) Associated values
of the feasible space represented by black dots and Pareto frontier represented by red
dots.}
\label{fig:cmobo}
\end{center}
\end{figure}

The process described is repeated iteratively until the budget of evaluations $T$ is consumed. 
Finally, BO recommends the point $\mathbf{x}$ whose observation value $y$ has been the best 
as the solution of the problem. In particular, the goodness of a solution of a constrained 
multi-objective problem can be measured in terms of the hyper-volume metric \citep{hernandez2016predictive}. 
Concretely, assuming minimization, the hyper-volume is simply the volume above the 
feasible Pareto frontier $\mathcal{Y}$ (\emph{i.e.}, the function values associated to the feasible 
Pareto set $\mathcal{F}$), which is maximized by the actual Pareto set. We can see in 
Figure \ref{fig:cmobo} a visual representation of the constrained multi-objective scenario. 
The figures at the top represent the two objectives and the constraint involved in the 
optimization problem. The figures below represent the feasible 
Pareto set $\mathcal{F}$ and its associated Pareto frontier $\mathcal{Y}$. The hyper-volume 
will be the region contained above the Pareto frontier with respect to a reference point, 
which is usually set as the worst result for each objective. 
To sum up, we enumerate the described steps in Algorithm \ref{alg:bo}.

\begin{figure}[tb]
\begin{algorithm}[H]
\label{alg:bo}
\textbf{Input:} Maximum number of evaluations $T$.
\caption{BO of a constrained multi-objective problem with objectives $f_1(\mathbf{x}),...,f_K(\mathbf{x})$ and constraints $c_1(\textbf{x}),....,c_C(\textbf{x})$.}
\For{$\text{t}=1,2,3,\ldots,T$}{
        {\bf 1:} {\bf if $N=1$:}\\
        \hspace{.7cm}Choose $\mathbf{x}_t$ at random from $\mathcal{X}$. \\
        \hspace{.5cm}{\bf else:} \\
        \hspace{.7cm}Find $\mathbf{x}_t$ by maximizing the acquisition function:
        $\mathbf{x}_t = \underset{\mathbf{x} \in \mathcal{X}}{\text{arg max}} \quad \alpha_t(\mathbf{x})$.

        {\bf 2:} Evaluate the black-boxes $f_1(\mathbf{x}),...,f_K(\mathbf{x})$ and $c_1(\textbf{x}),....,c_C(\textbf{x})$ at $\mathbf{x}_t$: $y_t=f_k(\mathbf{x}_\text{t}) + \epsilon_t$.

        {\bf 3:} Augment the dataset of each black-box with the new observation: $\mathcal{D}_{1:t}=\mathcal{D}_{1:t-1} \bigcup \{\mathbf{x}_t, y_t\}$.

        {\bf 4:} Fit again the GP models using the augmented dataset $\mathcal{D}_{1:t}$.
 }
{\bf 5:} Obtain the recommendation $\mathbf{x}^\star$: Point associated with the value that optimizes the hyper-volume of the Pareto frontier or with the best observed value. \\
\KwResult{Recommended point $\mathbf{x}^\star$}
\vspace{.5cm}
\end{algorithm}
\end{figure}
The following sections will explain in detail the GP models used to describe each
black-box and the fundamentals of parallel BO, where not only one but a batch of points are 
suggested at each iteration $t$.

\subsection{Modeling the Black-boxes Using Gaussian Processes}

We model each objective $f_k(\cdot)$ and constraint $c_j(\cdot)$ using a Gaussian process (GP)
\citep{rasmussen2003gaussian}. A Gaussian Process (GP) is a collection of random variables 
(of potentially infinite size), any finite number of which have (consistent) joint Gaussian 
distributions. Equivalently, it describes a stochastic process whose values, at any finite 
number of input locations, have joint Gaussian distributions. We can also think of GPs 
as defining a distribution over functions where inference takes place directly in the space 
of functions \citep{rasmussen2003gaussian}. A GP can also be defined as a distribution over 
functions, \textit{i.e.}, we can sample functions from it. So, if we assume that each 
of the black-boxes $f(\cdot)$ is a sample of the GP, it is reasonable to use a GP to 
model the black-box in BO given that it is a non-parametric flexible model, 
it is easy to estimate its hyper-parameters and it is a robust model that is not expected to
over-fit with small amounts of data.

More formally, a GP is fully characterized by a zero mean and a covariance 
function or kernel $k(\mathbf{x},\mathbf{x}')$, that is, 
$f(\mathbf{x}) \sim \mathcal{G}\mathcal{P}(\mathbf{0},k(\mathbf{x},\mathbf{x}'))$. 
The covariance function of the GP receives two points as an input, $\mathbf{x}$ and 
$\mathbf{x}'$. We define the prior mean function $m(\mathbf{x})$ and the covariance 
function $k(\mathbf{x}, \mathbf{x}')$ that computes the covariance 
between $f(\mathbf{x})$ and $f(\mathbf{x}')$ as:
\begin{align}
& m(\mathbf{x}) = \mathbb{E}[f(\mathbf{x})]\,, \nonumber \\
& k(\mathbf{x}, \mathbf{x}') = \mathbb{E}[(f(\mathbf{x}) - m(\mathbf{x}))(f(\mathbf{x}') - m(\mathbf{x}'))]\,.
\label{eq:gppp}
\end{align}

We assume independent GPs for each black-box function, objective or constraint. Consider the observations of a particular black-box function $\{(\mathbf{x}_i, y_i)\}_{i=1}^N$, where $y_i = f(\mathbf{x}_i) + \epsilon_i$, with $f(\cdot)$ the black-box function and $\epsilon_i$ some Gaussian noise.
A GP gives a distribution for the potential values of
$f(\cdot)$ at a new set of input points $\mathbf{X}^\star=(\mathbf{x}_1^\star,\ldots,\mathbf{x}_B^\star)^\text{T}$
of size $B$.  Let $\mathbf{f}^\star = (f(\mathbf{x}_1^\star),\ldots,f(\mathbf{x}_B^\star))^\text{T}$.
The predictive distribution for $\mathbf{f}^\star$ is Gaussian.
$p(\mathbf{f}^\star|\mathbf{y}) = \mathcal{N}(\mathbf{f}^\star|\mathbf{m}(\mathbf{X}^\star), \mathbf{V}(\mathbf{X}^\star))$,
where $\mathbf{y}=(y_1,\ldots,y_N)^\text{T}$ and the mean and covariances are, respectively:
\begin{align}
\mathbf{m}(\mathbf{X}^\star) & = \mathbf{K}_\star^\text{T} (\mathbf{K} + \sigma^2\mathbf{I})^{-1} \mathbf{y}\,, &
\mathbf{V}(\mathbf{X}^\star) & = \mathbf{K}_{\star,\star} - \mathbf{K}_\star^\text{T}
 (\mathbf{K} + \sigma^2 \mathbf{I})^{-1}\mathbf{K}_\star\,.
\label{eq:gp_predictive}
\end{align}
In (\ref{eq:gp_predictive}) $\sigma^2$ is the variance of the Gaussian noise; $\mathbf{K}_\star$ is
a $N\times B$ matrix with the prior covariances between $\mathbf{f}^\star$
and each $f(\mathbf{x}_i)$; and $\mathbf{K}$ is a $N \times N$ matrix with the prior
covariances among each $f(\mathbf{x}_i)$. That is $K_{ij} = k(\mathbf{x}_i,\mathbf{x}_j)$, for
some covariance function $k(\cdot,\cdot)$. Common examples of covariance functions 
used by GPs are the squared exponential or the Mat\'ern function \citep{rasmussen2003gaussian}.
Finally, $\mathbf{K}_{\star,\star}$ is a $B \times B$ matrix with the prior covariances for each entry in $\mathbf{f}^\star$.

A GP model has a set of hyper-parameters $\boldsymbol{\theta}$ that can be adjusted to better fit 
the data $\mathcal{D} = \{(\mathbf{x}_i, y_i)| i = 1,...,N\}$. These include the variance 
of the additive Gaussian noise $\sigma^2$ and any potential hyper-parameter of the covariance 
function $k(\cdot,\cdot)$. These can be, \emph{e.g.}, the amplitude and the length-scales. 
Two popular approaches to estimate the values for these hyper-parameters w.r.t the data are: 
maximizing the log marginal likelihood and approximately computing a posterior distribution for 
the hyper-parameters. In this work, we will generate approximate samples 
from the posterior distribution of the GP hyper-parameters using slice sampling as in \citep{snoek2012practical}.
The predictive distribution of the GP is averaged over these samples.

\subsection{Batch Bayesian Optimization}

Any sequential BO strategy can be transformed into a batch one by
iteratively applying the sequential strategy $B$ times, with $B$ the size of the batch.  
To avoid choosing similar points each time, one can simply 
hallucinate the results of the already chosen pending evaluations \citep{snoek2012practical}. 
For this, the acquisition function is simply 
updated from $\alpha(\mathbf{x}|\mathcal{D})$ to $\alpha(\mathbf{x}|\mathcal{D} 
\bigcup {(\mathbf{x}_i , \mathbf{h}_i)\,, \forall \mathbf{x}_i \in \mathcal{P}})$, 
where $\mathcal{D}$ are the data collected so far, $\mathcal{P}$ is the set of pending 
evaluations, and $\mathbf{h}_i$ denotes the hallucinated evaluation result for the 
pending evaluation $\mathbf{x}_i$. A simple approach is to update the surrogate model after choosing 
each batch point by setting $\mathbf{h}_i$ equal to the mean of the predictive distribution 
given by the GPs \citep{desautels2014parallelizing}. Of course, this strategy, which we
refer to as parallel sequential, has the disadvantage of requiring the optimization of the
acquisition $B$ times, and also updating the GPs using hallucinated 
observations. This is expected to lead to extra computational cost than the one of our 
proposed approach, PPESMOC. 

\begin{figure}[ht]
\begin{center}
\includegraphics[width=0.475\textwidth]{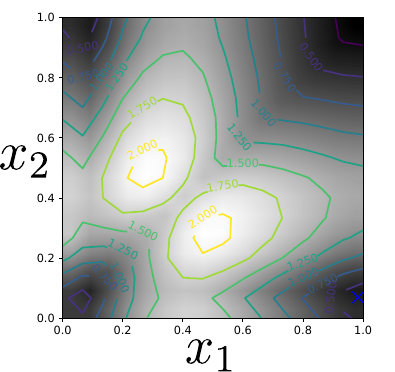}
\caption{Acquisition function visualization. The batch size is $B=2$.
The problem is in one dimension. Each axis corresponds to different values for one
of the $2$ points in the batch, taking values in the interval $[0,1]$.}
\label{fig:acq_fun_batch}
\end{center}
\end{figure}

As will be further developed in 
Section \ref{section3:PPESMOC}, PPESMOC chooses $\mathbf{X}_{N+1}$ as the batch 
of points that maximizes the expected reduction in the entropy of the Pareto set in 
the feasible space, $\mathcal{X}^\star$. Hence, the acquisition function will now have not only the $D$ dimensions
associated with the dimensions of the input space $\mathcal{X}$, but $B\times D$ dimensions, where $B$ is the 
size of the batch. We show in Figure \ref{fig:acq_fun_batch} an acquisition function example that 
takes into account every possible combination value of a batch size of two a one-dimensional problem 
in the interval $[0,1]$. As we can see, the acquisition function ends up having two dimensions despite 
being a one-dimensional input space $\mathcal{X}$. Of course, this is because the batch size is two. The bigger the
batch size the bigger the dimensionality of the acquisition function.

Because of this bigger dimensionality, optimizing the acquisition function is more challenging 
in the parallel setting than in the sequential setting. The use of approximations based on differences 
to compute the gradient is no longer feasible. These approximations of the gradient 
of the acquisition are common in the literature for sequential BO methods. For example, 
they are used, \emph{e.g.}, in PESMOC to approximate the gradient of the acquisition function
and to optimize it \citep{garrido2019predictive}. Approximating the gradient in the parallel setting using 
differences would require $2BD$ evaluations of the acquisition function, where $B$ is 
the batch size and $D$ the input dimensionality. As the gradient of the acquisition 
function needs to be computed at each step of its optimization (as when L-BFGS is used for this), 
it is necessary to compute the exact gradients of the acquisition function w.r.t the inputs. Hopefully,
automatic differentiation tools such as Auto-grad can be used for that purpose \citep{maclaurin2015autograd}.

\section{Parallel Predictive Entropy Search for Multi-Objective Bayesian Optimization with Constraints}
\label{section3:PPESMOC}

We now explain in detail the PPESMOC acquisition function. 
In Section \ref{subsec:pes}, we begin by explaining how we can measure
the entropy of the minimizer in the classical BO scenario and then describe the 
generalizations performed to that approach until the parallel constrained multi-objective 
scenario. We describe in Section \ref{subsec:ppesmoc} how PPESMOC can be used 
to identify such a batch of points by maximizing an acquisition function.

\subsection{Information-based Acquisition Functions}
\label{subsec:pes}

It is possible to model the minimizer of an optimization problem $\mathbf{x}^\star$ as a 
random variable \citep{hennig2012entropy,villemonteix2009informational,hernandez2014predictive}. 
Hence, we can compute the entropy $H[p(\mathbf{x}^\star)]$ of that random variable with associated probability 
density function $p(\mathbf{x}^\star)$. Intuitively, if we want to discover the location of the 
minimizer $\mathbf{x}^\star$, we need to minimize the entropy of the location of the 
extremum $H[p(\mathbf{x}^\star)]$ at every BO iteration. In order to do so, 
we can measure the amount of information that we have about the minimizer 
$\mathbf{x}^\star$ of the problem and try to increase it the most at each iteration.

The first acquisition function that used this approach is entropy search (ES) 
\citep{villemonteix2009informational, hennig2012entropy}. Specifically, ES, uses the differential 
entropy of probability distributions to choose the next point to evaluate. 
Consider $N$ observations $\mathcal{D}=\{ (\mathbf{x}_i,\mathbf{y}_i)\}_{i=1}^N$
of the black-boxes obtained so far.  In particular, we are interested in the point $\mathbf{x}$, whose expected
evaluation $y$, minimizes the expected differential entropy of the conditioned posterior 
distribution $\mathbb{E}_{\mathbf{y}}\{H[p(\mathbf{x}^\star|\mathcal{D} \cup (\mathbf{x},y))]\}$ of the 
optimum $\mathbf{x}^\star$. That is, the point $\mathbf{x}$ that maximizes the expected reduction 
in the differential entropy $H[\cdot]$ of the posterior for $\mathbf{x}^\star$. The analytical expression 
of ES represents this idea. We can write the ES acquisition function $\alpha(\mathbf{x})$ as:
\begin{equation}
\text{ES}(\mathbf{x}) = H[p(\mathbf{x}^\star|\mathcal{D})] - 
\mathbb{E}_{\mathbf{y}}\{H[p(\mathbf{x}^\star|\mathcal{D} \cup \{(\mathbf{x}, y)\})]\}\,,
\label{eq:acq_es}
\end{equation}
where the expectation is taken with respect to the predictive distribution 
$p(y|\mathcal{D}, \mathbf{x})$ of the black-box function 
$f(\mathbf{x})$. The problem with ES is that the exact computation of the above 
expression is infeasible in practice. Concretely, computing the probability distribution 
of the minimizer given the data $p(\mathbf{x}^\star|\mathcal{D})$ is intractable, 
having to resort to complex approximations \citep{villemonteix2009informational, hennig2012entropy}. 

To circumvent this issue, the PES acquisition function $\alpha(\mathbf{x})$ is an 
ES equivalent expression that does not require as many approximations as ES and it 
is easier to implement \citep{hernandez2014predictive}. Specifically, it is possible to perform a trick 
to convert the analytical expression of ES into another analytical expression that is 
easier to approximate. This trick is based on the concept of mutual information $I(X, Y)$ 
of random variables $X$ and $Y$. In particular, PES uses the fact that mutual information 
$I(X, Y)$ is symmetric. Let $\mathcal{D}$ be the dataset of all evaluations processed 
by BO until a given iteration $t$. Namely, $\mathcal{D} = \{(\mathbf{x}_i, y_i)| i = 1,...,t\}$. 
The ES expression in (\ref{eq:acq_es}) can be equivalently written as the mutual information 
between $\mathbf{x}^\star$ and $y$ given $\mathcal{D}$, namely, 
$ES(\mathbf{x}) = I(\mathbf{x}^\star, y)$. Since the mutual information 
$I(X, Y)$ is a symmetric function, we can swap the roles of 
$y$ and $\mathbf{x}^\star$ in the $\text{ES}(\mathbf{x})$ analytical 
expression. By doing it so, (\ref{eq:acq_es}) can be rewritten as:
\begin{equation}
\label{pesequation}
\text{PES}(\mathbf{x}) = H[p(y|\mathcal{D}, \mathbf{x})] - 
	\mathbb{E}_{p(\mathbf{x}^\star|\mathcal{D})}[H[p(y|\mathcal{D},\mathbf{x},\mathbf{x}^\star)]]\,,
\end{equation}
where $p(y|\mathcal{D},\mathbf{x},\mathbf{x}^\star)$ is the conditional 
predictive distribution for $y$ at $\mathbf{x}$ given the observed data $\mathcal{D}$ and 
the location $\mathbf{x}^\star$ of the global optimum of the objective function 
$f(\mathbf{x})$. The first term of PES's acquisition function is analytic. The second can 
be easily approximated. More precisely, the first term of PES $H[p(y|\mathcal{D}, \mathbf{x})]$ 
corresponds to the entropy of the GP predictive distribution 
$p(y|\mathcal{D}, \mathbf{x})$ of the objective $f(\mathbf{x})$, potentially 
contaminated with noise. This is essentially the entropy of an univariate 
Gaussian distribution. Nevertheless, the second term of PES in 
(\ref{pesequation}) must be approximated by the expectation propagation (EP) 
algorithm \citep{minka2001expectation,hernandez2014predictive}. 

The PES acquisition function has been generalized to optimize several objectives via 
PESMO \citep{hernandez2016predictive}, modifying the acquisition function to take 
into account several objectives. In PESMO, the acquisition function is modified 
as: 
\begin{align}
\alpha(\mathbf{x}) & = \text{H}[p(\mathbf{y}|\mathcal{D})] -
\mathbb{E}_{p(\mathbf{y}|\mathcal{D},\mathbf{x})}[\text{H}[p(\mathcal{X}^\star|\mathcal{D},\mathbf{x},\mathcal{X}^\star)]]\,,
        \label{eq:initial_acq_orig}
\end{align}
where $\mathcal{X}^\star$ is the Pareto set, which represents the solution to the optimization problem,
and $p(\mathbf{y}|\mathcal{D})$ is the GP predictive distribution for each objective.
PES was also generalized to take into account 
several constraints in PESC \citep{hernandez2015predictive}. Here, the notion of feasibility is 
introduced and a point is only valid if it satisfies all the constraints, that is, 
$c_1(\mathbf{x}) \geq 0, \ldots, c_C(\mathbf{x}) \geq 0$. Every constraint is modeled 
using a GP and the acquisition function takes into account the predictive distributions of each
black-box, including the constraints. As in the other PES acquisition functions, EP is used to approximate 
the intractable factors. Finally, PESMO and PESC were generalized to tackle multiple 
objectives under several constraints simultaneously in PESMOC \citep{garrido2019predictive}. 
PESMOC introduced the feasible Pareto set and uses also EP to approximate both the 
intractable factors of the conditional distributions that appear in the acquisition function.
In the next subsection, we present PPESMOC, that generalizes the previous works to 
the parallel constrained multi-objective scenario. Such an scenario leads to several issues that are addressed
in this work.  We note that a parallel version of PES for the unconstrained single objective case 
is described in \citep{shah2015parallel}.

\subsection{Specification of the Acquisition Function}
\label{subsec:ppesmoc}

In this section we describe the proposed acquisition function which leads to the proposed method PPESMOC.
Let us define $\mathbf{X}_{N+1}=\{\mathbf{x}_1,\ldots,\mathbf{x}_B\}$ as the 
batch of $B$ points where the black-boxes should be evaluated at the next iteration. 
We choose $\mathbf{X}_{N+1}$ as the batch of points that maximizes the expected reduction
in the entropy of the Pareto set in the feasible space, $\mathcal{X}^\star$. This is a popular strategy 
that has shown good empirical results in other sequential optimization settings, including the 
PESMOC acquisition function \citep{garrido2019predictive}.
Starting from this idea, the PPESMOC acquisition function is defined as:
\begin{align}
\alpha(\mathbf{X}) & = \text{H}[p(\mathcal{X}^\star|\mathcal{D})] - 
\mathbb{E}_{p(\mathbf{Y}|\mathcal{D},\mathbf{X})}[\text{H}[p(\mathcal{X}^\star|\mathcal{D} \cup (\mathbf{X},\mathbf{Y})]]\,,
	\label{eq:initial_acq}
\end{align}
where $\mathbf{X}$ is the candidate batch of $B$ points at which to evaluate the black-boxes;
$\mathbf{Y}$ is a matrix with the set of $B$ noisy evaluations associated to $\mathbf{X}$, for each black-box function; 
$\text{H}[p(\mathbf{x})] = - \int p(\mathbf{x}) \log p(\mathbf{x})d\mathbf{x}$ is the differential entropy of 
the distribution $p(\mathbf{x})$; the expectation is with respect to the posterior predictive 
distribution of $\mathbf{Y}$ at the candidate batch $\mathbf{X}$, given the data we have observed so far, $\mathcal{D}$; 
finally, $p(\mathcal{X}^\star|\mathcal{D})$ is the probability distribution of potential Pareto sets $\mathcal{X}^\star$ 
given the data we have observed so far $\mathcal{D}$.
Note that the distribution $p(\mathbf{Y}|\mathcal{D},\mathbf{X})$, is given by the product of the predictive
distributions of each GP, as indicated in (\ref{eq:gp_predictive}), 
for each black-box function. Namely, $p(\mathbf{Y}|\mathcal{D},\mathbf{X})=\prod_{k=1}^K p(\mathbf{y}_k^o|\mathcal{D},\mathbf{X})
\prod_{j=1}^J p(\mathbf{y}_j^c|\mathcal{D},\mathbf{X})$, where $\mathbf{y}_k^o$ and $\mathbf{y}_j^c$ are $B$-dimensional
vectors with the potential observations of each black-box function, objective or constraint, for each point in the 
batch $\mathbf{X}$. Recall that an independent GP is modeling each black-box function. 

Note that (\ref{eq:initial_acq}) involves the entropy of the Pareto set, $\mathcal{X}^\star$, which can be very 
difficult to compute.  To simplify the computation of the acquisition function, we use the same trick
based on the symmetry of mutual information that we described in the previous section.
For this, we observe that (\ref{eq:initial_acq}) is the mutual information between $\mathcal{X}^\star$ and $\mathbf{Y}$,
$\text{I}(\mathcal{X}^\star,\mathbf{Y})$. Since the mutual information is symmetric, 
\emph{i.e.}, $\text{I}(\mathcal{X}^\star,\mathbf{Y})=\text{I}(\mathbf{Y},\mathcal{X}^\star)$,
we swap the roles of $\mathcal{X}^\star$ and $\mathbf{Y}$ obtaining:
\begin{align}
\alpha(\mathbf{X}) = \text{H}[p(\mathbf{Y}|\mathcal{D},\mathbf{X})] - 
	\mathbb{E}_{p(\mathcal{X}^\star|\mathcal{D})}[\text{H}[p(\mathbf{Y}|\mathcal{D},\mathbf{X},\mathcal{X}^\star)]],
\label{eq:acq_simplified1}
\end{align}
where $p(\mathbf{Y}|\mathcal{D},\mathbf{X},\mathcal{X}^\star)$ is the predictive distribution for 
the values of the black-boxes at $\mathbf{X}$, given the observed data $\mathcal{D}$, 
and given that the solution of the optimization problem. Namely, the Pareto set in the 
feasible space, given by $\mathcal{X}^\star$. Furthermore, the expectation is with respect to
$p(\mathcal{X}^\star|\mathcal{D})$. Namely, the posterior distribution of $\mathcal{X}^\star$
given the data we have observed so far $\mathcal{D}$.

Importantly, the first term in (\ref{eq:acq_simplified1}) can be evaluated analytically since it is just 
the entropy of the predictive distribution, $\text{H}[p(\mathbf{Y}|\mathcal{D},\mathbf{X})]$, 
which is a factorizing $K+J$ dimensional multivariate Gaussian. In particular,
\begin{align}
\text{H}[p(\mathbf{Y}|\mathcal{D},\mathbf{X})] 
 & = 0.5 ((K+J)B\log(2\pi e)+\sum_{k=1}^{K} \log|\mathbf{V}_k^o(\mathbf{X})| + \sum_{j=1}^{J}|\mathbf{V}_j^c(\mathbf{X})| )
\end{align}
where $\mathbf{V}_k^o(\mathbf{X})$ and $\mathbf{V}_j^c(\mathbf{X})$ are the covariance matrices of the predictive
distribution for each black-box function (objective or constraint, respectively) given by (\ref{eq:gp_predictive}),
plus the corresponding variance of the additive Gaussian noise (\emph{i.e.}, $\mathbf{I}\sigma^2$).

Note that the expectation in (\ref{eq:acq_simplified1}) can be approximated by a Monte Carlo average. 
More precisely, one can generate random samples of the black-box functions using a random-feature 
approximation of each GP, as in \cite{garrido2019predictive}. These samples can then be easily 
optimized to generate a sample from $p(\mathcal{X}^\star|\mathcal{D})$.  
Because the samples of the black-box function are cheap to evaluate, this optimization process has little cost 
and can be done using, \emph{e.g.}, a grid of points. In practice, we use a finite Pareto set approximated by 
$50$ points. A problem, however, is evaluating the second term that appears in (\ref{eq:acq_simplified1}). Namely, 
the entropy of $p(\mathbf{Y}|\mathcal{D},\mathbf{X},\mathcal{X}^\star_s)$, for
a particular sample of $\mathcal{X}^\star$, $\mathcal{X}^\star_s$. Such a distribution is intractable.
As in the case of PESMOC, we resort to expectation propagation to approximate its value \citep{minka2001expectation}.

\subsection{Approximating the Conditional Predictive Distribution}

Assume both $\mathcal{X}$ and $\mathcal{X}^\star$ have finite size and that $\mathcal{X}^\star$ is known. 
Later on, we will show how to approximate $\mathcal{X}$ with a finite size set.
Let $\mathbf{F}$ and $\mathbf{C}$ be a matrix with the actual
objective and constraint values associated to $\mathcal{X}$. Then,
\begin{align}
p(\mathbf{Y}|\mathcal{D},\mathbf{X},\mathcal{X}^\star) = & 
	\int p(\mathbf{Y}|\mathbf{X},\mathbf{F},\mathbf{C}) p(\mathcal{X}^\star|\mathbf{F},\mathbf{C})
	p(\mathbf{F}|\mathcal{D}) p(\mathbf{C}|\mathcal{D}) d\mathbf{F} d \mathbf{C}\,,
	\label{eq:conditional_pred1}
\end{align}
where $p(\mathbf{Y}|\mathbf{X},\mathbf{F},\mathbf{C})=\prod_{b=1}^B \prod_{k=1}^K 
\delta(y^k_b-f_k(\mathbf{x}_b)) \prod_{j=1}^J \delta(y^j_b - c_j(\mathbf{x}_b))$, with
$y^k_b$ the evaluation corresponding to the $k$-th objective associated to the batch point $\mathbf{x}_b$,
$y^j_b$ the evaluation corresponding to the $j$-th constraint associated to the batch point $\mathbf{x}_b$,
$\delta(\cdot)$ a Dirac's delta function and $B$ the batch size.
We have assumed no additive Gaussian noise. In the case of noisy observations, one
simply has to replace the delta functions with Gaussians with
the corresponding variance, $\sigma^2$.

In (\ref{eq:conditional_pred1}) $p(\mathbf{F}|\mathcal{D})$ and $p(\mathbf{C}|\mathcal{D})$
denote the posterior predictive distribution for the objectives and constraints, respectively. Note that
we assume independent GPs. Therefore, these distributions factorize across objectives and constraints.
They are Gaussians with parameters given in (\ref{eq:gp_predictive}).
Last, in (\ref{eq:conditional_pred1}) $p(\mathcal{X}^\star|\mathbf{F},\mathbf{C})$ is an informal probability
distribution that takes value different from zero, only for a valid Pareto set $\mathcal{X}^\star$. 
More precisely, $\mathcal{X}^\star$ has to satisfy that $\forall \mathbf{x}^\star \in 
\mathcal{X}^\star$, $\forall \mathbf{x}' \in \mathcal{X}$, $c_j(\mathbf{x}^\star) \geq 0$, 
$\forall j$, and if $c_j(\mathbf{x}') \geq 0$, $\forall j$, then $\exists k$ s.t.
$f_k(\mathbf{x}^\star) < f_k(\mathbf{x}')$. Namely, each point of the Pareto set
has to be better than any other feasible point in at least one of the objectives.
These conditions can be summarized as:
\begin{align}
p(\mathcal{X}^\star|\mathbf{F},\mathbf{C}) \propto 
	\prod_{\mathbf{x}^\star \in \mathcal{X}^\star}
	\left( \left[ \prod_{j=1}^J \Phi_j(\mathbf{x}^\star) \right] \left[ \prod_{\mathbf{x}'\in \mathcal{X}}
	\Omega(\mathbf{x}',\mathbf{x}^\star)\right]
	\right) 
	\label{eq:prob_pareto_set1}
\end{align}
where $\Phi_j(\mathbf{x}^\star)=\Theta(c_j(\mathbf{x}^\star))$, with $\Theta(\cdot)$ the
Heaviside step function, using the convention that $\Theta(0)=1$. 
Furthermore, 
\begin{align}
\Omega(\mathbf{x}',\mathbf{x}^\star) &= \left[ \prod_{j=1}^J \Theta(c_j(\mathbf{x}')) \right]
\Psi(\mathbf{x}',\mathbf{x}^\star) + \left[1 - \prod_{j=1}^J \Theta(c_j(\mathbf{x}')) \right] \cdot 1\,,
\end{align}
where $\Psi(\mathbf{x}',\mathbf{x}^\star) = 1 - \prod_{k=1}^K \Theta(f_k(\mathbf{x}^\star) - f_k(\mathbf{x}'))$.
The goal of $\prod_{j=1}^J \Phi_j(\mathbf{x}^\star) $ in Equation (\ref{eq:prob_pareto_set1}) is to guarantee that every
point in $\mathcal{X}^\star$ is feasible. Otherwise, $p(\mathcal{X}^\star|\mathbf{F},\mathbf{C})$ takes value zero.
Similarly, $\Omega(\mathbf{x}',\mathbf{x}^\star)$ can be understood as follows:
$\prod_{j=1}^J \Theta(c_j(\mathbf{x}'))$ checks that $\mathbf{x}'$ is feasible. If $\mathbf{x}'$
is infeasible, we do not care and simply multiply everything by $1$. Otherwise, $\mathbf{x}'$ has to be
dominated by $\mathbf{x}^\star$. That is checked by $\Psi(\mathbf{x}',\mathbf{x}^\star)$. This last factor takes
value one if $\mathbf{x}^\star$ dominates $\mathbf{x}'$ and zero otherwise.
Summing up, the r.h.s. of Equation (\ref{eq:prob_pareto_set1}) takes value $1$ only if $\mathcal{X}^\star$ is a valid Pareto set.

Critically, in Equation (\ref{eq:conditional_pred1}) all the factors that appear in the r.h.s are Gaussian, 
except for $p(\mathcal{X}^\star|\mathbf{F},\mathbf{C})$. The non-Gaussian factors contained in this
distribution are approximated by Gaussians using expectation propagation (EP) \citep{minka2001expectation}.
Each $\Phi_j(\mathbf{x}^\star)$ factor is approximated by a univariate Gaussian that need not be normalized.
Namely, $\Phi_j(\mathbf{x}^\star)\approx 
\tilde{\mathcal{N}}(c_j(\mathbf{x}^\star|\tilde{m}_j^{\mathbf{x}^\star},\tilde{v}_j^{\mathbf{x}^\star}))$.
The parameters of this Gaussian are tuned by EP. Similarly, each $\Omega(\mathbf{x}',\mathbf{x}^\star)$ is approximated
by a product of $K$ bivariate Gaussians and $J$ univariate Gaussians that need not be normalized. 
That is, 
\begin{align}
\Omega(\mathbf{x}',\mathbf{x}^\star)\approx \prod_{k=1}^K 
\tilde{\mathcal{N}}(\bm{\upsilon}|
\tilde{\mathbf{m}}_k^{\mathbf{x}^\star,\mathbf{x}'}, \tilde{\mathbf{V}}_k^{\mathbf{x}^\star,\mathbf{x}'})
\prod_{j=1}^J \tilde{\mathcal{N}}(c_j(\mathbf{x}')|\tilde{m}_j^{\mathbf{x}'},\tilde{v}_j^{\mathbf{x}'})),
\end{align}
where $\bm{\upsilon}=(f_k(\mathbf{x}^\star),f_k(\mathbf{x}'))^\text{T}$. The parameters of these Gaussians are also
adjusted by EP. The approximate factors are refined iteratively until their
parameters do not change. This ensures that they look similar to 
the corresponding exact factors. 

In our experiments, and when running EP, we replace the set $\mathcal{X}$ in Equation (\ref{eq:prob_pareto_set1}) 
by a finite set given by $\{\mathbf{x}_n\}_{n=1}^N \bigcup \mathbf{X} \bigcup \mathcal{X}^\star$. Namely, the union
of all points that have already being evaluated, the candidate batch $\mathbf{X}$ and the current Pareto
set $\mathcal{X}^\star$ that has been sampled from $p(\mathcal{X}^\star|\mathcal{D}$) when using a Monte
Carlo approximation of the expectation in the r.h.s. of Equation (\ref{eq:acq_simplified1}). 
These are the input points we have so far. In practice, the $\Omega(\mathbf{x}',\mathbf{x}^\star)$ factors 
corresponding to $\mathbf{X}$, \emph{i.e.}, the batch of points at which to evaluate the 
acquisition are only refined one time by EP, as in PESMOC \citep{garrido2019predictive}. This ensures that the 
acquisition function can be quickly evaluated.

\subsection{PPESMOC's Acquisition Function}
\label{sec:ppesmoc}

After EP has converged, the conditional predictive distribution in Equation (\ref{eq:conditional_pred1}) is approximated 
by replacing each non-Gaussian factor by the corresponding Gaussian approximation obtained by EP. Because all 
factors are then Gaussian, and the Gaussian family is closed under the product operation, their product can 
be easily evaluated, resulting in another Gaussian distribution for 
$p(\mathbf{Y}|\mathcal{D},\mathbf{X},\mathcal{X}^\star)$. 
Consider $S$ Monte Carlo samples of $\mathcal{X}^\star$ to approximate the expectation in the r.h.s. 
of Equation (\ref{eq:acq_simplified1}). Let $\mathbf{V}_k^o(\mathbf{X};s)^\text{CPD}$ and 
$\mathbf{V}_j^c(\mathbf{X};s)^\text{CPD}$ denote the covariance matrices of the Gaussian approximation of 
$p(\mathbf{Y}|\mathcal{D},\mathbf{X},\mathcal{X}^\star)$ for each objective $k$ and constraint $j$, for sample 
$\mathcal{X}^\star_s$. The PPESMOC's acquisition is simply given by the difference in the entropy 
before and after conditioning to $\mathcal{X}^\star$. Namely,
\begin{align}
\alpha(\mathbf{X}) &= \sum_{k=1}^K \log |\mathbf{V}_k^o(\mathbf{X})| + \sum_{j=1}^J \log |\mathbf{V}_k^c(\mathbf{X})| 
	\nonumber \\
	& \quad 
	- \frac{1}{S} \sum_{s=1}^S \left[  \sum_{k=1}^K \log |\mathbf{V}_k^o(\mathbf{X};s)^\text{CPD}| + 
	\sum_{j=1}^J \log |\mathbf{V}_k^c(\mathbf{X};s)^\text{CPD}|  \right]\,.
	\label{eq:acquisition}
\end{align}
The cost of computing Equation (\ref{eq:acquisition}), assuming a constant number of iterations of EP until convergence,
is in $\mathcal{O}((K+J)N^3+(K+J)B^3)$, where $N$ is the number of points observed so far, and $B$ is the batch size.
This cost is a consequence of the GP inference process and having to compute the determinant of matrices of size $B \times B$.
Importantly, the gradients of Equation (\ref{eq:acquisition}) w.r.t $\mathbf{X}$ can be readily computed using automatic
differentiation tools such as Autograd (\url{https://github.com/HIPS/autograd}) \citep{maclaurin2015autograd}.
This is key to guarantee that the acquisition can be optimized using, \emph{e.g.}, quasi-Newton methods (L-BFGS), 
to find $\mathbf{X}_{N+1}$. In our implementation we run EP until convergence one time to approximate the factors 
corresponding to the observed data $\mathcal{D}$ and the sampled Pareto set $\mathcal{X}^\star_s$. These factors 
are reused and those corresponding to the candidate batch $\mathbf{X}_{N+1}$ are refined only one time. This 
saves  computational time. We implemented PPESMOC in the software for BO Spearmint. Our implementation is 
available at \url{https://github.com/EduardoGarrido90/spearmint_ppesmoc}. 

Note that (\ref{eq:acquisition}) includes a sum across each objective and constraint.  The consequence is that
the acquisition function can be expressed as a sum of one acquisition function per objective and constraint. Namely,
$\alpha(\mathbf{X})=\sum_{k=1}^K \alpha_k^o(\mathbf{X}) + \sum_{j=1}^J \alpha_j^c(\mathbf{X})$.
This enables the use of PPESMOC decoupled evaluations settings in which one chooses not only where to evaluate
the black-boxes, but which black-box to evaluate \citep{garrido2019predictive}. 
For this, one only has to maximize independently each $\alpha_k^o(\mathbf{X})$
and $\alpha_j^c(\mathbf{X})$ to choose then the black-box that leads to the largest expected decrease in the
entropy of $\mathcal{X}^\star$. A decoupled evaluation setting is expected to lead to better optimization results.
However, we leave its empirical evaluation for future work.

\begin{figure}[ht]
\begin{center}
\includegraphics[width=0.45\textwidth]{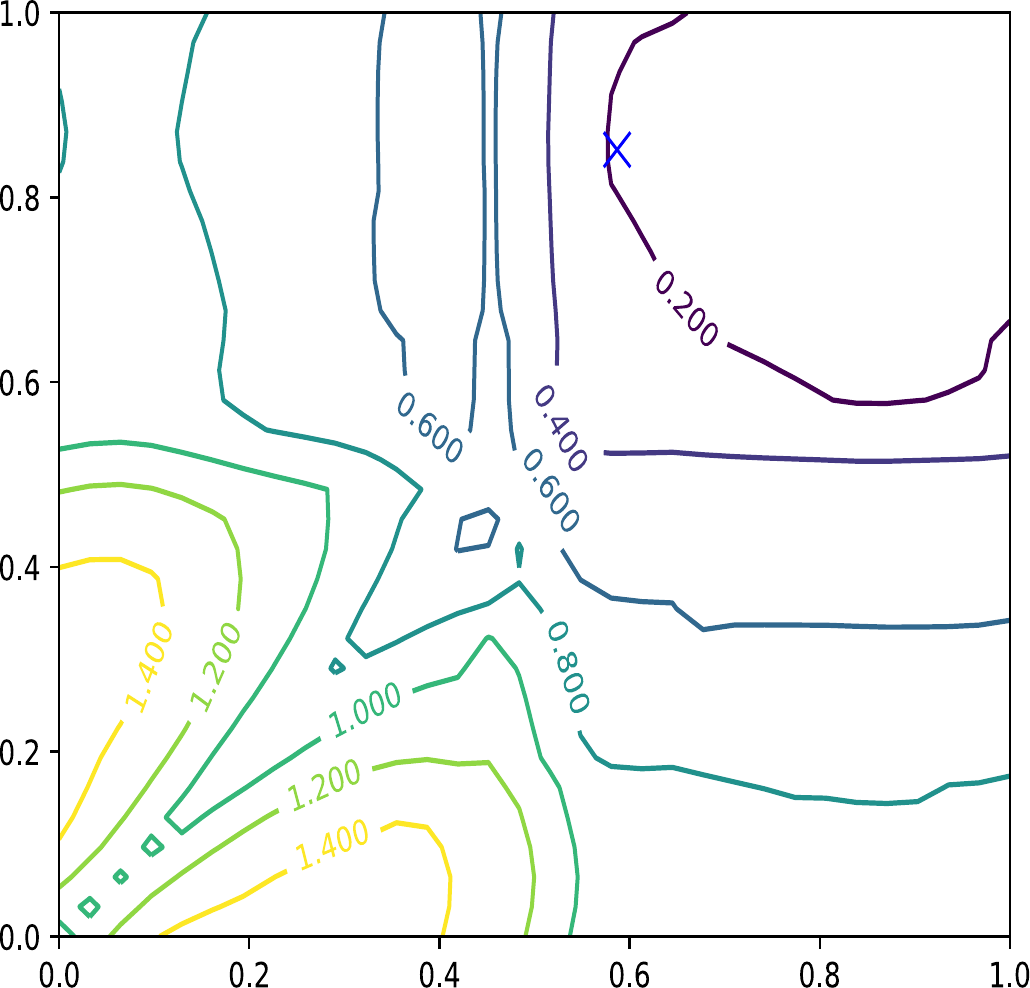}
\includegraphics[width=0.45\textwidth]{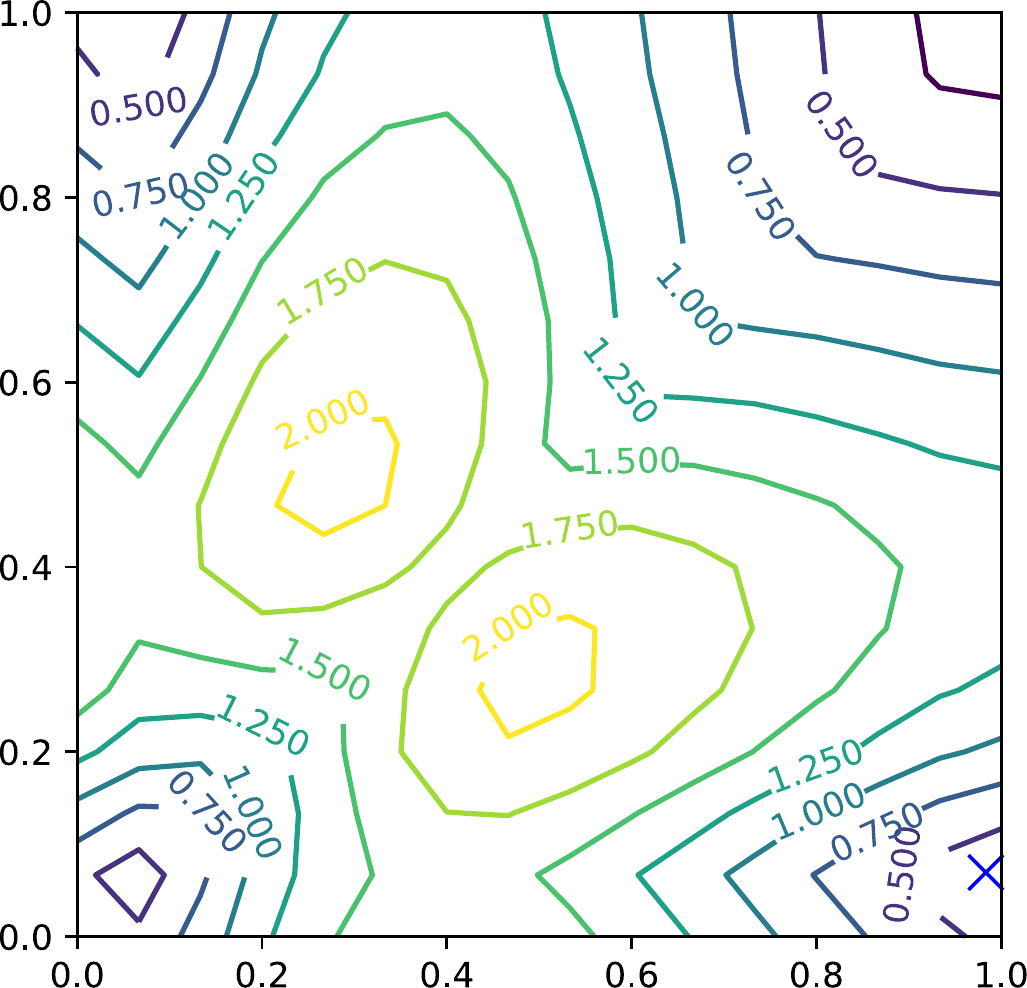}
\caption{Acquisition function visualization.
Each figure corresponds to a different repetition of the optimization problem. The batch size is $B=2$.
The problem is in one dimension. Each axis corresponds to different values for one
of the $2$ points in the batch, taking values in the interval $[0,1]$. 
Blue crosses show already evaluated locations. }
\label{fig:ppesmoc}
\end{center}
\end{figure}

We illustrate the PPESMOC's acquisition function in two scenarios of a $1$-dimensional problem (for the sake of 
visualization) with batch size $B=2$ in Figure \ref{fig:ppesmoc} (left) and (right). 
This problem has two objectives and two constraints. The observations obtained so 
far are displayed with a blue cross (previous evaluated batch). Each axis corresponds to 
the potential values (in the interval $[0,1]$) for each one of the two points in the batch. 
Figure \ref{fig:ppesmoc} displays the contour curves of the acquisition function. 
We remark here some of its properties: (i) It is symmetric w.r.t the diagonal, meaning that the order 
of the points in the batch does not affect its value. (ii) In the neighborhood of the observed 
point the acquisition value is low, meaning that the acquisition favors unexplored regions. 
(iii) In the diagonal the acquisition takes lower values, meaning that it favors diversity in 
the batch. These are expected properties of a batch BO acquisition function.
In these experiments the number of Monte Carlo samples $S$ is set equal to $10$. This is also 
the number of samples used for the GP hyper-parameters.

\subsection{Quality of the Approximation to the Acquisition Function}

As described previously, the acquisition function of the proposed method, PPESMOC, is intractable and needs to be
approximated. The exact evaluation requires computing an expectation that has no closed form solution
and computing the conditional predictive distribution of the probabilistic models given some Pareto set $\mathcal{X}^\star$.
In Section \ref{sec:ppesmoc} we propose to approximate these quantities using Monte Carlo samples and expectation propagation, respectively.
In this section we check the accuracy of this approximation to see if it resembles the actual acquisition function.
For this, we consider a simple one dimensional problem with a batch of two points, two objectives and one constraint generated from a
GP prior. In this simple setting, it is possible to compute a more accurate estimate of the acquisition function
using a more expensive sampling technique, combined with a non-parametric estimator of entropy \citep{singh2003nearest}.
More precisely, we discretize the input space and generate a sample of the Pareto set $\mathcal{X}^\star$ by optimizing
a sample of the black-box functions. This sample is generated as in the PPESMOC approximation. We then generate
samples of the black-box functions and keep only those that are compatible with $\mathcal{X}^\star$ being the
solution to the optimization problem. This process is repeated $100,000$ times. Then, a non-parametric method is used
to estimate the entropy of the predictive distribution at each possible batch of the input space before and after the conditioning.
The difference in the entropy at each input location gives a more accurate estimate of the acquisition function of PPESMOC.
Of course, this approach is too expensive to be used in practice for solving optimization problems.

\begin{figure}[ht]
\begin{center}
\begin{tabular}{cc}
\multicolumn{2}{c}{\includegraphics[width=0.8\textwidth]{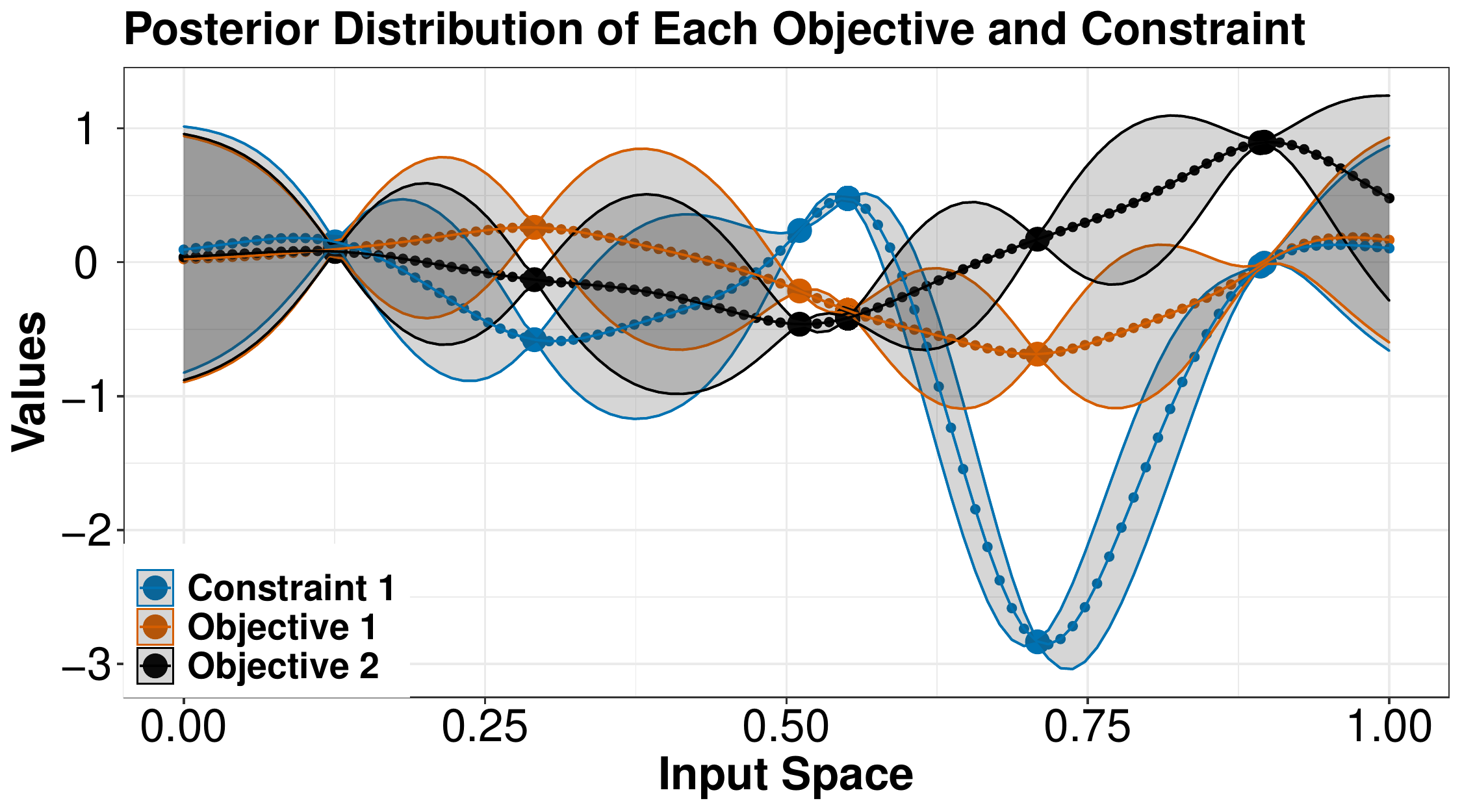}} \\
\includegraphics[width=0.475\textwidth]{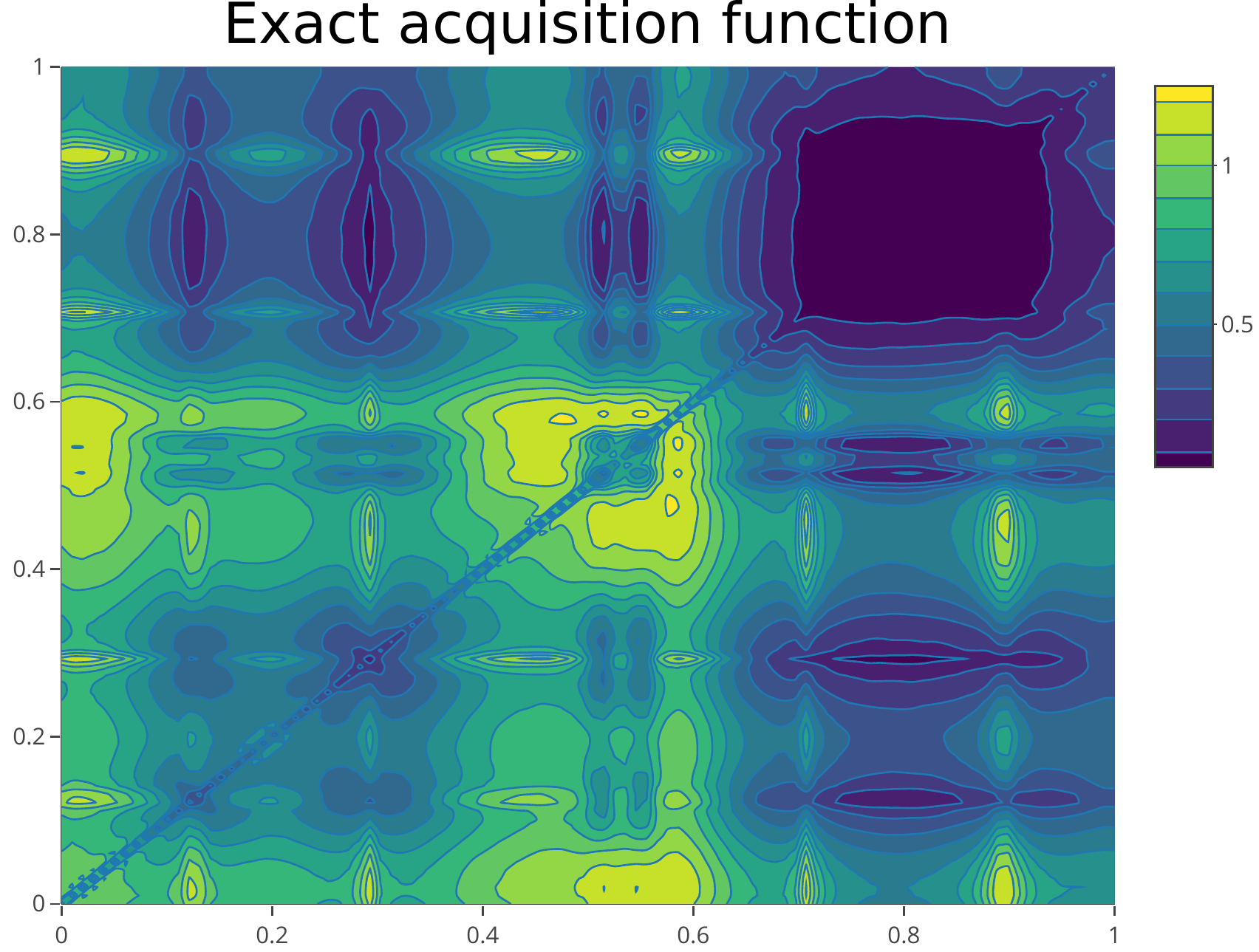}&
\includegraphics[width=0.475\textwidth]{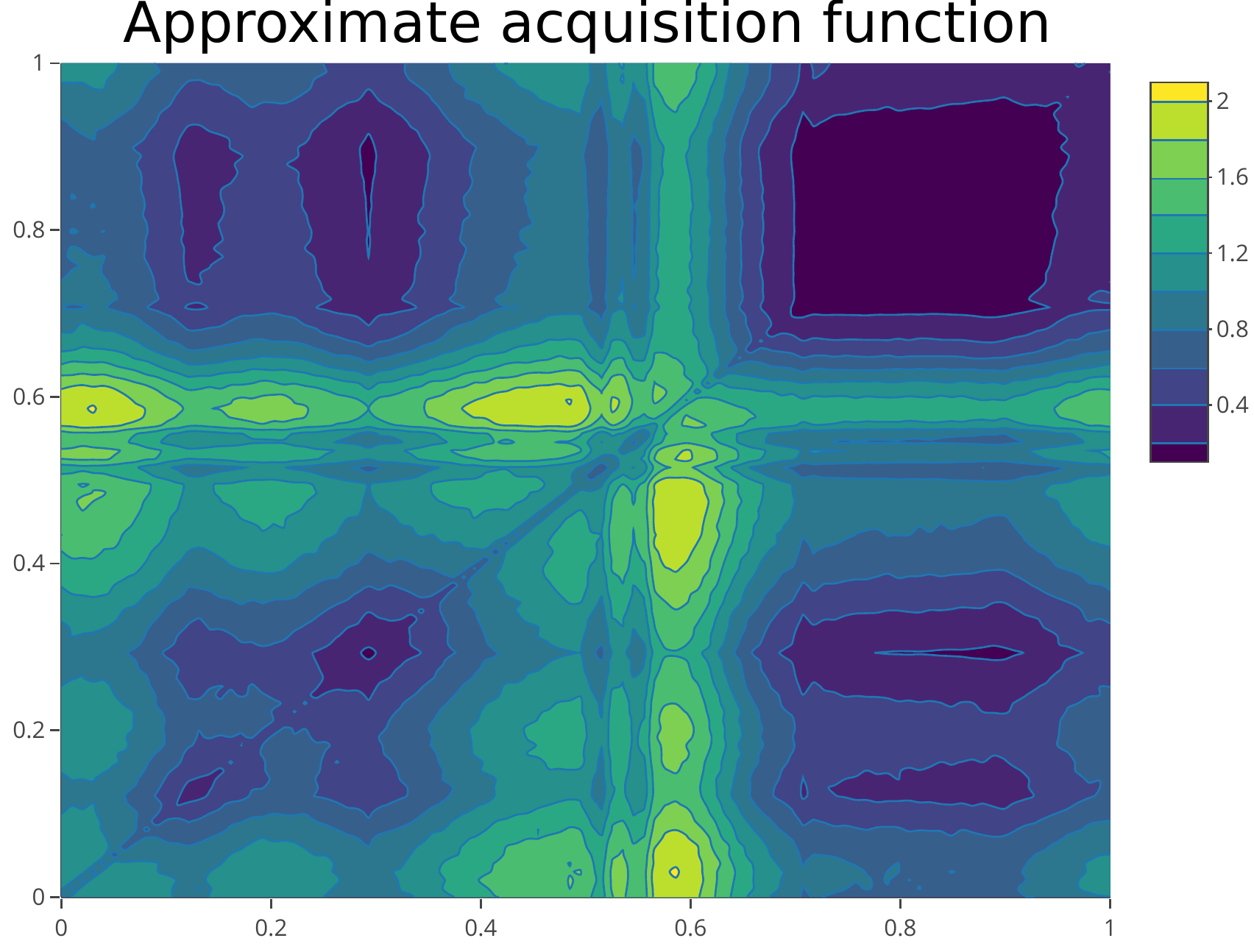}
\end{tabular}
\caption{Acquisition function visualization. The batch size is $B=2$.
The problem is in one dimension. 
(top) GP fit to each objective and constraint. (bottom right) Approximate acquisition function. (bottom left) Exact 
acquisition function computed via expensive Monte Carlo sampling.
Each axis corresponds to different values for one of the $2$ points in the batch, taking values in the interval $[0,1]$.
Note that both acquisitions are low in unfeasible regions.
}
\label{fig:exact_coupled_ppesmoc}
\end{center}
\end{figure}

Figure \ref{fig:exact_coupled_ppesmoc} shows a comparison between the two estimates of the exact acquisition function.
In both plots, the acquisition function values are displayed for all the batch combinations of the input space. 
The one described above (left, exact) and the one suggested in Section \ref{sec:ppesmoc} (right, approximate).
We observe that both estimates of the acquisition function take higher values in regions with high
uncertainty and promising predictions. Similarly, both estimates take lower values in regions with low uncertainty.
Importantly, both acquisition functions are pretty similar in the sense that they take high and low values in the
same regions of the input space. Therefore, both acquisition functions are extremely correlated. This
empirical result supports that the approximation proposed in this work is an accurate estimate of the
actual acquisition function.

\section{Related Work}\label{sec:related_work_ppesmoc}
\label{section4:rw}
In this section we review other works from the literature that describe related batch BO 
methods and BO methods that can address multi-objective problems with constraints.
Nevertheless, to our knowledge PPESMOC is the only batch BO method for constrained multi-objective 
problems. PPESMOC is related to parallel predictive entropy search (PPES) 
\citep{shah2015parallel}. At each iteration, PPES also selects a batch of points 
maximizing the expected information gain about the global maximizer of the objective. 
The computations are also approximated using expectation propagation (EP).
The main difference is that PPES is limited to single-objective and un-constrained
optimization problems, unlike PPESMOC, which can address multiple objectives and several 
constraints. This is a non-trivial extension of PPES. In particular, several objectives and constraints 
require the use of several GPs, not only one. Furthermore, including constraints and several
objectives leads to more complicated non-Gaussian factors that need to be approximated using
EP. Therefore, the EP update operations of PPESMOC are significantly more complicated than
those of PPES. Moreover, the evaluation of the acquisition function in PPES was expensive, 
making the optimization of the acquisition function unfeasible for batch sizes greater than 3. In order 
to circumvent this issue, the PPESMOC acquisition function has been optimized via reverse mode differentiation
using Autograd \citep{maclaurin2015autograd}. Additionally, solving multi-objective problems is also 
more complicated than solving a single-objective problem. In particular, the solution of 
the later is the Pareto set, a set of potentially infinite size.

Other batch BO methods form the literature include local penalization. This is 
an heuristic based on an estimate of the Lipschitz constant of the function that penalizes 
the acquisition function in each neighborhood of already selected points 
\citep{gonzalez2016batch}. The penalized acquisition function is used to collect batches of points, minimizing 
the  non-parallelizable computational effort. The advantage of this method is 
that it can be used with arbitrary acquisition functions. A limitation, however, 
is that it requires to fix the amount of penalization, which depends on the 
Lipschitz constant of the objective.  Such a constant needs to be estimated from data. Another limitation is that this
approach computes every point of the batch separately, it transforms the acquisition function based on the previous
selected batch points but it does not compute them simultaneously, as in the PPESMOC case, selecting the best possible
combination. Furthermore, \cite{gonzalez2016batch} only addresses 
unconstrained single-objective problems.

Hybrid batch Bayesian optimization dynamically switches, based on the current 
state, between sequential and batch evaluation policies with 
variable batch sizes \citep{azimi2012hybrid}. This strategy uses expected improvement as the 
acquisition function. However, it can only address unconstrained single-objective 
problems. A batch BO approach can also be implemented via a multi-objective ensemble  of multiple acquisition 
functions \citep{lyu2018batch}. In each iteration, a multi-objective optimization of multiple 
acquisition functions is carried out. A sample of points from the Pareto set is then selected 
as the batch of points to evaluate. Even though this strategy can address multi-objective problems, 
it cannot deal with constraints in the optimization process. 
Other strategies for batch 
BO that do not address the constrained multi-objective setting 
include \cite{desautels2014parallelizing,gupta2018exploiting,kathuria2016batched,daxberger2017distributed}.

Any sequential BO strategy can be transformed into a batch one by
iteratively applying the sequential strategy $B$ times.  
To avoid choosing similar points each time, one can simply 
hallucinate the results of the already chosen pending evaluations \cite{snoek2012practical}. 
For this, the acquisition function is simply 
updated from $\alpha(\mathbf{x}|\mathcal{D})$ to $\alpha(\mathbf{x}|\mathcal{D} 
\bigcup {(\mathbf{x}_i , \mathbf{h}_i)\,, \forall \mathbf{x}_i \in \mathcal{P}})$, 
where $\mathcal{D}$ are the data collected so far, $\mathcal{P}$ is the set of pending 
evaluations, and $\mathbf{h}_i$ denotes the hallucinated evaluation result for the 
pending evaluation $\mathbf{x}_i$. A simple approach is to update the surrogate model after choosing each 
batch point by setting $\mathbf{h}_i$ equal to the mean of the predictive distribution 
given by the GPs \cite{desautels2014parallelizing}. Of course, this strategy, which we
refer to as parallel sequential, has the disadvantage of requiring the optimization of the
acquisition $B$ times, and also updating the GPs using hallucinated 
observations. This is expected to lead to extra computational cost than in PPESMOC.

PPESMOC is a generalization of PESMOC, as described in \cite{garrido2019predictive}. PESMOC is 
the current state-of-the-art for solving constrained multi-objective BO problems. 
Nevertheless, PESMOC is a sequential BO method that can only suggest one point at a time to 
be evaluated. It cannot suggest a batch of points as PPESMOC. PESMOC also works by choosing the 
next candidate point as the one that is expected to reduce the most the entropy of the Pareto set in the feasible 
space. The required computations are also approximated using expectation propagation algorithm \cite{minka2001expectation}. 
Notwithstanding, the extension of PPESMOC over PESMOC is not trivial. In particular, in PPESMOC the 
acquisition function involves additional non-Gaussian factors (one per each point in the batch) and 
requires the computation of its gradients. These are not needed in PESMOC 
as the input dimensionality in that method is smaller, \emph{i.e.}, $D$ vs. $B \times D$.
PESMOC approximates the gradient by differences, which is too expensive in the parallel setting.

Bayesian Multi-objective optimization (BMOO) is another strategy for constrained multi-objective BO 
in the sequential setting \cite{feliot2017bayesian}. BMOO is based on the expected hyper-volume improvement 
acquisition function (EHI) in which the expected increase in the hyper-volume of the Pareto front is computed 
for each candidate point. The hyper-volume is simply the volume of points in functional space above the Pareto front 
(\emph{i.e.}, the function values associated to the Pareto set), which is maximized by the actual Pareto set. 
It is hence a natural measure of utility. When several constraints are introduced in the problem, this criterion 
boils down to the product of a modified EHI criterion (where only feasible points are considered) and the 
probability of feasibility, as indicated by the probabilistic models, which are also GPs.
BMOO is, however, restricted to the sequential evaluation setting. BMOO was designed originally for
a noiseless evaluation setting. However, it has also been shown to be able to solve optimization problems in 
the noisy setting \citep{garrido2019predictive}.
 
\section{Experiments}\label{seq_experiments}
\label{section5:exps}

We evaluate the performance of PPESMOC and
compare results with two base-lines. Namely, a strategy that  chooses 
at each iteration a random batch of points at which to evaluate the 
black-boxes. We also compare results with two parallel sequential methods (see
Section \ref{sec:related_work_ppesmoc}) that use the acquisition function of 
PESMOC and BMOO, respectively. We refer to these methods as PS\_PESMOC and 
PS\_BMOO. In each experiment, we report average results and error bars 
across 100 repetitions. We measure the logarithm of the relative difference 
between the hyper-volume of the problem's solution and the hyper-volume of the recommendation 
provided by each BO method. The solution of the optimization problem is found via exhaustive search in 
synthetic problems. In real-world problems we use the best recommendation obtained 
by any method. In each BO method we use a Mat\'ern covariance function for the GPs. The GP 
hyper-parameters are also sampled from the posterior using $10$ 
slice samples, as in \citep{snoek2012practical}. The predictive distribution 
and acquisition functions are averaged over the samples.
In PPESMOC, the number of samples $S$ of $\mathcal{X}^\star$ is also set to 
$10$ as it was done for PESMOC \citep{garrido2019predictive}. In each method, at each iteration, we 
output a recommendation obtained by optimizing the GP means. For this, we use a grid of points. 
For the optimization of the acquisition function of each method, we select an initial random point at 
random and start L-BFGS from it. Finally, we recommend only feasible solutions with 
high probability as in \citep{garrido2019predictive}. 

\subsection{Synthetic Experiments}

We compare the performance of each method when the objectives and the 
constraints are sampled from a GP prior. The problem considered has 2 
objectives and 2 constraints in a 4-dimensional input space. 
We consider a noiseless and a noisy scenario. In this last case, the 
evaluations are contaminated with Gaussian noise with variance equal to $0.1$.
We report results for different batch sizes. Namely, $4, 8 ,10 $ and $20$ points. 
We allow for $100$ evaluations of each black-box (\emph{i.e.}, a total of 400 evaluations). 
In the case that the recommendation produced contains an infeasible point, 
we simply set the hyper-volume of the recommendation equal to zero. 

\begin{figure}[htb!]
\begin{center}
\begin{tabular}{cc}
\includegraphics[width=0.49\textwidth]{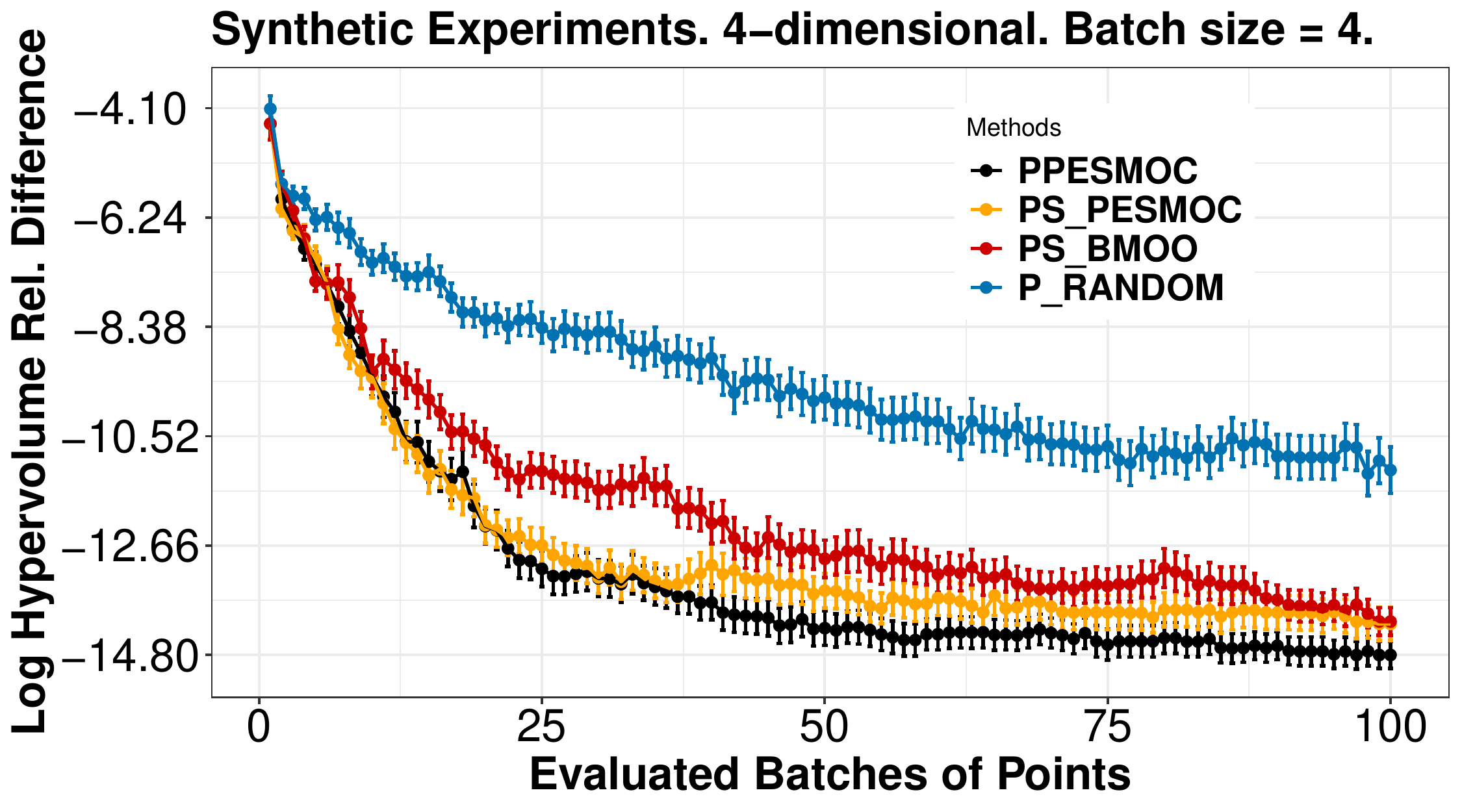} &
\includegraphics[width=0.49\textwidth]{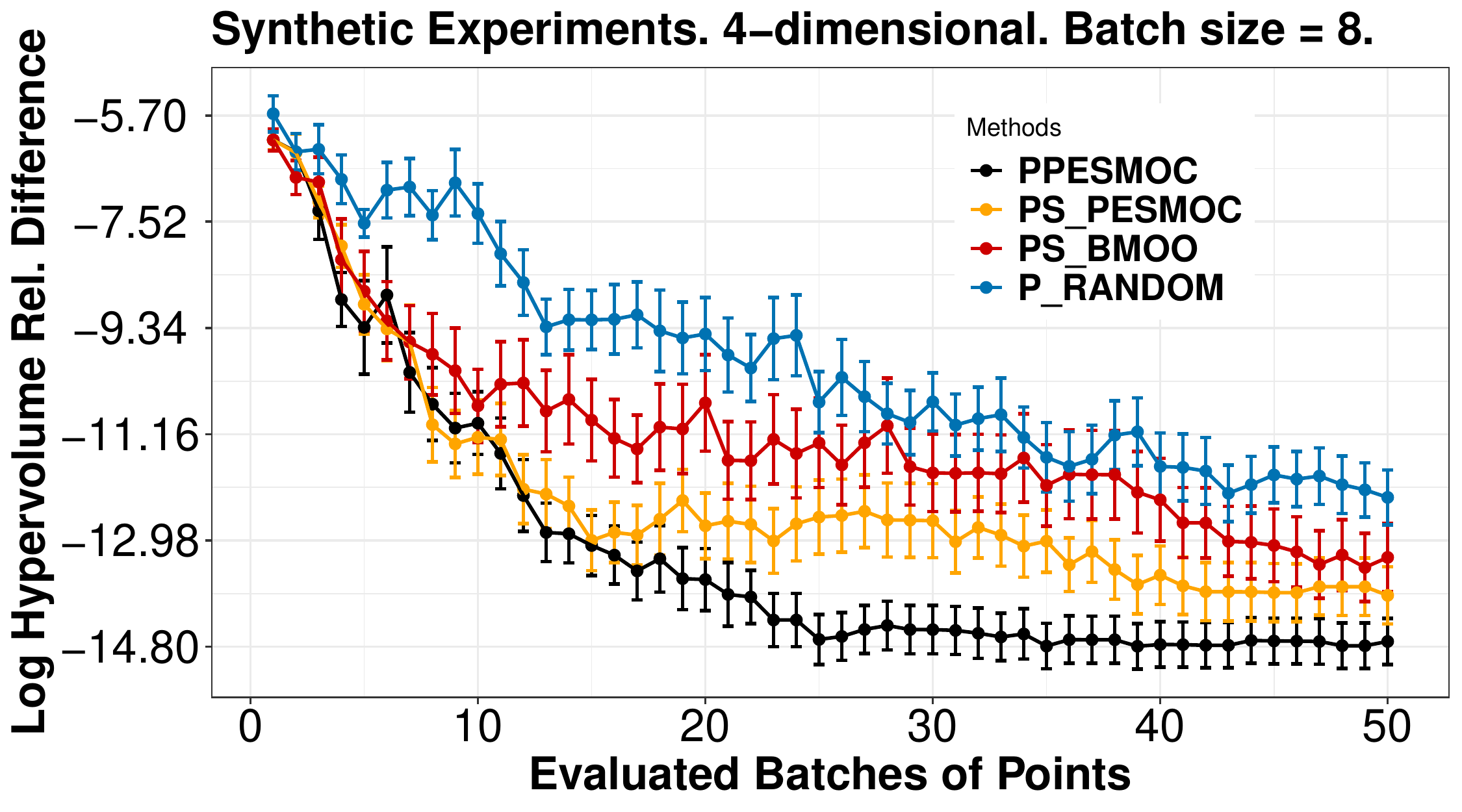}\\
\includegraphics[width=0.49\textwidth]{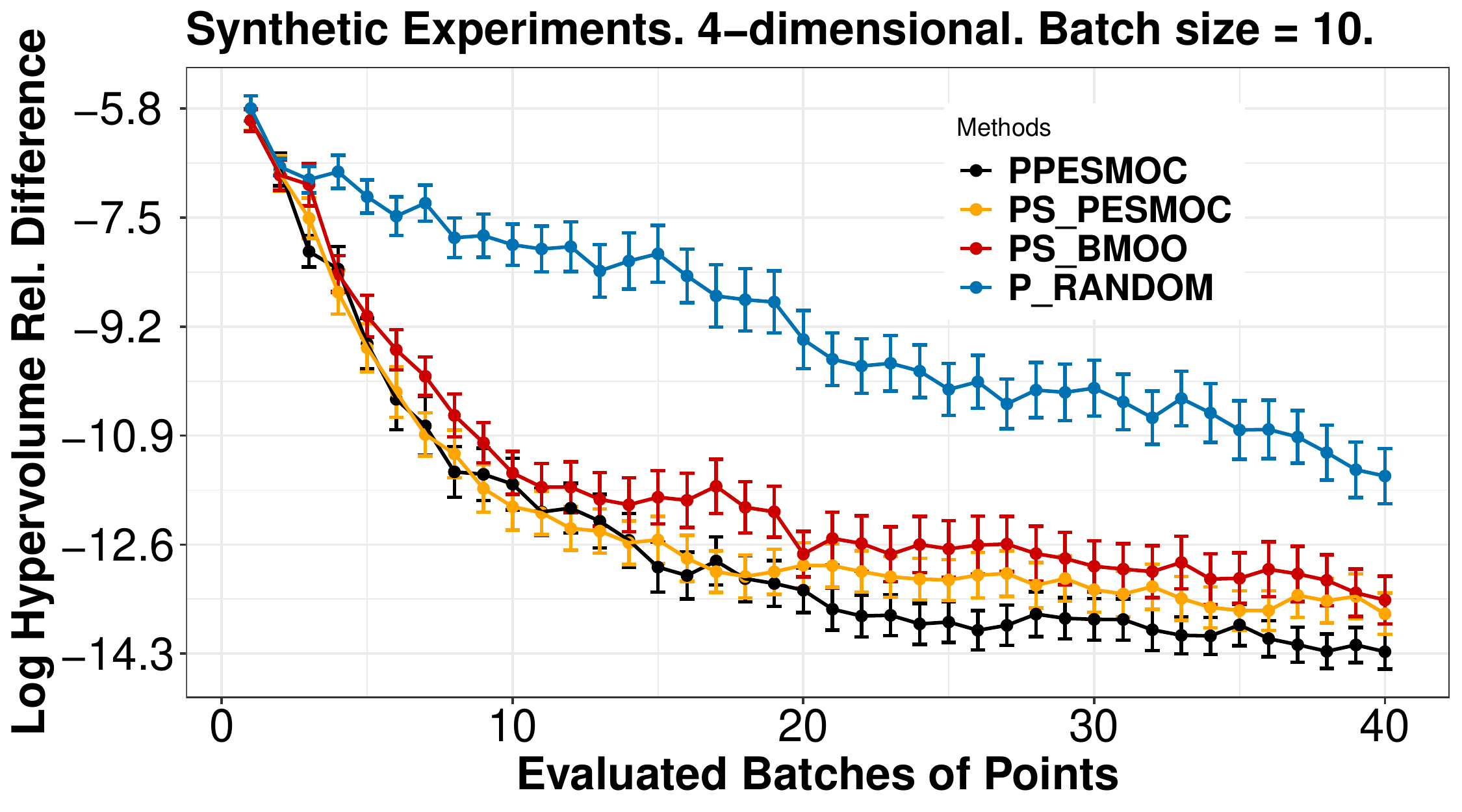} &
\includegraphics[width=0.49\textwidth]{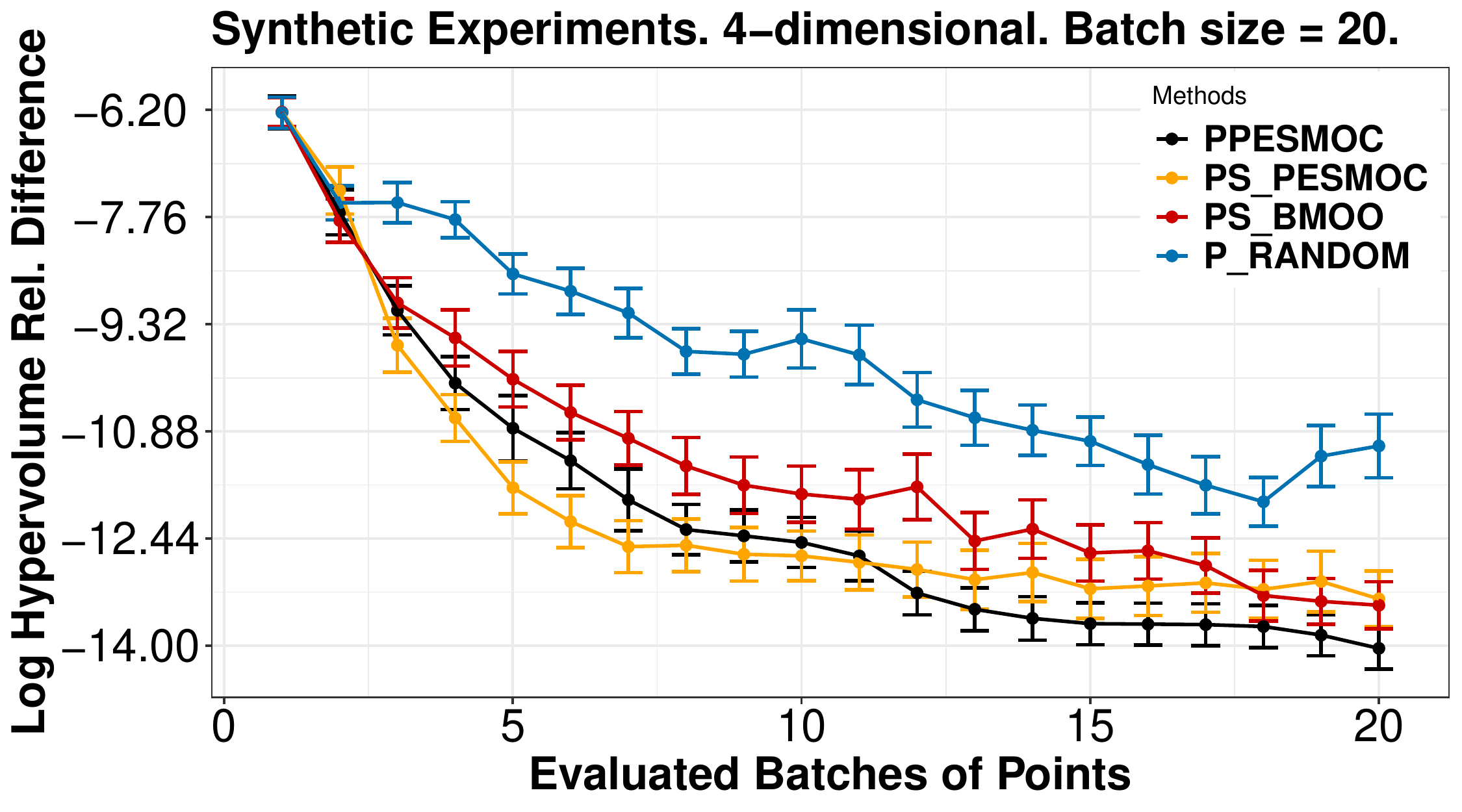}
\end{tabular}
\caption{Average results for the synthetic experiments in the noiseless evaluation setting.}
\label{fig:ppesmoc_synthetic}
\end{center}
\end{figure}

\begin{figure}[htb!]
\begin{center}
\begin{tabular}{cc}
\includegraphics[width=0.49\textwidth]{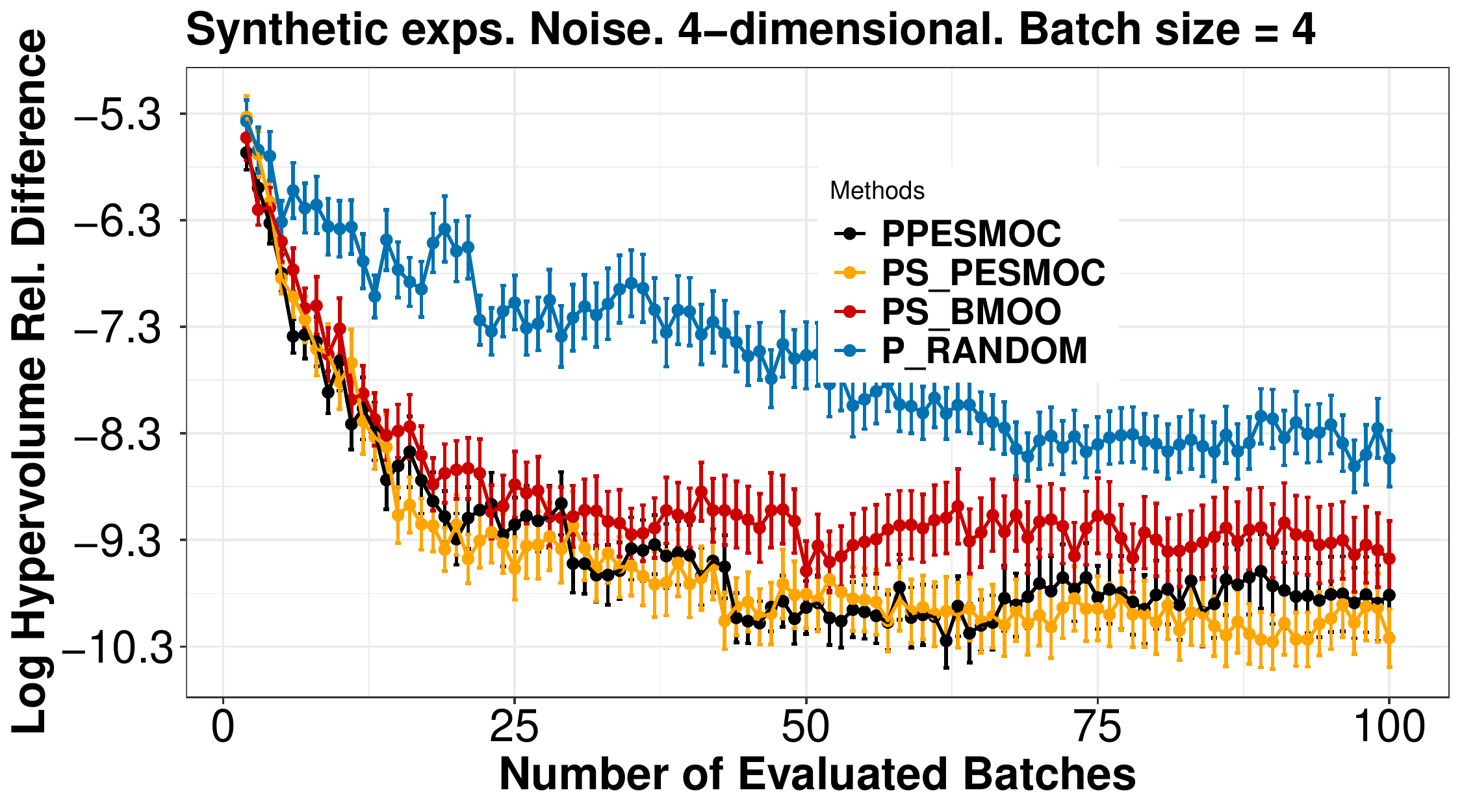} & 
\includegraphics[width=0.49\textwidth]{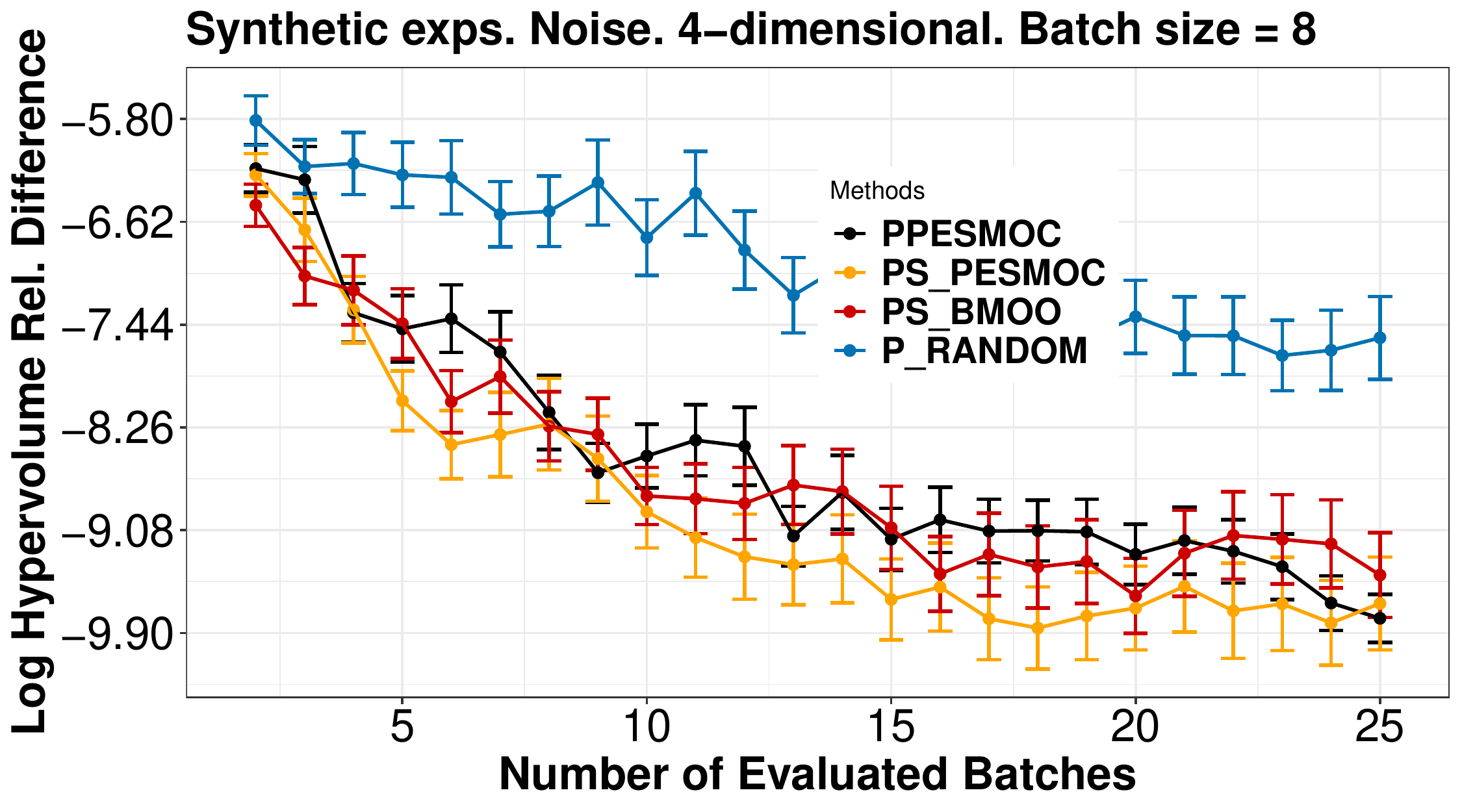} \\
\includegraphics[width=0.49\textwidth]{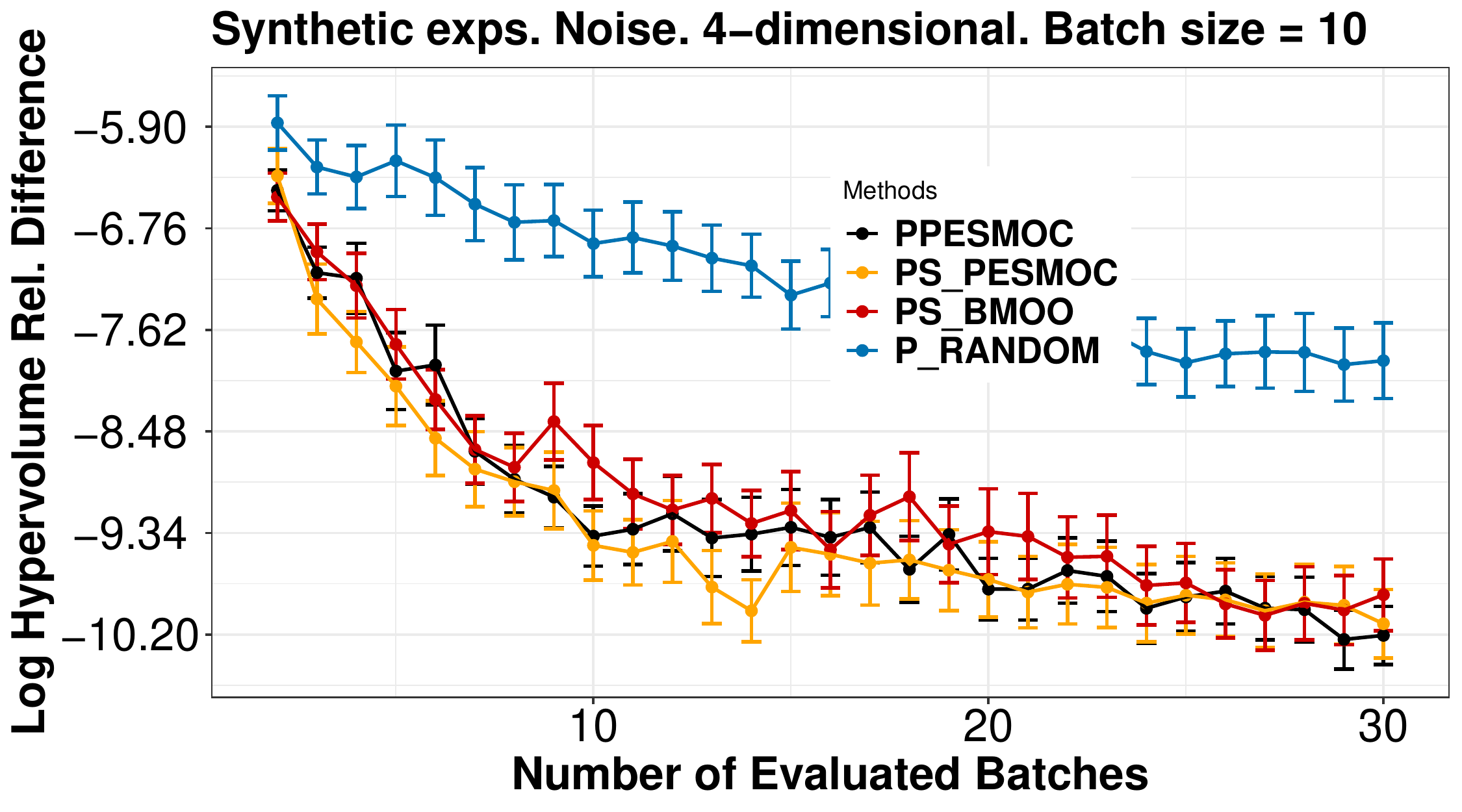} &
\includegraphics[width=0.49\textwidth]{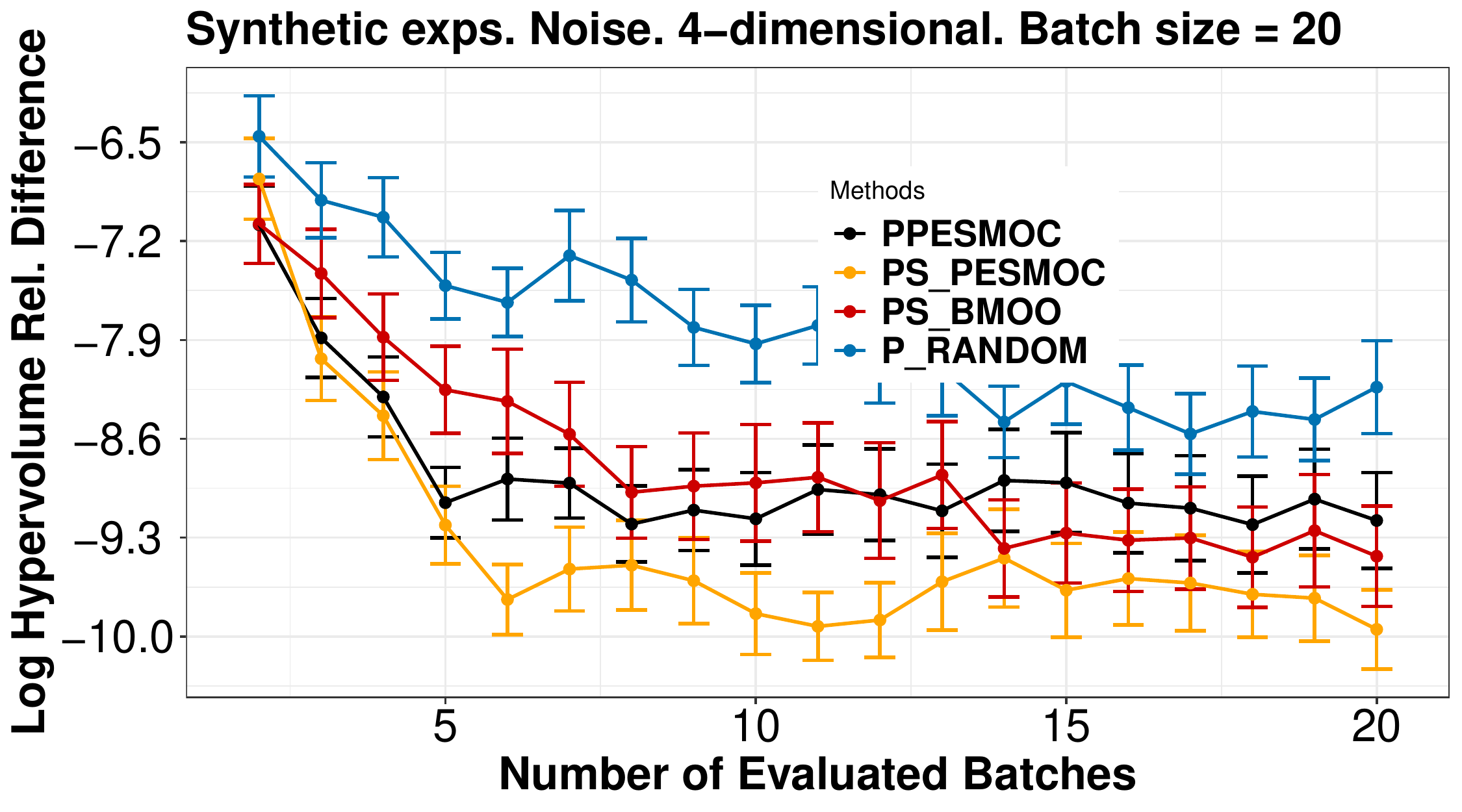}
\end{tabular}
\caption{Average results for the synthetic experiments in the noisy evaluation setting.}
\label{fig:ppesmoc_synthetic_noisy}
\end{center}
\end{figure}

The results obtained are displayed in Figure \ref{fig:ppesmoc_synthetic} and  
Figure \ref{fig:ppesmoc_synthetic_noisy} for the noiseless and noisy case, 
respectively. We observe that PPESMOC performs better than PS\_PESMOC and PPESMOC and the other 
methods in the noiseless setting. The random search strategy gives the worst results followed by PS\_BMOO.
In the noisy setting, however, the differences between the methods are less clear and PPESMOC
and PS\_BMOO and PS\_PESMOC perform similarly. All of them outperform the random search strategy.


The disadvantage of the parallel sequential strategies, PS\_PESMOC and PS\_BMOO, is that they have
a bigger computational cost with respect to the batch size $B$ than PPESMOC. More precisely, they require 
updating the GP models and optimizing the acquisition $B$ times. PPESMOC is hence expected to be computationally 
cheaper for larger batch sizes $B$. We add empirical evidence for this claim by showing, in Table 
\ref{table:pesmoc_times}, the median of the computational time used by each method to determine the next 
batch of points to evaluate. We note that the parallel sequential strategies are faster for smaller
batches of points. This is because the acquisition function is simpler. However, for larger batch sizes, 
\emph{i.e.}, $B=50$, PPESMOC has a computational time that scales better than the parallel sequential approaches. 
By contrast, PS\_PESMOC requires, on average, more and more computation time as the batch size $B$ increases.

\begin{table}[htb]
\centering
\caption{Median of the time in seconds to choose the next batch of points by PPESMOC and the parallel sequential 
approaches.}
\label{table:pesmoc_times}
\begin{tabular}{lr@{$\pm$}lr@{$\pm$}lr@{$\pm$}lr@{$\pm$}lr@{$\pm$}l}
\hline
\bf{Method} & \multicolumn{2}{c}{B=4} & \multicolumn{2}{c}{B=8} & \multicolumn{2}{c}{B=10} & \multicolumn{2}{c}{B=20} & \multicolumn{2}{c}{B=50}\\
\hline
PPESMOC & 697.3 & 27.5 & 913.9 & 28.1 & 960.7 & 27.8 & 1044.8 & 30.7 & \bf{1273.9} & \bf{30.1} \\
PS\_PESMOC & \bf{191.3} & \bf{7.0} & \bf{346.2} & \bf{6.1} & \bf{406.2} & \bf{6.5} & \bf{799.8} & \bf{26.6} & 1960.3 & 31.6 \\
PS\_BMOO & 379.4 & 12.3 & 551.7 & 22.8 & 594.2 & 19.9 & 895.9 & 27.9 & 1874.2 & 42.4 \\
\hline
\end{tabular}
\end{table}



\subsection{Benchmark Experiments}

\begin{figure}[ht]
\begin{center}
\includegraphics[width=0.49\textwidth]{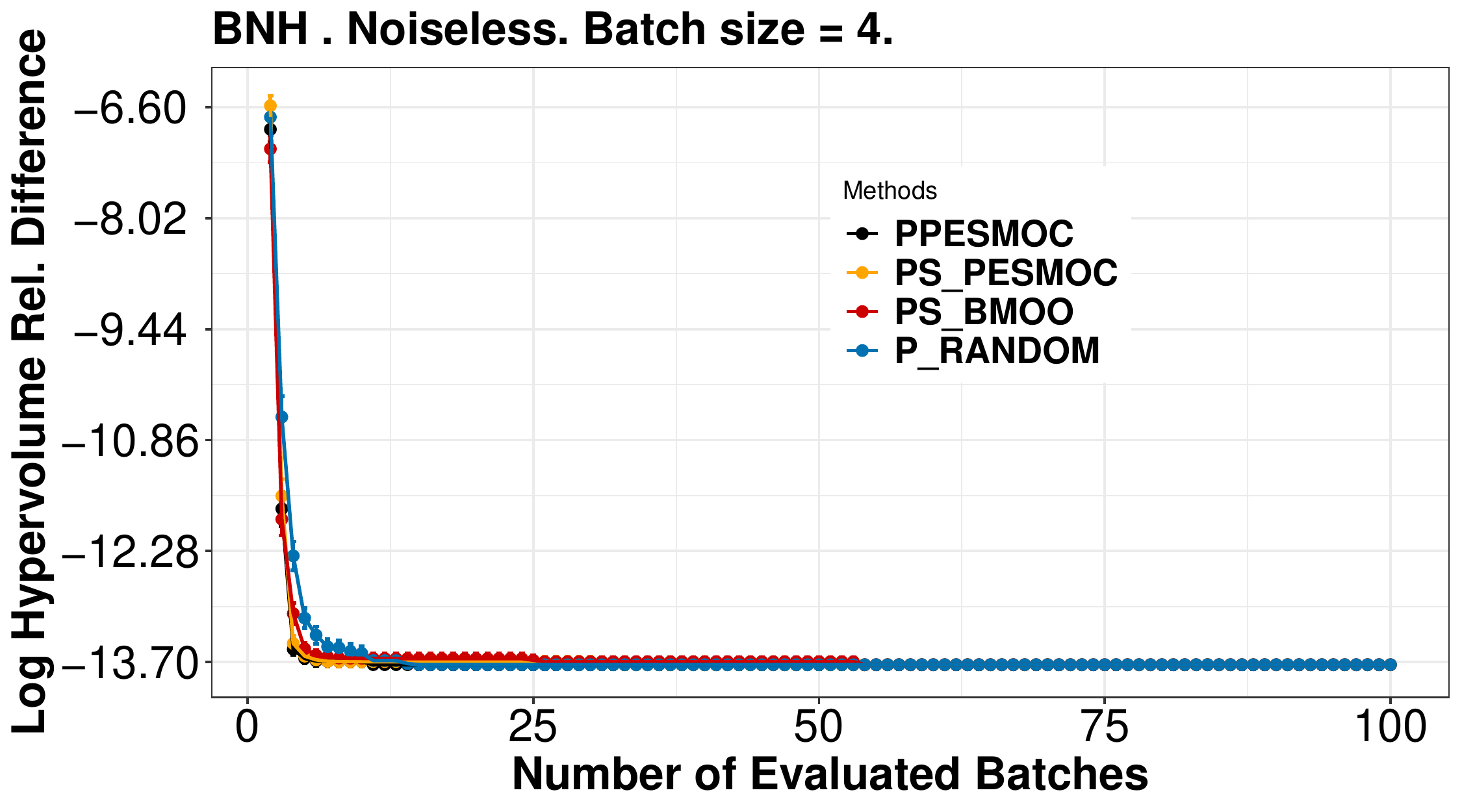}
\includegraphics[width=0.49\textwidth]{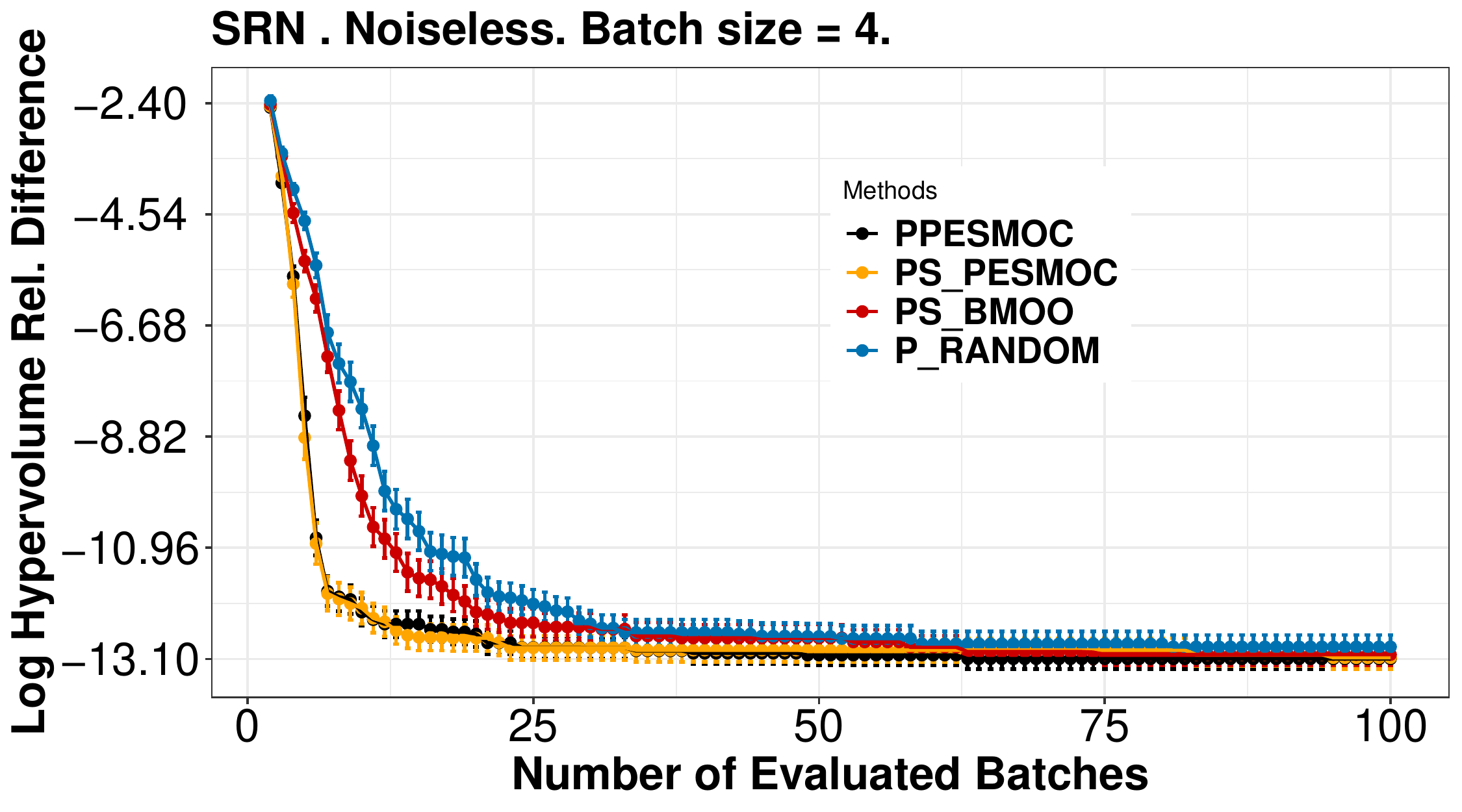} \\
\includegraphics[width=0.49\textwidth]{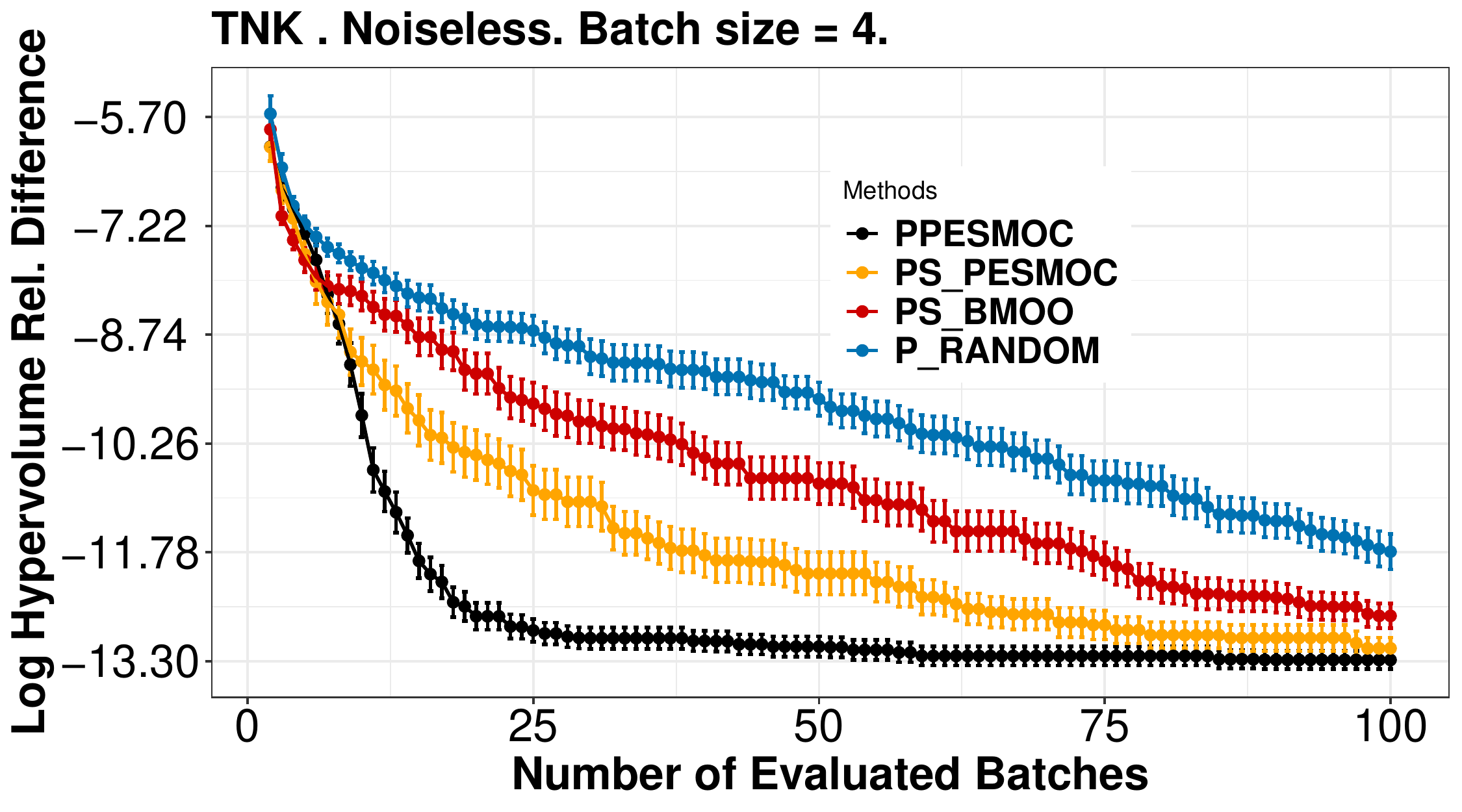}
\includegraphics[width=0.49\textwidth]{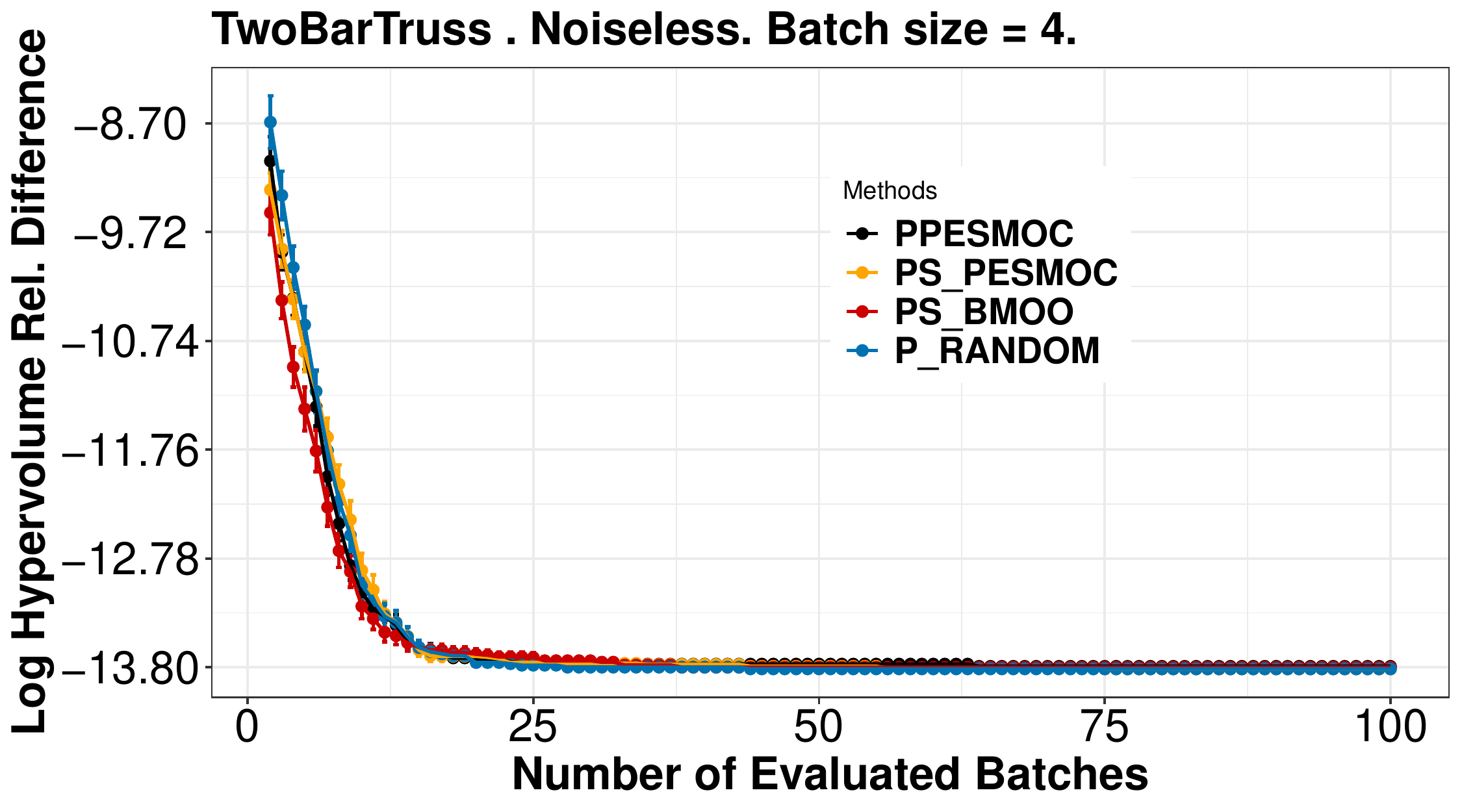} \\
\includegraphics[width=0.49\textwidth]{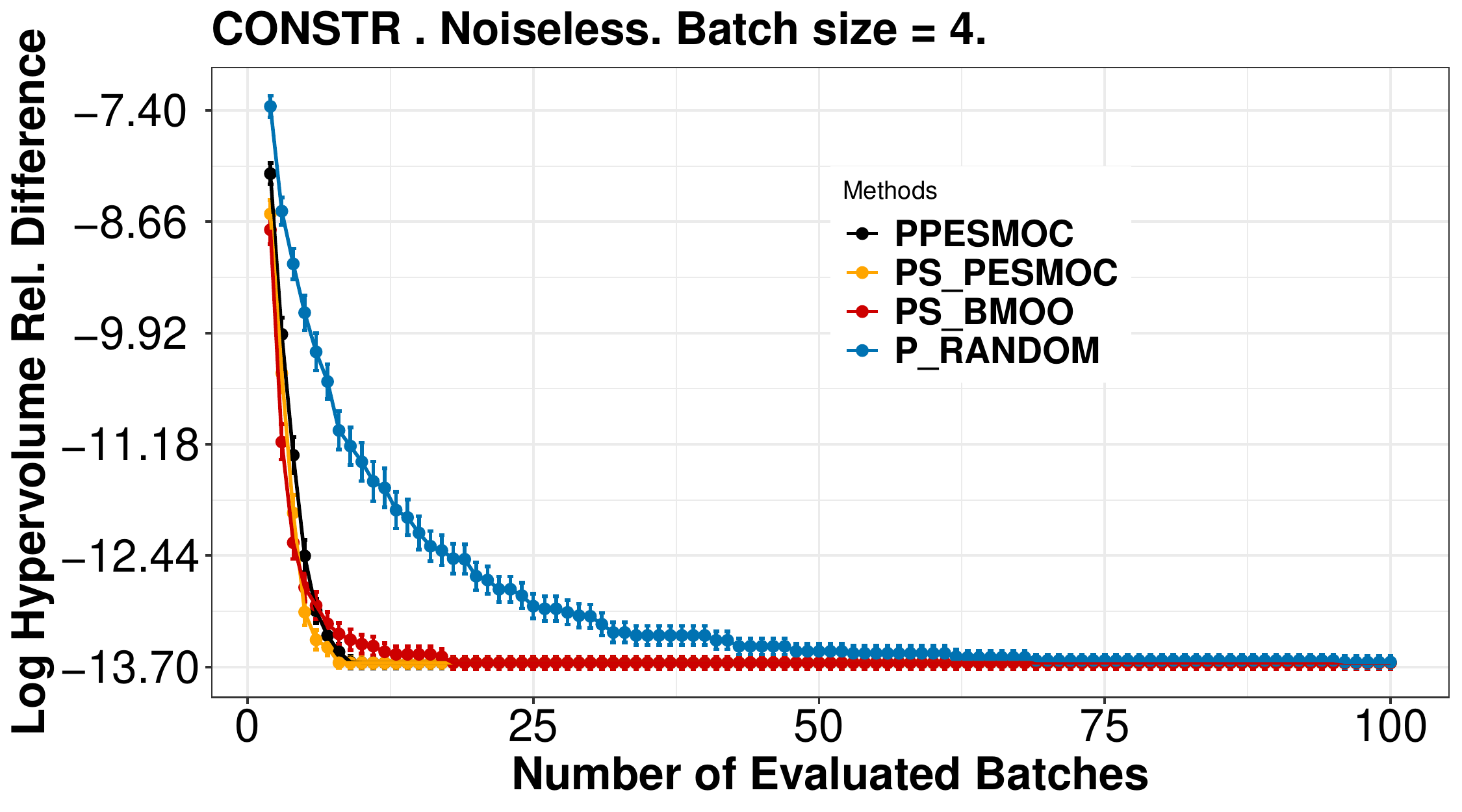}
\includegraphics[width=0.49\textwidth]{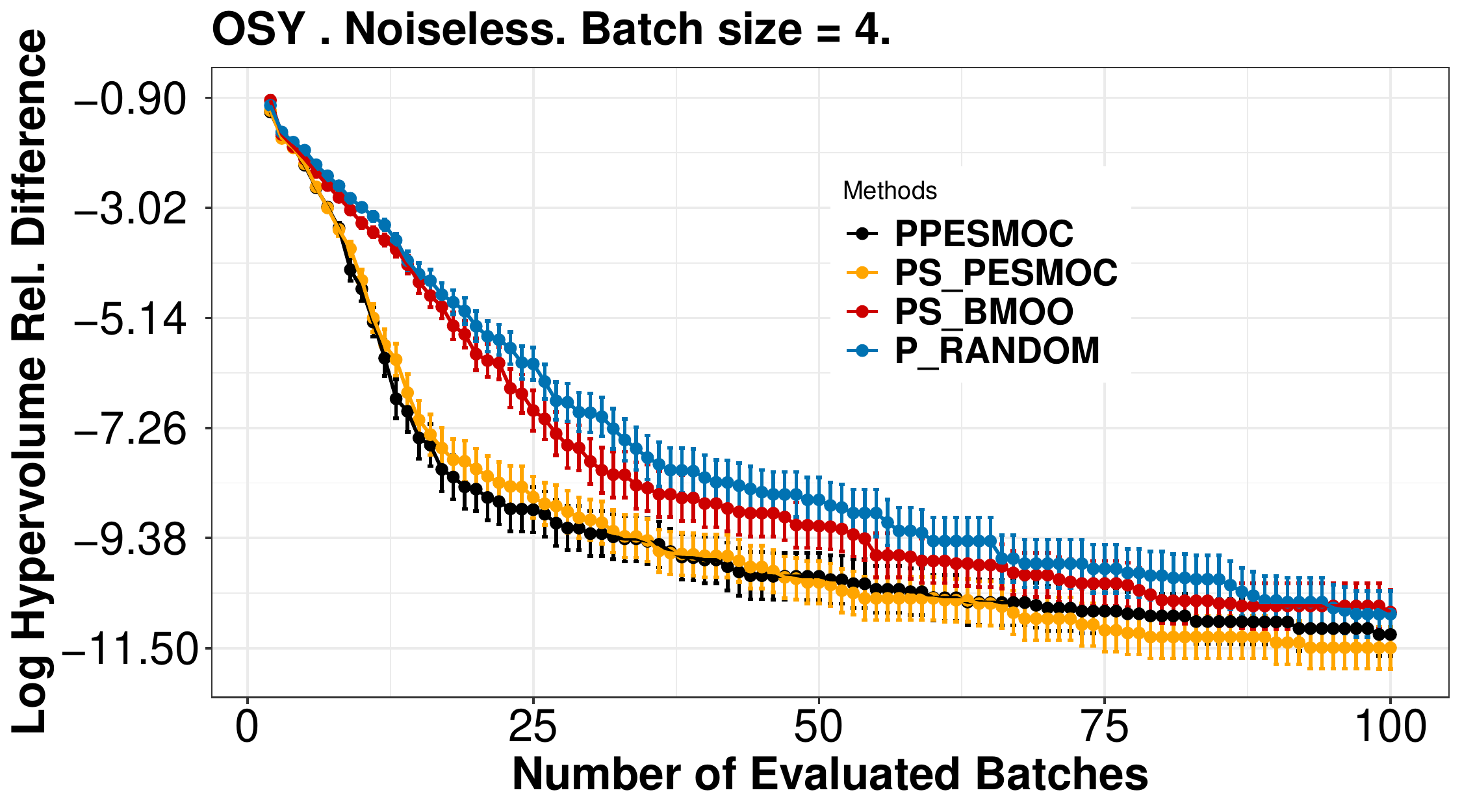} \\
\caption{Logarithm of the relative difference between the hyper-volume of the recommendation obtained by each method and the hyper-volume of the actual solution. We report results after each evaluation of the black-box functions. Benchmark functions corrupted by noise.}
\label{fig:bench}
\end{center}
\end{figure}

\begin{figure}[ht]
        \begin{tabular}{cc}
                \includegraphics[width=0.475\linewidth]{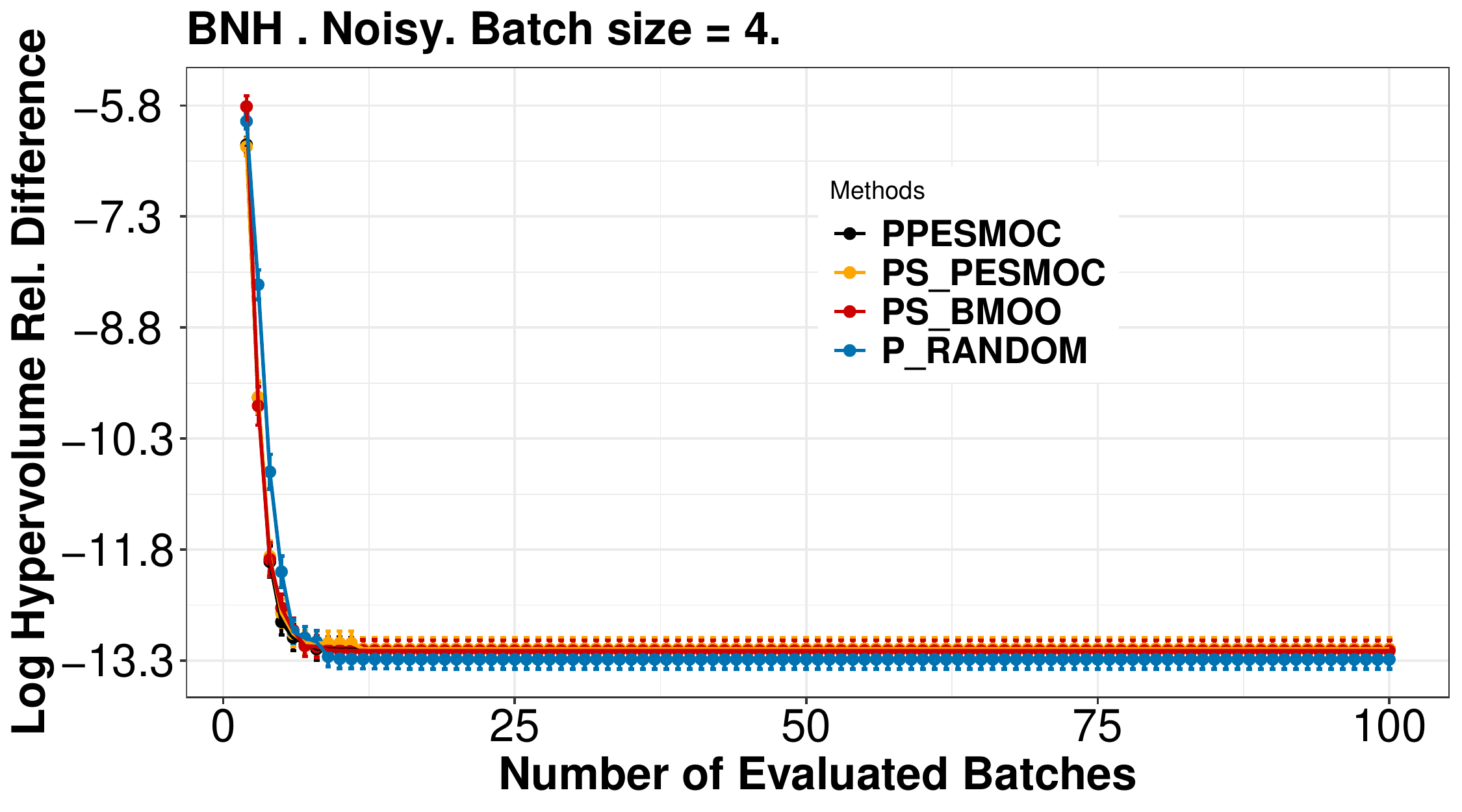} &
                \includegraphics[width=0.475\linewidth]{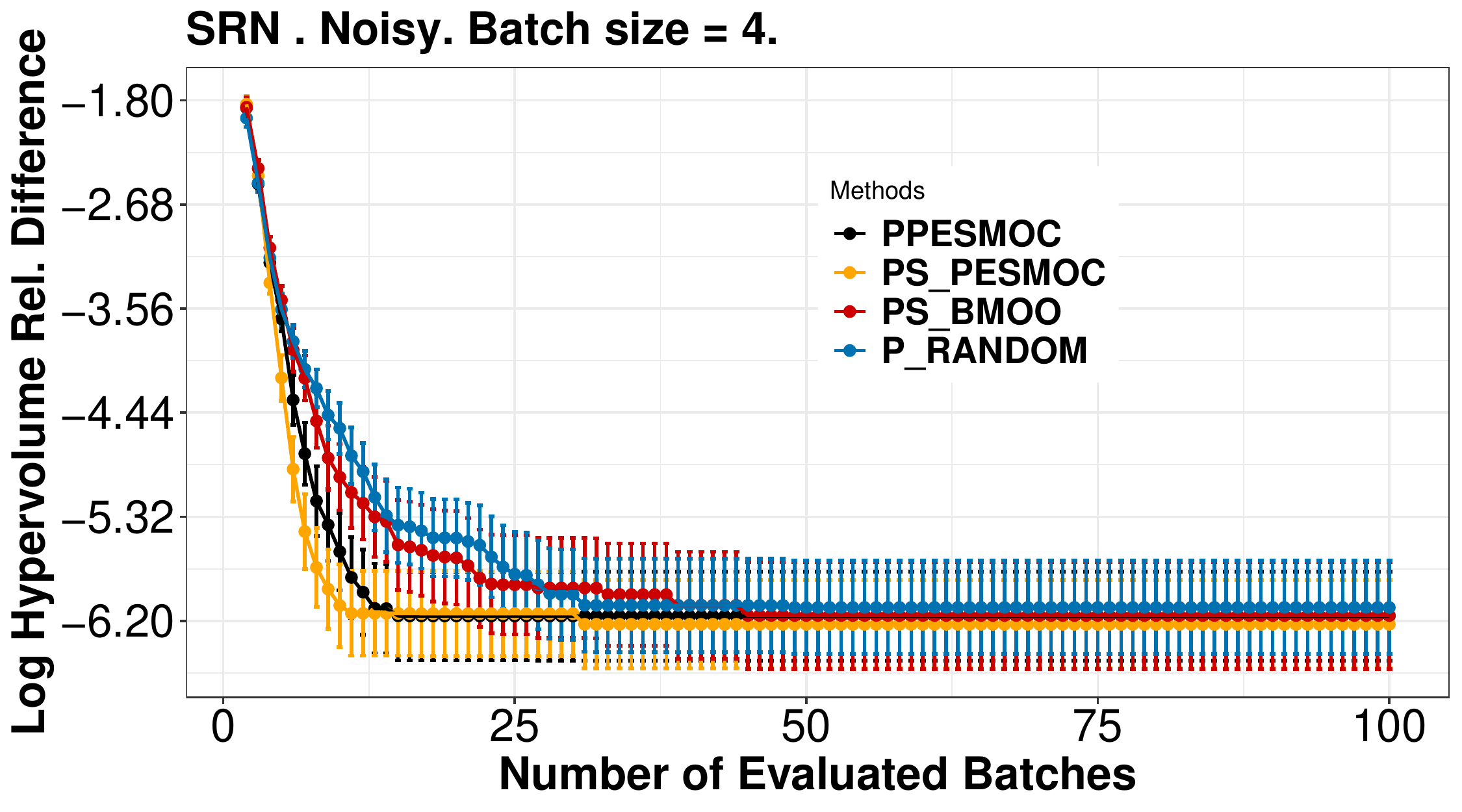} \\
                \includegraphics[width=0.475\linewidth]{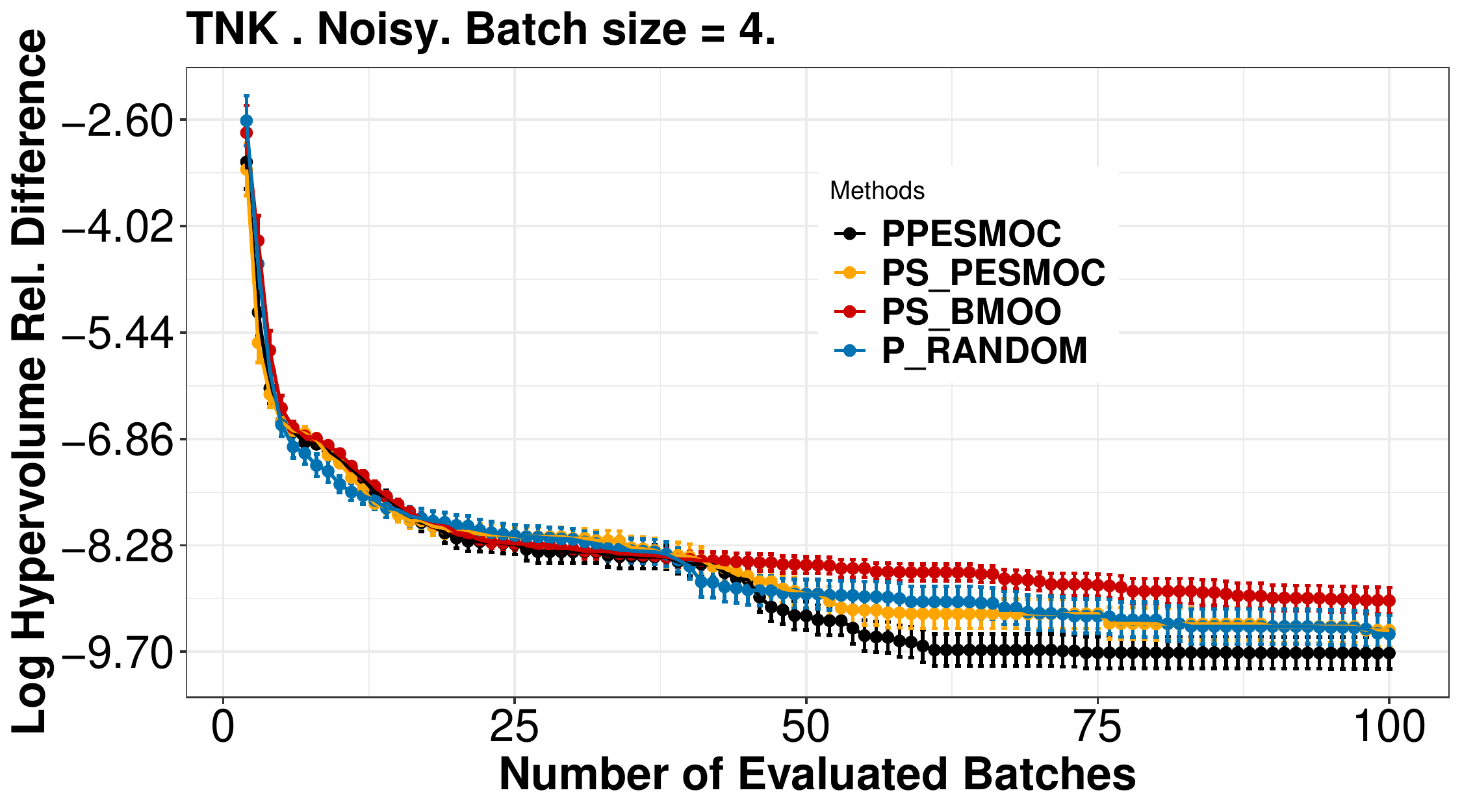} &
                \includegraphics[width=0.475\linewidth]{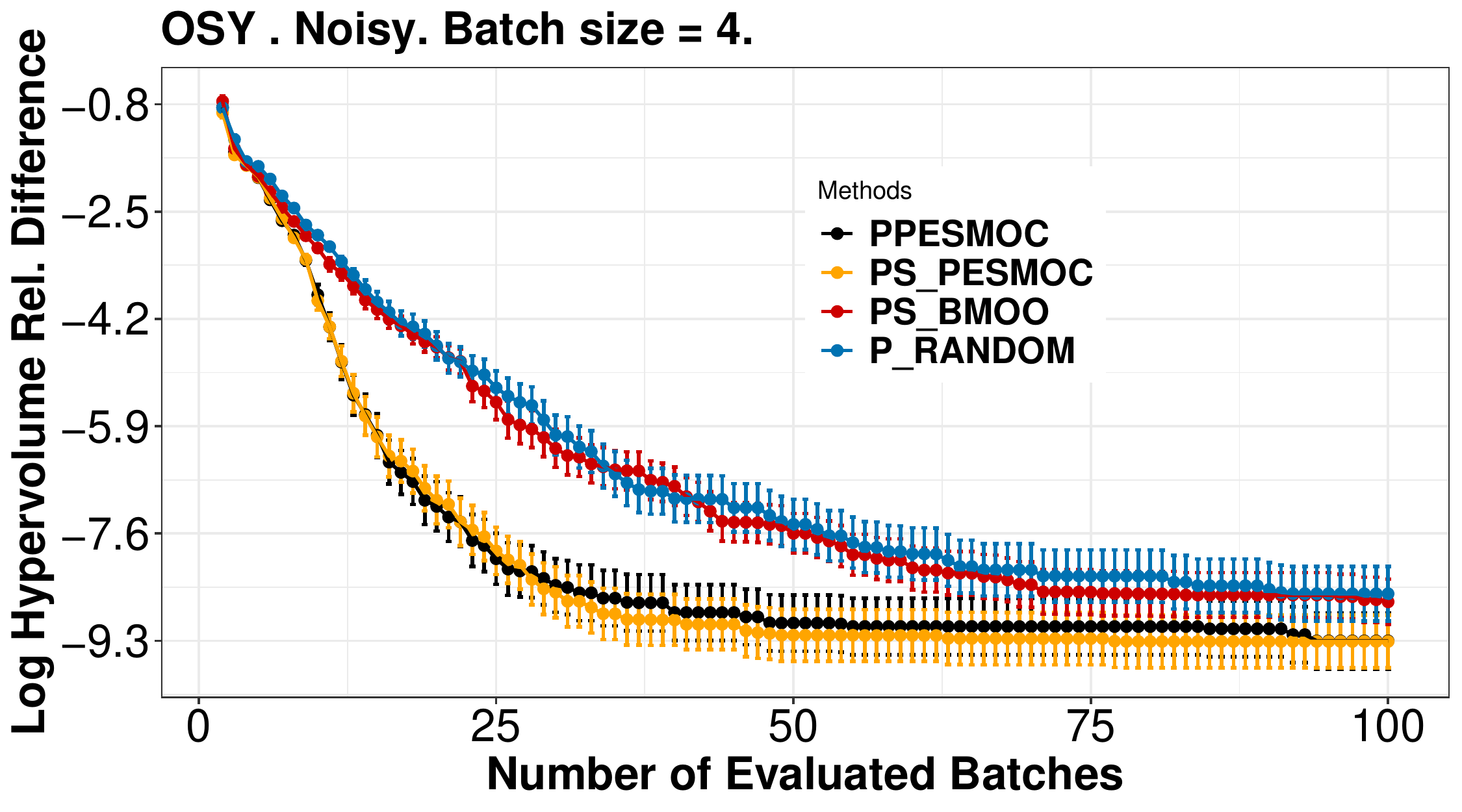} \\
                \includegraphics[width=0.475\linewidth]{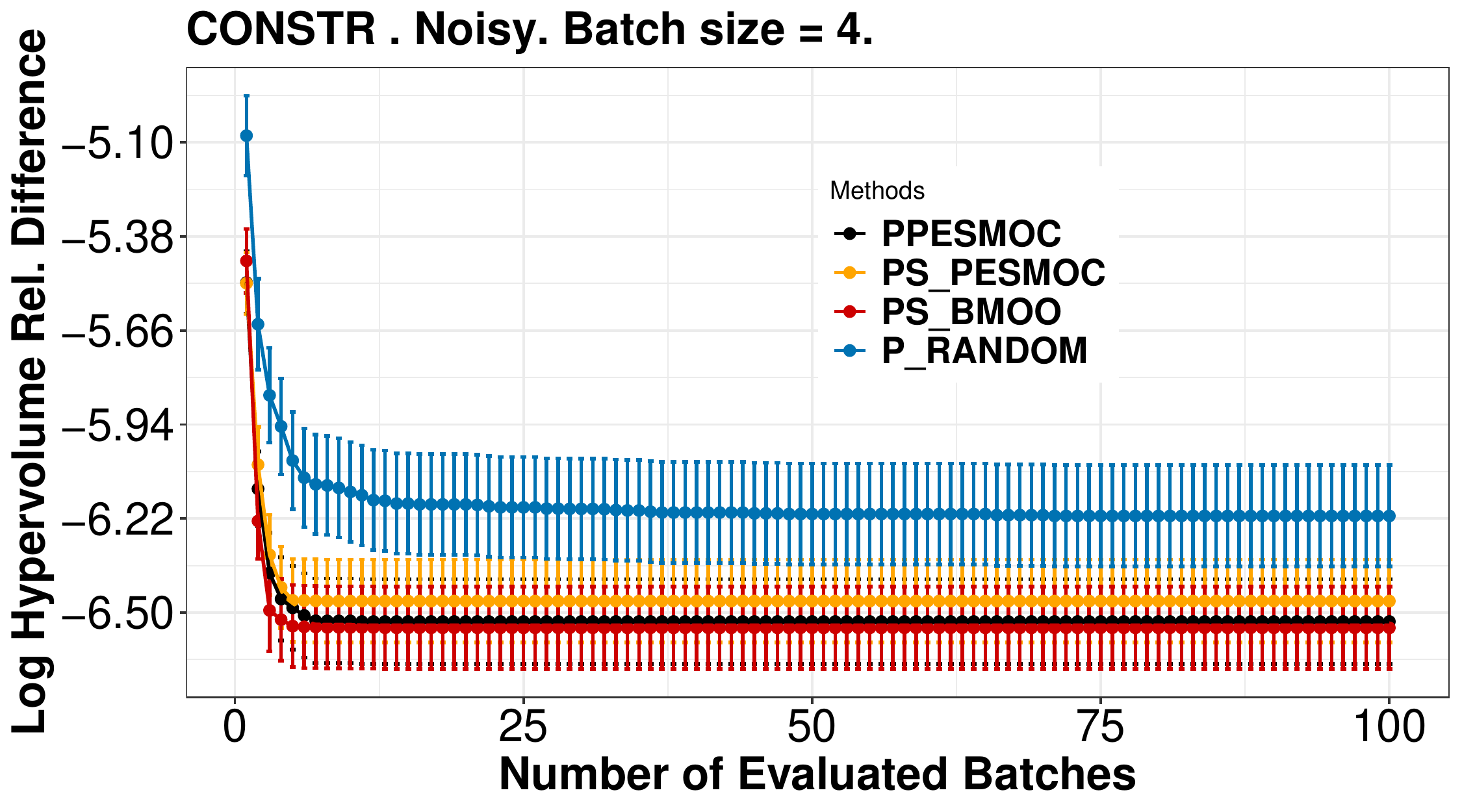} &
                \includegraphics[width=0.475\linewidth]{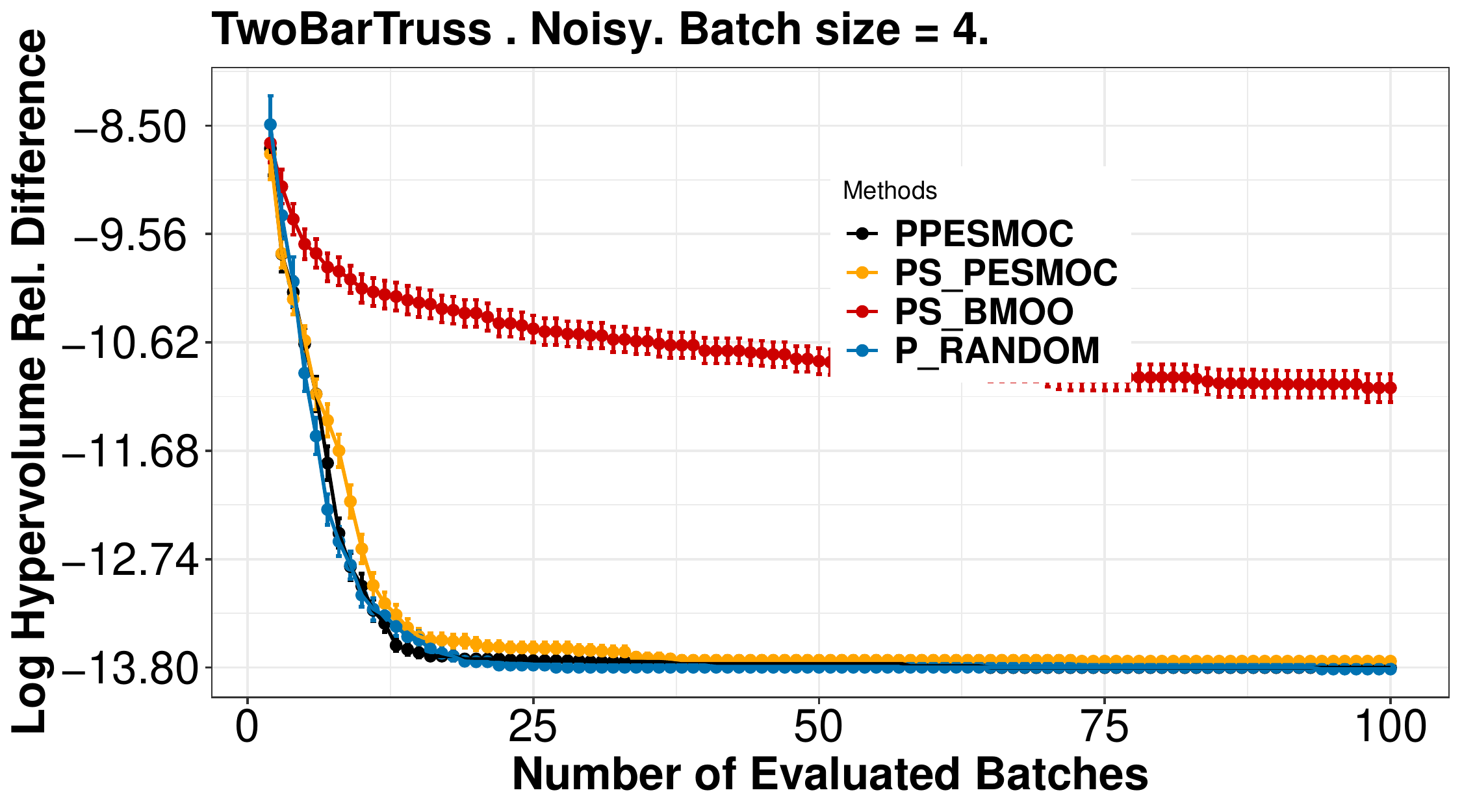} 
        \end{tabular}
        \caption{{Average results for the problems BNH, SRN, TNK and OSY, CONSTR and TwoBar Truss. Noisy setting. }}
        \label{fig:benchmark_1}
\end{figure}

We carry out extra experiments in which the black-boxes are not sampled from a 
GP. For this, we consider 6 classical constrained multi-objective optimization 
problems \citep{chafekar2003constrained}. The analytical expression for 
the objectives and constraints of each optimization problem are described in the 
supplementary material.  We consider two scenarios. A noiseless scenario and a
noisy scenario, where the black-box evaluations are contaminated with 
additive Gaussian noise. The variance of the noise is set to 1\% of
the range of potential values of the corresponding black-box.
The batch size considered is $B=4$.
The average results obtained in these experiments, for each method and each scenario, 
are displayed in Figure \ref{fig:bench}, that shows plots of the noiseless case, 
and Figure \ref{fig:benchmark_1}, showing the noisy setting. We observe that, most of 
the times, PPESMOC perform similar or better than the other methods, in each scenario, 
noiseless and noisy, as it was expected. 

\subsection{Real-world Experiments}

We compare each method in the task of finding an optimal 
gradient-boosting ensemble, as in \citep{garrido2019predictive}. We consider two objectives: 
the prediction error of the ensemble and its size. These objectives are conflictive since smaller ensembles 
will have in general higher error rates and the other way around. 
We introduce as a constraint of the problem, 
that the average speed up factor of the classification process given by a dynamic 
ensemble pruning technique  is at least 25$\%$ \citep{hernandez2009statistical}. We 
have carefully chosen this value to guarantee that the constraint is active at 
the optimal solution. The dataset considered is the German credit dataset extracted 
from the UCI repository \citep{Dua2019}. This is a binary classification dataset with $1,000$ 
instances and $20$ attributes. The prediction error is measured using 10-fold-cross validation, 
repeated 5 times. The ensemble size is the logarithm of the 
sum of the total number of nodes in the trees of the ensemble.

The adjustable parameters of the ensemble are: the number of trees 
(between 1 and $1,000$), the number of random attributes considered for split in 
each tree (between 1 and 20), the minimum number of samples required to split a node (between 2 and
200), the fraction of randomly selected training data used to build each tree (between 0.5 and 1.0),
and the fraction of training instances whose labels are changed after the sub-sampling process
(between $0.0$ and $0.7$).

\begin{figure}[ht]
\begin{center}
\begin{tabular}{cc}
\includegraphics[width=0.49\textwidth]{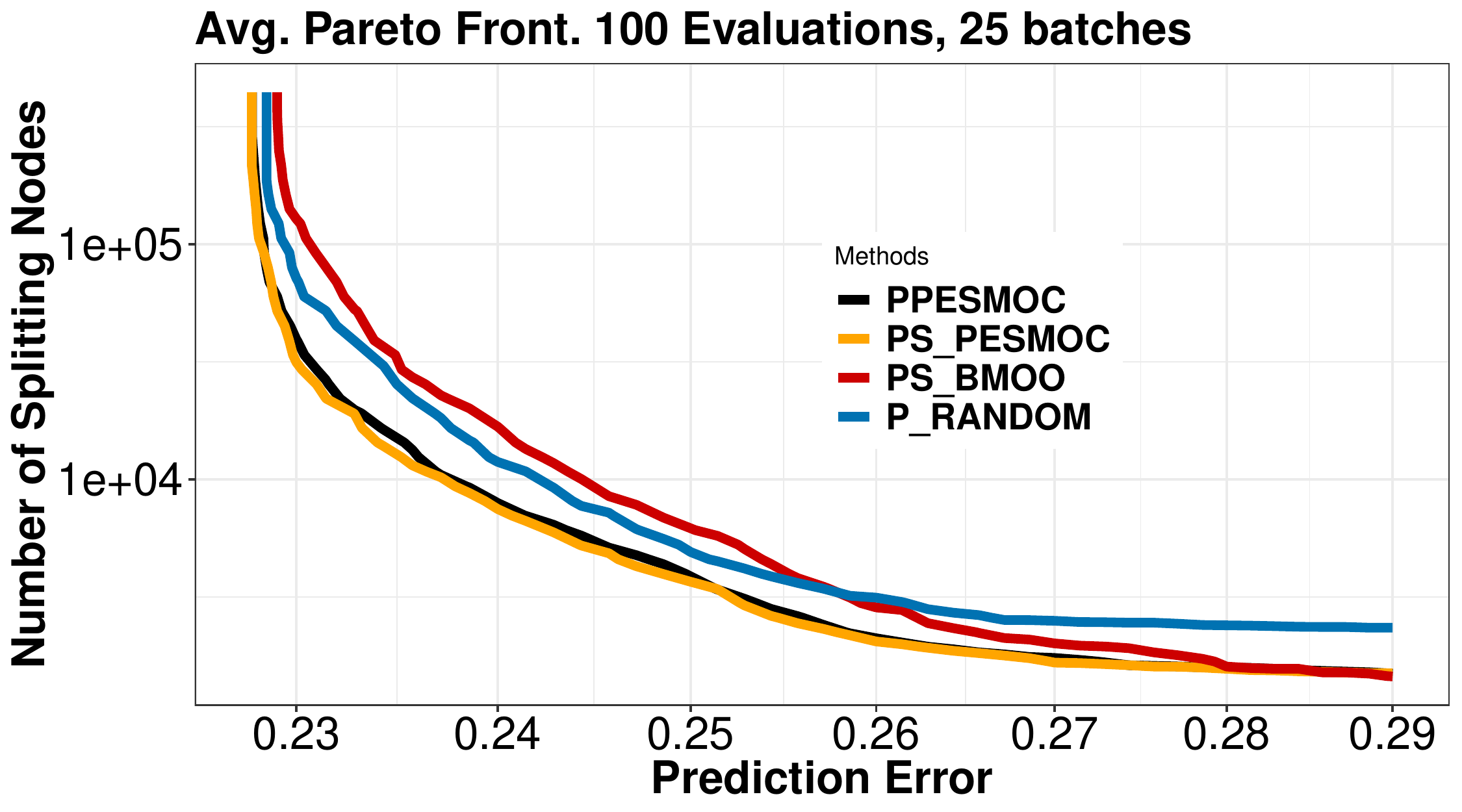} & \includegraphics[width=0.49\textwidth]{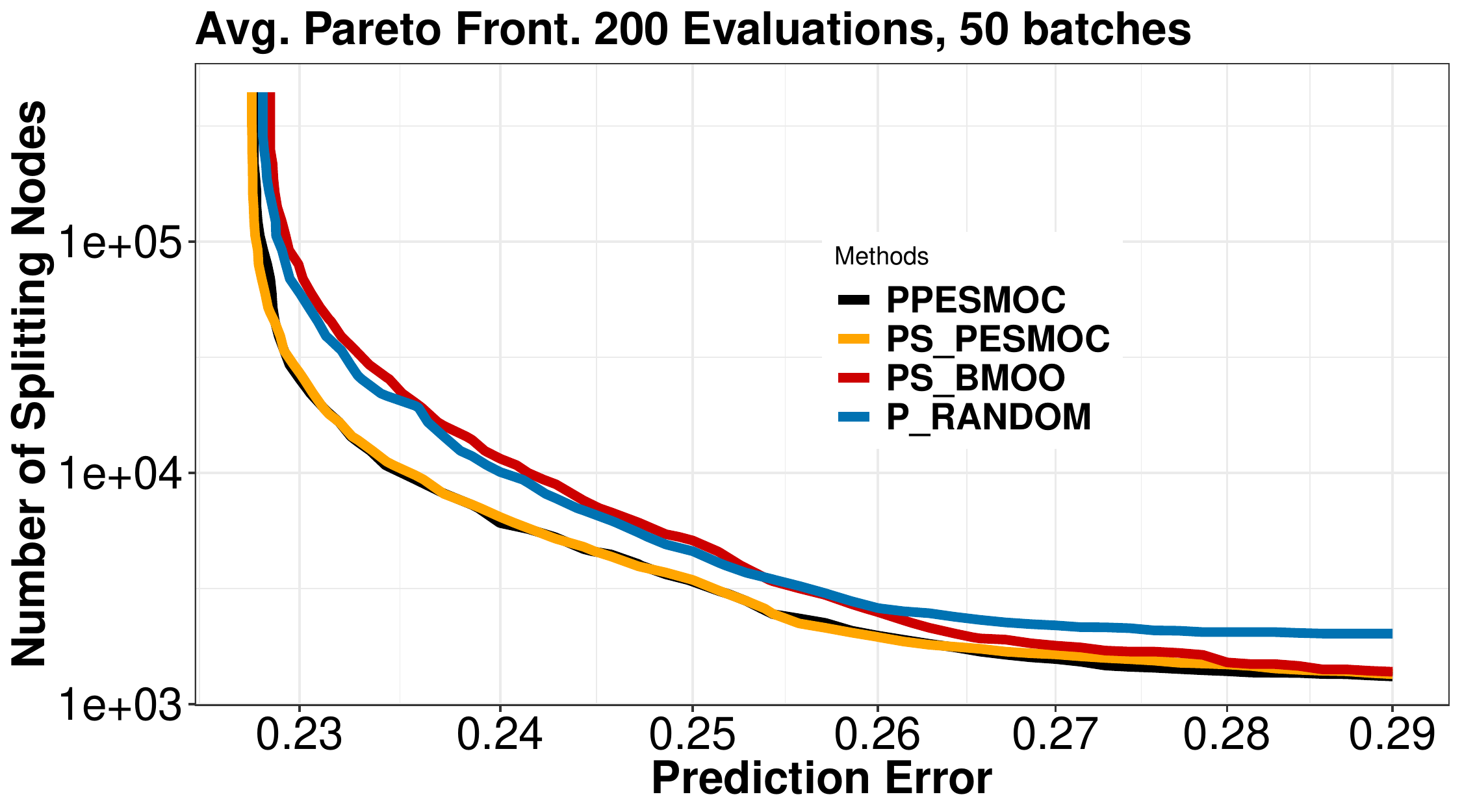} \\
\end{tabular}
\caption{Average Pareto front in the task of finding an optimal ensemble for 100 (left) and 200 evaluations (right).
}
\label{fig:ensemble}
\end{center}
\end{figure}

Table \ref{table:ensemble_hypervolumes} shows the average hyper-volume obtained in this task, for each method, 
after $100$ and $200$ evaluations using a batch size of $4$. Figure \ref{fig:ensemble} shows also the
average Pareto front obtained by each method. The Pareto front is simply given by the objective values 
associated to the recommendation made by each method. The higher the volume of points that is above this set
of points in the objective values the better the performance of a method. We observe that 
for $100$ evaluations PS\_PESMOC is the best strategy closely followed by PPESMOC. However, after $200$ evaluations
PPESMOC performs better. The random strategy is the worst method and PS\_BMOO and performs slightly worse than PS\_PESMOC.

\begin{table}[htb]
\centering
\caption{Average hyper-volume in the task of finding an optimal ensemble of trees.}
\begin{tabular}{cr@{$\pm$}l@{\hspace{.5cm}}r@{$\pm$}l@{\hspace{.5cm}}r@{$\pm$}l@{\hspace{.5cm}}r@{$\pm$}l}
\hline
{\bf \# Eval.} & \multicolumn{2}{c}{\bf PPESMOC} & \multicolumn{2}{c}{\bf PS\_PESMOC} & \multicolumn{2}{c}{\bf PS\_BMOO} & 
	\multicolumn{2}{c}{\bf P\_RANDOM} \\
\hline
100 & \hspace{.5cm}0.284 & 0.016 & \hspace{.5cm}{\bf 0.287} & {\bf 0.017} & \hspace{.5cm} 0.263 & 0.020 & \hspace{.5cm}0.247 & 0.017 \\
200 & \hspace{.5cm}\bf{0.316} & \bf{0.006} & \hspace{.5cm}0.315 & 0.008 & \hspace{.5cm} 0.298 & 0.011 & \hspace{.5cm}0.278 & 0.010 \\
\hline
\end{tabular}
\label{table:ensemble_hypervolumes}
\end{table}

As described in PESMOC,
we also evaluate each method on the task of finding an optimal deep neural network (DNN) on the 
MNIST dataset \citep{lecun2010mnist}. The objectives 
are the prediction error of the DNN on a validation 
dataset of $10,000$ instances (extracted from the original training set) and the time that such a DNN
will take for making predictions (normalized with respect to the prediction time of the fastest possible network). 
These are conflictive objectives in the sense that 
minimizing the prediction error will often lead to bigger DNN with a bigger prediction times. 
We are also interested in codifying such a DNN into a chip.
Thus, we constrain the problem by enforcing that the area of the
resulting DNN, after being codified into a chip, is below $1$ $\text{mm}^2$.
We have carefully chosen this value to guarantee that the constraint is active at the optimal solution.
To measure the chip area we use the Aladdin simulator, which 
given a computer program describing the operations of the DNN, 
outputs an estimate of the area of a chip implementing those operations \citep{shao2014aladdin}.
To train the DNN we use the Keras library.
Prediction time is normalized by the smallest possible prediction time, which 
corresponds to a DNN of a single layer with $5$ hidden units.

The input parameters to be optimized are:
The logarithm of the $\ell_1$  and $\ell_2$ weight regularizers;
the dropout probability; the logarithm of the initial learning
rate; the number of hidden units per layer; and the number of hidden layers. We have 
also considered two variables that have an impact in the hardware implementation
of the DNN. Namely, the logarithm (in base 2) of the array partition 
factor and the loop unrolling factor.
 
We report the performance after $60$ evaluations using a batch 
size $B=4$. The DNN is trained  using ADAM with the default parameters. 
The loss function is the  cross-entropy. The last layer of 
the DNN contains 10 units and a soft-max activation function. All other layers use
Re-Lu as the activation function. Finally, each DNN is trained during a total of
$150$ epochs using mini-batches of size $4,000$.

\begin{figure}[ht]
\begin{center}
\begin{tabular}{cc}
\includegraphics[width=0.85\textwidth]{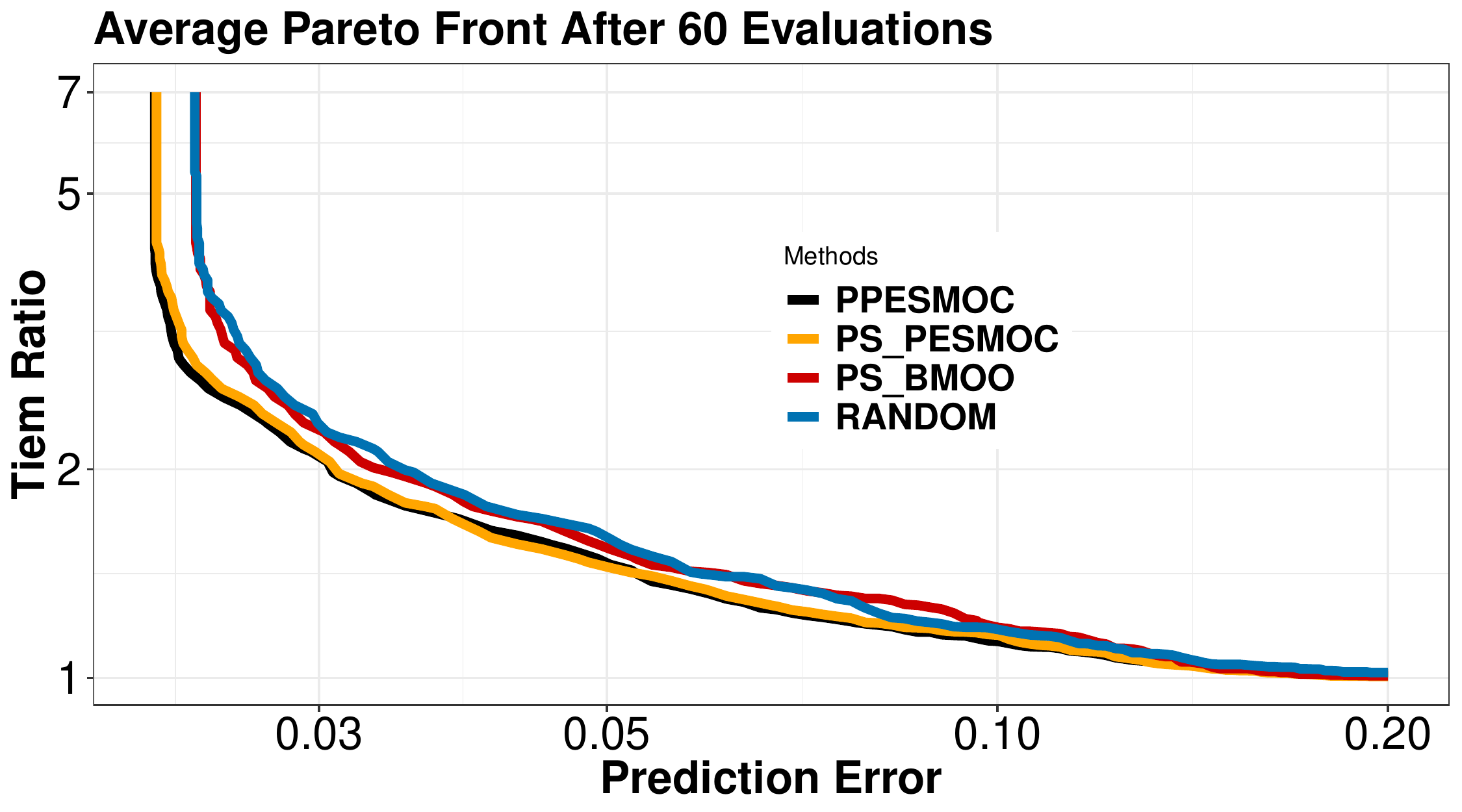} 
\end{tabular}
\caption{Average Pareto front in the task of finding an optimal neural network for 60 evaluations.
}
\label{fig:rrnn_op}
\end{center}
\end{figure}

The average Pareto front obtained by each method is shown in
Figure \ref{fig:rrnn_op}. Time ratio corresponds to the ratio between 
the time needed for making predictions normalized by
the time of the fastest possible network.
Table \ref{table:aladdin_hypervolume} shows the 
average hyper-volume of each method. Here, PPESMOC outperforms by little 
PS\_PESMOC. PS\_BMOO gives worse results than PS\_PESMOC and performs 
almost equal to the random search strategy. We can see in the frontier how 
the difference between PPESMOC and PS\_PESMOC method is fairly small, however
PPESMOC is able to find neural networks with better prediction error for the same
prediction time.

\begin{table}[htb]
\centering
\caption{Avg. hyper-volume of each method in the neural network experiment. }
\label{table:aladdin_hypervolume}
\begin{tabular}{cr@{$\pm$}l@{\hspace{.5cm}}r@{$\pm$}l@{\hspace{.5cm}}r@{$\pm$}l@{\hspace{.5cm}}r@{$\pm$}l}
\hline
{\bf \# Eval.} & 
\multicolumn{2}{c}{\bf PPESMOC} & \multicolumn{2}{c}{\bf PS\_PESMOC} & \multicolumn{2}{c}{\bf PS\_BMOO} & 
	\multicolumn{2}{c}{\bf P\_RANDOM} \\
\hline
60 & \hspace{.5cm}\bf{29.23} & \bf{0.38} & \hspace{.5cm}29.09 & 0.97 & \hspace{.5cm} 28.46 & 1.54 & \hspace{.5cm}28.58 & 0.81 \\
\hline
\end{tabular}
\end{table}

\section{Conclusions} \label{seq-conclusions}

In this work we have described PPESMOC, the first
method to address Bayesian optimization problems
with several objectives and constraints in which the
black-boxes can be evaluated in parallel. More precisely,
PPESMOC suggests, at each iteration, a batch of
points at which the objectives and constraints should
be evaluated simultaneously. We have compared the performance of PPESMOC
on several optimization problems, including synthetic, benchmark and real-world problems.
Furthermore, we have compared results with two simple base-lines.
Namely, a random exploration strategy and two methods derived
from the literature about sequential Bayesian optimization,
PS\_PESMOC and PS\_BMOO.  We have observed that PPESMOC performs
well in general, giving similar and sometimes better results than PS\_PESMOC and PS\_BMOO.
The main advantage is, however, that the PPESMOC acquisition function scales much better
with respect to the batch size. Unlike PPESMOC, the sequential strategies
require repeating an iterative process as many times as the batch size.
This process includes hallucinating observations, re-fitting the underlying GP models,
and optimizing a sequential acquisition function. This leads to
a prohibitive computational cost for large batch sizes.	
For small batch sizes, however, these simple strategies are cheap and they
give over-all good results that are often similar to those of PPESMOC. Therefore, in that
setting, one may consider using them to solve optimization problems with
several objectives and constraints in which the evaluations have to be done in parallel.

\section*{Acknowledgments}

Authors gratefully acknowledge the use of the facilities of Centro de Computacion Cientifica (CCC) at Universidad Autónoma de Madrid. The authors also acknowledge financial support from Spanish Plan Nacional I+D+i, PID2019-106827GB-I00 /AEI / 10.13039/501100011033.  

\bibliographystyle{apalike}  
\bibliography{main_article}

\begin{thebibliography}{}

\bibitem[Ariizumi et~al., 2014]{ariizumi2014expensive}
Ariizumi, R., Tesch, M., Choset, H., and Matsuno, F. (2014).
\newblock Expensive multiobjective optimization for robotics with consideration
  of heteroscedastic noise.
\newblock In {\em 2014 IEEE/RSJ International Conference on Intelligent Robots
  and Systems}, pages 2230--2235. IEEE.

\bibitem[Azimi et~al., 2012]{azimi2012hybrid}
Azimi, J., Jalali, A., and Fern, X. (2012).
\newblock Hybrid batch {B}ayesian optimization.
\newblock {\em arXiv preprint arXiv:1202.5597}.

\bibitem[Bergstra et~al., 2011]{bergstra2011algorithms}
Bergstra, J., Bardenet, R., Bengio, Y., and K{\'e}gl, B. (2011).
\newblock Algorithms for hyper-parameter optimization.
\newblock In {\em Advances in neural information processing systems}, pages
  2546--2554.

\bibitem[Brochu et~al., 2010]{brochu2010tutorial}
Brochu, E., Cora, V., and De~Freitas, N. (2010).
\newblock A tutorial on {B}ayesian optimization of expensive cost functions,
  with application to active user modeling and hierarchical reinforcement
  learning.
\newblock {\em arXiv preprint arXiv:1012.2599}.

\bibitem[Chafekar et~al., 2003]{chafekar2003constrained}
Chafekar, D., Xuan, J., and Rasheed, K. (2003).
\newblock Constrained multi-objective optimization using steady state genetic
  algorithms.
\newblock In {\em Genetic and Evolutionary Computation Conference}, pages
  813--824.

\bibitem[Daxberger and Low, 2017]{daxberger2017distributed}
Daxberger, E.~A. and Low, B. K.~H. (2017).
\newblock Distributed batch {G}aussian process optimization.
\newblock In {\em International Conference on Machine Learning-Volume 70},
  pages 951--960.

\bibitem[Desautels et~al., 2014]{desautels2014parallelizing}
Desautels, T., Krause, A., and Burdick, J.~W. (2014).
\newblock Parallelizing exploration-exploitation tradeoffs in {G}aussian
  process bandit optimization.
\newblock {\em Journal of Machine Learning Research}, 15:3873--3923.

\bibitem[Dua and Graff, 2017]{Dua2019}
Dua, D. and Graff, C. (2017).
\newblock {UCI} machine learning repository.

\bibitem[Feliot et~al., 2017]{feliot2017bayesian}
Feliot, P., Bect, J., and Vazquez, E. (2017).
\newblock A bayesian approach to constrained single-and multi-objective
  optimization.
\newblock {\em Journal of Global Optimization}, 67(1-2):97--133.

\bibitem[Feurer and Hutter, 2019]{feurer2019hyperparameter}
Feurer, M. and Hutter, F. (2019).
\newblock Hyperparameter optimization.
\newblock In {\em Automated Machine Learning}, pages 3--33. Springer, Cham.

\bibitem[Garrido-Merch{\'a}n and Hern{\'a}ndez-Lobato,
  2019]{garrido2019predictive}
Garrido-Merch{\'a}n, E. and Hern{\'a}ndez-Lobato, D. (2019).
\newblock Predictive entropy search for multi-objective {B}ayesian optimization
  with constraints.
\newblock {\em Neurocomputing}.

\bibitem[Gonz{\'a}lez et~al., 2016]{gonzalez2016batch}
Gonz{\'a}lez, J., Dai, Z., Hennig, P., and Lawrence, N. (2016).
\newblock Batch {B}ayesian optimization via local penalization.
\newblock In {\em Artificial intelligence and statistics}, pages 648--657.

\bibitem[Gupta et~al., 2018]{gupta2018exploiting}
Gupta, S., Shilton, A., Rana, S., and Venkatesh, S. (2018).
\newblock Exploiting strategy-space diversity for batch {B}ayesian
  optimization.
\newblock In {\em International Conference on Artificial Intelligence and
  Statistics}, pages 538--547.

\bibitem[Hennig and Schuler, 2012]{hennig2012entropy}
Hennig, P. and Schuler, C. (2012).
\newblock Entropy search for information-efficient global optimization.
\newblock {\em Journal of Machine Learning Research}, 13(Jun):1809--1837.

\bibitem[Hern{\'a}ndez-Lobato et~al., 2016]{hernandez2016predictive}
Hern{\'a}ndez-Lobato, D., Hernandez-Lobato, J.~M., Shah, A., and Adams, R.~P.
  (2016).
\newblock Predictive entropy search for multi-objective {B}ayesian
  optimization.
\newblock In {\em International Conference on Machine Learning}, pages
  1492--1501.

\bibitem[Hern{\'a}ndez-Lobato et~al., 2009]{hernandez2009statistical}
Hern{\'a}ndez-Lobato, D., Mart{\'\i}nez-Mu{\~n}oz, G., and Su{\'a}rez, A.
  (2009).
\newblock Statistical instance-based pruning in ensembles of independent
  classifiers.
\newblock {\em IEEE Transactions on Pattern Analysis and Machine Intelligence},
  31:364--369.

\bibitem[Hern{\'a}ndez-Lobato et~al., 2015]{hernandez2015predictive}
Hern{\'a}ndez-Lobato, J.~M., Gelbart, M.~A., Hoffman, M.~W., Adams, R.~P., and
  Ghahramani, Z. (2015).
\newblock Predictive entropy search for {B}ayesian optimization with unknown
  constraints.
\newblock In {\em International Conference on Machine Learning}, pages
  1699--1707.

\bibitem[Hern{\'a}ndez-Lobato et~al., 2014]{hernandez2014predictive}
Hern{\'a}ndez-Lobato, J.~M., Hoffman, M., and Ghahramani, Z. (2014).
\newblock Predictive entropy search for efficient global optimization of
  black-box functions.
\newblock In {\em Advances in neural information processing systems}, pages
  918--926.

\bibitem[Kathuria et~al., 2016]{kathuria2016batched}
Kathuria, T., Deshpande, A., and Kohli, P. (2016).
\newblock Batched {G}aussian process bandit optimization via determinantal
  point processes.
\newblock In {\em Advances in Neural Information Processing Systems}, pages
  4206--4214.

\bibitem[LeCun et~al., 2010]{lecun2010mnist}
LeCun, Y., Cortes, C., and Burges, C. (2010).
\newblock Mnist handwritten digit database. at\&t labs.

\bibitem[Lyu et~al., 2018]{lyu2018batch}
Lyu, W., Yang, F., Yan, C., Zhou, D., and Zeng, X. (2018).
\newblock Batch bayesian optimization via multi-objective acquisition ensemble
  for automated analog circuit design.
\newblock In {\em International conference on machine learning}, pages
  3306--3314. PMLR.

\bibitem[Maclaurin et~al., 2015]{maclaurin2015autograd}
Maclaurin, D., Duvenaud, D., and Adams, R.~P. (2015).
\newblock Autograd: Effortless gradients in numpy.
\newblock In {\em ICML 2015 AutoML Workshop}, volume 238, page~5.

\bibitem[Minka, 2001]{minka2001expectation}
Minka, T. (2001).
\newblock Expectation propagation for approximate {B}ayesian inference.
\newblock In {\em Proceedings of the Seventeenth conference on Uncertainty in
  artificial intelligence}, pages 362--369. Morgan Kaufmann Publishers Inc.

\bibitem[Mockus et~al., 1978]{mockus1978application}
Mockus, J., Tiesis, V., and Zilinskas, A. (1978).
\newblock The application of bayesian methods for seeking the extremum.
\newblock {\em Towards global optimization}, 2(117-129):2.

\bibitem[Rasmussen, 2003]{rasmussen2003gaussian}
Rasmussen, C. (2003).
\newblock {G}aussian processes in machine learning.
\newblock In {\em Summer School on Machine Learning}, pages 63--71. Springer.

\bibitem[Shah and Ghahramani, 2015]{shah2015parallel}
Shah, A. and Ghahramani, Z. (2015).
\newblock Parallel predictive entropy search for batch global optimization of
  expensive objective functions.
\newblock In {\em Advances in Neural Information Processing Systems}, pages
  3330--3338.

\bibitem[Shahriari et~al., 2015]{shahriari2015taking}
Shahriari, B., Swersky, K., Wang, Z., Adams, R., and De~Freitas, N. (2015).
\newblock Taking the human out of the loop: A review of {B}ayesian
  optimization.
\newblock {\em Proceedings of the IEEE}, 104(1):148--175.

\bibitem[Shao et~al., 2014]{shao2014aladdin}
Shao, Y.~S., Reagen, B., Wei, G., and Brooks, D. (2014).
\newblock Aladdin: A pre-rtl, power-performance accelerator simulator enabling
  large design space exploration of customized architectures.
\newblock In {\em International Symposium on Computer Architecture}, pages
  97--108.

\bibitem[Siarry and Collette, 2003]{siarry2003multiobjective}
Siarry, P. and Collette, Y. (2003).
\newblock Multiobjective optimization: principles and case studies.

\bibitem[Singh et~al., 2003]{singh2003nearest}
Singh, H., Misra, N., Hnizdo, V., Fedorowicz, A., and Demchuk, E. (2003).
\newblock Nearest neighbor estimates of entropy.
\newblock {\em American journal of mathematical and management sciences},
  23:301--321.

\bibitem[Snoek et~al., 2012]{snoek2012practical}
Snoek, J., Larochelle, H., and Adams, R. (2012).
\newblock Practical {B}ayesian optimization of machine learning algorithms.
\newblock In {\em Advances in neural information processing systems}, pages
  2951--2959.

\bibitem[Villemonteix et~al., 2009]{villemonteix2009informational}
Villemonteix, J., Vazquez, E., and Walter, E. (2009).
\newblock An informational approach to the global optimization of
  expensive-to-evaluate functions.
\newblock {\em Journal of Global Optimization}, 44(4):509.

\end{thebibliography}


\begin{thebibliography}{}

\bibitem[Gerven et~al., 2009]{gerven2009bayesian}
Gerven, M., Cseke, B., Oostenveld, R., and Heskes, T. (2009).
\newblock {B}ayesian source localization with the multivariate {L}aplace prior.
\newblock In {\em Advances in Neural Information Processing Systems}, pages
  1901--1909.

\bibitem[Maclaurin et~al., 2015]{maclaurin2015autograd}
Maclaurin, D., Duvenaud, D., and Adams, R.~P. (2015).
\newblock Autograd: Effortless gradients in numpy.
\newblock In {\em ICML 2015 AutoML Workshop}, volume 238, page~5.

\bibitem[Minka and Lafferty, 2012]{minka2012expectation}
Minka, T. and Lafferty, J. (2012).
\newblock Expectation-propogation for the generative aspect model.
\newblock {\em arXiv preprint arXiv:1301.0588}.

\bibitem[Opper and Archambeau, 2009]{opper2009variational}
Opper, M. and Archambeau, C. (2009).
\newblock The variational {G}aussian approximation revisited.
\newblock {\em Neural computation}, 21(3):786--792.

\end{thebibliography}

\end{document}


\title{Supplementary material for Parallel Predictive Entropy Search for Multi-objective Bayesian Optimization with Constraints}

\author{
Eduardo C. Garrido-Merchán \\
Universidad Aut\'onoma de Madrid\\
\texttt{eduardo.garrido@uam.es} \\
\and
Daniel Hern\'andez-Lobato\\
Universidad Aut\'onoma de Madrid\\
\texttt{daniel.hernandez@uam.es}
}

\date{}

\maketitle

\begin{abstract}
In this supplementary material, we give additional information about PPESMOC. In particular, we provide information about the EP algorithm used to compute the approximation to the PPESMOC acquisition function, the computation of the PPESMOC gradients to introduce them into the L-BFGS optimization algorithm in order to getter better results from the optimization of the acquisition function and results of additional experiments.
\end{abstract}

\section{Optimization of the PPESMOC acquisition function approximation}
The PPESMOC acquisition function $\alpha(\cdot)$ has $B$ times more dimensions than the sequential PESMOC acquisition function, where $B$ is the number of points considered in the batch. Due to the curse of dimensionality, obtaining an estimate of the maximum of the acquisition function is not longer feasible by the procedure done in the PESMOC acquisition function, based only on using a grid and a local search procedure, like the L-BFGS algorithm, approximating the gradients by differences. Hence, it is necessary to compute the exact gradients of the PPESMOC acquisition function w.r.t the inputs, $\mathbf{X}$.

Importantly, as explained in the main document, the parameters of the approximate factors can be considered to be fixed after EP has converged. This simplifies significantly the gradient computation. We compute the gradients of the PPESMOC acquisition function by using the Autograd tool \citep{maclaurin2015autograd}. To optimize PPESMOC, we choose an initial point of PPESMOC at random. Then, we run the L-BFGS algorithm with the initial point and the gradients of PPESMOC computed by Autograd. We only compute the factors of the observations and Pareto set point in the first iteration of the optimization as they are the same for all the optimization process. For every new point in the optimization done by L-BFGS we only refine once with EP the test factors. We set a maximum of $100$ iterations of the L-BFGS optimization procedure.

\section{The EP Approximation to the $\Phi(\cdot)$ and $\Omega(\cdot,\cdot)$ factors}

The EP algorithm updates each of the approximate factors presented in the previous section until convergence. The following sections will describe the necessary operations needed for the EP algorithm to update each of the factors. It the following subsection, it is assumed that we have already obtained the mean and variances of each of the $\mathcal{K}$ and $\mathcal{J}$ conditional predictive distributions, which will be explained in detail in section 4.3.

\subsection{EP update operations for the $\Phi(\cdot)$ factors}

As it was explained in section 3, for the $\mathcal{M}$ Pareto Set points defined by the set $\mathcal{X}^*$, in every point $\boldsymbol{x}_i \in \mathcal{X}^*$, the EP algorithm will generate $J$ approximations for the $\Phi(\boldsymbol{x}_j)$ factors for every constraint $c_j$ that will be defined by its mean $\hat{f}_{j}^{\boldsymbol{x}}$ and its variance $\hat{e}_{j}^{\boldsymbol{x}}$. Computations are done for all the points $\boldsymbol{x}_i \in \mathcal{X}^*$. The operations for these factors are described as follows.

\subsubsection{Computation of the Cavity Distribution}

The first step performed by the EP algorithm is the computation of the Cavity Distribution $\tilde{q}^{\setminus n}(\boldsymbol{x})$. In order to make the computations easier, we first obtain the natural parameters of the Gaussian Distributions for all the $\mathcal{M}$ Pareto Set points by using the equations:
\begin{align}
& \boldsymbol{\hat{m}}_{j} = \frac{\boldsymbol{\xi}_j}{diag(\boldsymbol{\Xi}_j)}\,, \nonumber \\
& \boldsymbol{\hat{v}}_{j} = \frac{1}{diag(\boldsymbol{\Xi}_j)}\,.
\end{align}
Where $\boldsymbol{\Xi}_j$ is a vector of the variances of the $\mathcal{M}$ points for the constraint $c_j$ and $\boldsymbol{\Xi}$ is the matrix of all the variances of all $\mathcal{M}$ and $\mathcal{N}$ points which construction will be explained in detail in section 4.3. The term $diag$ holds for the diagonal of $\boldsymbol{\Xi}$ as we are only interested in the variance of the $M$ points and not the variance of these points with the $N$ points for the factor $\Phi(\cdot)$. In the same way, $\boldsymbol{\xi}_j$, is the vector of means for the constraint $c_j$ and $\boldsymbol{\xi}$ contains all the means of all the points for every constraint in $\boldsymbol{c}$. $\boldsymbol{\hat{m}}_{j}$ and $\boldsymbol{\hat{v}}_{j}$ hold the mean and variance natural parameters corresponding for all the points in the set $\mathcal{X}^*$.

Once we have obtain the natural parameters $\boldsymbol{\hat{m}}_{j}$ and $\boldsymbol{\hat{v}}_{j}$, we obtain the cavity distribution. As we are dealing with natural parameters, it is not necessary to use the formula for the ratio of Gaussian Distributions, the cavity distribution defined by mean $\boldsymbol{\hat{m}}^{\setminus j}$ and variance $\boldsymbol{\hat{v}}^{\setminus j}$ will simply be obtained by the subtraction of the natural parameters between the approximated distribution defined by parameters $\boldsymbol{\hat{m}_{j}}$ and $\boldsymbol{\hat{v}_{j}}$ (which is equivalent to the product of all the factors for all the constraints) and the factor $\boldsymbol{\hat{e}}_{j}$ and $\boldsymbol{\hat{f}}_{j}$ corresponding to the constraint $c_j$ that we want to update:
\begin{align}
& \boldsymbol{\hat{v}}_{nat}^{\setminus j} = \boldsymbol{\hat{v}_{j}} - \boldsymbol{\hat{e}}_{j}\,, \nonumber \\
& \boldsymbol{\hat{m}}_{nat}^{\setminus j} = \boldsymbol{\hat{m}_{j}} - \boldsymbol{\hat{f}}_{j}\,.
\end{align}
Once the subtraction is done, we transform the natural parameters of the cavity distribution into Gaussian parameters again by using the formula that converts natural to Gaussian parameters.
\begin{align}
& \boldsymbol{\hat{v}}^{\setminus j} = \frac{1}{\boldsymbol{\hat{v}}_{nat}^{\setminus j}}\,, \nonumber \\
& \boldsymbol{\hat{m}}^{\setminus j} = \boldsymbol{\hat{m}}_{nat} \boldsymbol{\hat{v}}^{\setminus j}\,.
\end{align}
The variances $\boldsymbol{\hat{v}}^{\setminus j}$ need to be positive for the following operations.

\subsubsection{Computation of the Partial Derivatives of the Normalization Constant}

Once the cavities $\boldsymbol{\hat{v}}^{\setminus j}$ and $\boldsymbol{\hat{m}}^{\setminus j}$ have been computed, the EP need to compute the quantities required for the update of the factors $\boldsymbol{\hat{e}}_{j}$ and $\boldsymbol{\hat{f}}_{j}$ in order to minimize the KL divergence between $\Phi(\cdot)$ and the approximation distribution. These quantities are the first and second moments of the distribution that we want to approximate. These are given by the log of the partial derivatives of $Z_j$, the constant that normalizes the distribution that we want to approximate, in this case, $\hat{\Phi}(\cdot)$.
\begin{align}
Z_j = \int \hat{\Phi}(\boldsymbol{x}^*)\ dc_j\,.
\end{align}
As $\Phi(\boldsymbol{x}^*)$ is approximated by a Gaussian Distribution $\hat{\Phi}(\boldsymbol{x}^*)$ with mean $\boldsymbol{\hat{m}}^{\setminus j}$ and variance $\boldsymbol{\hat{v}}^{\setminus j}$, the normalization constant $Z_j$ can be computed in closed form and its given by the cumulative distribution function ,$\Phi(\cdot)$, of this Gaussian Distribution:
\begin{align}
Z_j = \Phi\Bigg(\frac{\boldsymbol{\hat{m}}^{\setminus j}}{\sqrt{\boldsymbol{\hat{v}}^{\setminus j}}}\Bigg)\,.
\end{align}
Let $\alpha =  \frac{\boldsymbol{\hat{m}}^{\setminus j}}{\sqrt{\boldsymbol{\hat{v}}^{\setminus j}}}$, then $\log(Z_j) = \log(\Phi(\alpha))$. For numerical robustness, if $a,b \in \mathbb{R}$, we apply the rule $ \frac{a}{b} = \exp{(\log(a)-\log(b))}$. Using these expressions, the log-derivatives are computed as follows:
\begin{align}
& \frac{\partial \log(Z_j)}{\partial \boldsymbol{\hat{m}}^{\setminus j}} = \frac{\exp\{\log(N(\alpha))-\log(Z_j)\}}{\sqrt{\boldsymbol{\hat{v}}^{\setminus j}}}\,, \nonumber \\
& \frac{\partial \log(Z_j)}{\partial \boldsymbol{\hat{v}}^{\setminus j}} = - \frac{\exp\{\log(N(\alpha))-\log(Z_j)\}\alpha}{2\boldsymbol{\hat{v}}^{\setminus j}}\,.
\end{align}
Where $N(\cdot)$ represent the Gaussian probability density function. These expressions are valid for computing the first and second moments, but they do not present numerical robustness in all experiments. Since the lack of robustness of $\frac{\partial \log(Z_j)}{\partial \boldsymbol{\hat{v}}^{\setminus j}}$, we use the formula given by the work by \citet{opper2009variational}, and use the second partial derivative $\frac{\partial^{2} \log(Z_j)}{\partial [\boldsymbol{\hat{m}}^{\setminus j}]^2}$ rather than $\frac{\partial \log(Z_j)}{\partial \boldsymbol{\hat{v}}^{\setminus j}}$. This derivative is given by the following expression:
\begin{align}
\frac{\partial^{2} \log(Z_j)}{\partial [\boldsymbol{\hat{m}}^{\setminus j}]^2} = - \exp\{\log(N(\alpha))-\log(Z_j)\}\frac{\alpha\exp\{\log(N(\alpha))-\log(Z_j)\}}{\boldsymbol{\hat{v}}^{\setminus j}}\,.
\end{align}
Given these derivatives, in the next section it will be explained how to obtain the individual approximate factors $\boldsymbol{\hat{e}}_{j}$ and $\boldsymbol{\hat{f}}_{j}$.

\subsection{EP update operations for the $\Omega(\cdot,\cdot)$ factors}

Recalling section 3, for the $\mathcal{M}$ Pareto Set points defined by the set $\mathcal{X}^*$ and the $\mathcal{N}$ input space observation points defined by the set $\hat{\mathcal{X}}$, for every pair of points $\boldsymbol{x}_i \in \hat{\mathcal{X}}$ and $\boldsymbol{x}_j \in \mathcal{X}^*$, the EP will generate $K$ two-dimensional Gaussian approximations for every objective $\boldsymbol{f}_k$ that will be defined for the pair observation and Pareto set point by factors defined by mean $\hat{\boldsymbol{b}}_{ij}$ and variance $\hat{\boldsymbol{A}}_{ij}$ and for the pair of Pareto set points by factors defined by mean $\hat{\boldsymbol{d}}_{ij}$ and variance $\hat{\boldsymbol{C}}_{ij}$. It will also define $J$ one-dimensional Gaussian approximations for every constraint $c_j$ that will be defined for the pair observation and Pareto set point by factors defined by mean $\hat{bc}_j$ and variance $\hat{ac}_j$ and for the pair of Pareto set points by factors defined by mean $\hat{dc}_j$ and variance $\hat{cc}_j$. Computations are done for all the pairs of points from the sets $\mathcal{X}^*$ and $\hat{\mathcal{X}}$. The necessary operations for computing these factors are described in the following sections.

\subsubsection{Computation of the Cavity Distribution}
For the factors $\hat{ac}_j$, $\hat{bc}_j$, $\hat{cc}_j$ and $\hat{dc}_j$ that approximate the $\mathcal{J}$ one-dimensional Gaussian approximations for every constraint $c_j$, the operations needed to extract the cavity distribution from the approximate distribution are the same ones as the ones described in Section 4.1.1. These operations are done for the observation points in $\hat{\mathcal{X}}$ for the factors $\hat{ac}_j$, $\hat{bc}_j$ and for the Pareto Set points in $\mathcal{X}^*$ for the factors $\hat{cc}_j$ and $\hat{dc}_j$. That is, obtaining the natural parameters of $\boldsymbol{\Xi}_j$ as in Eq. (16), subtracting the natural parameters of the factor that is approximated, Eq. (17), and obtaining the Gaussian parameters of the cavity that we define for a point $\boldsymbol{x}_i$, $m_{ij}^{\setminus b_j}$ and $v_{ij}^{\setminus a_j}$, as shown in Eq. (18).

Obtaining the cavity distribution for the factors $\hat{\boldsymbol{A}}_{ij}$, $\hat{\boldsymbol{b}}_{ij}$, $\hat{\boldsymbol{C}}_{ij}$ and $\hat{\boldsymbol{d}}_{ij}$ that approximate the $\mathcal{K}$ two-dimensional Gaussian approximations for every objective $\boldsymbol{f}_k$ follow different expressions as in this case the Gaussian Distributions are bivariate for every pair of points considered.

In the first case, for the case of approximating a distribution that consider a point $\boldsymbol{x}_i$ belonging to the observations set $\hat{\mathcal{X}}$ and a point $\boldsymbol{x}_j$ from the Pareto set $\mathcal{X}^*$, that is, the factors $\hat{\boldsymbol{A}}_{ij}$ and $\hat{\boldsymbol{b}}_{ij}$, it is necessary to obtain, for every objective $k$ and each of the pair of points mentioned, the natural parameters $\boldsymbol{m}_{ij(nat)}^{k}$ and ${\boldsymbol{V}_{ij}^{k}}^{-1}$ of the Gaussian Process that models each of the $\mathcal{K}$ objectives $f(\cdot)_j$.
These natural parameters are obtained by the following expressions:
\begin{align}
   & \boldsymbol{m}_{ij(nat)}^{k} = {\boldsymbol{V}_{ij}^{k}}^{-1}\boldsymbol{m}_{ij}^{k}\,, \nonumber \\
   & {\boldsymbol{V}_{ij}^{k}}^{-1} = (\boldsymbol{V}_{ij}^{k})^{-1}\,,
\end{align}
where $\boldsymbol{V}_{ij}^{k}$ is a 2x2 matrix that represent in the points $\boldsymbol{x}_i$ and $\boldsymbol{x}_j$ the variance of the Gaussian approximation of the objective $k$ and $\boldsymbol{m}_{ij}^{k}$ is a vector that represent in the points $\boldsymbol{x}_i$ and $\boldsymbol{x}_j$ the mean of the Gaussian approximation of the objective $k$.

As in the constraints case, we now extract the cavity distribution that we define by the natural parameters $\boldsymbol{m}_{ijk(nat)}^{\setminus b}$ and $\boldsymbol{V}_{ijk(nat)}^{\setminus A}$, by subtracting to the computed natural parameters $\boldsymbol{m}_{ij(nat)}^{k}$ and ${\boldsymbol{V}_{ij}^{k}}^{-1}$, computed in the previous step, the factors that we want to update $\boldsymbol{b}_{ij}^{k}$ and $\boldsymbol{A}_{ij}^{k}$. That is:
\begin{align}
    & \boldsymbol{m}_{ijk(nat)}^{\setminus b} = \boldsymbol{m}_{ij(nat)}^{k} - \boldsymbol{b}_{ij}^{k}\,, \nonumber \\
    & \boldsymbol{V}_{ijk(nat)}^{\setminus A} = {\boldsymbol{V}_{ij}^{k}}^{-1} - \boldsymbol{A}_{ij}^{k}\,.
\end{align}
For the bivariate Gaussian distribution, the step of obtaining the Gaussian parameters from the natural parameters is defined by the following expressions:
\begin{align}
    & \boldsymbol{m}_{ijk}^{\setminus b} = \boldsymbol{V}_{ijk}^{\setminus A}\ \boldsymbol{m}_{ijk(nat)}\,, \nonumber \\
    & \boldsymbol{V}_{ijk}^{\setminus A} = (\boldsymbol{V}_{ijk(nat)}^{\setminus A})^{-1}\,,
\end{align}
where $\boldsymbol{V}_{ijk}^{\setminus A}$ is a 2x2 matrix with the variances of each of the points and the correlation between each of them and $\boldsymbol{m}_{ijk}^{\setminus b}$ is a two position vector that represent the means. In the case of the factors $\hat{\boldsymbol{C}}_{ij}$ and $\hat{\boldsymbol{d}}_{ij}$ that consider two Pareto Set points, the operations for extracting the cavity distribution are the same ones as in the previous case.

\subsubsection{Computation of the Partial Derivatives of the Normalization Constant}

In this section, the operations needed to compute the partial derivatives for all the $\hat{\boldsymbol{A}}_{ij}$, $\hat{\boldsymbol{b}}_{ij}$, $\hat{\boldsymbol{C}}_{ij}$, $\hat{\boldsymbol{d}}_{ij}$, $\hat{ac}_j$, $\hat{bc}_j$, $\hat{cc}_j$ and $\hat{dc}_j$ are described. These derivatives need previous computations in order to compute the normalization constant $Z_\Omega$ of the factor $\Omega(\cdot,\cdot)$ that we want to approximate. These computations are given by the following expressions, all of which depend upon terms computed in the previous section. The shown computations are the result of applying rules in order to be robust such as $a/b = \exp\{\log(a)-\log(b)\}$ and $ab = \exp\{\log(a)+\log(b)\}$.These operations are equivalent for the two points cases, but here, the necessary operations for computing the normalization constant $Z_\Omega$ are described for the case of the factors $\hat{\boldsymbol{A}}_{ij}$, $\hat{\boldsymbol{b}}_{ij}$, $\hat{ac}_j$ and $\hat{bc}_j$:
\begin{align}
    & \boldsymbol{s}_k = \boldsymbol{V}_{ijk[0,0]}^{\setminus A} + \boldsymbol{V}_{ijk[1,1]}^{\setminus A} - 2\boldsymbol{V}_{ijk[0,1]}^{\setminus A}\,, \\
    & \boldsymbol{\alpha}_k = \frac{\boldsymbol{m}_{ijk[0]}^{\setminus b}-\boldsymbol{m}_{ijk[1]}^{\setminus b}}{\sqrt{s_k}}\,, \\
    & \boldsymbol{\beta}_j = \frac{m_{ij}^{\setminus b_j}}{\sqrt{v_{ij}^{\setminus a_j}}}\,, \\
    & \boldsymbol{\phi} = \Phi(\boldsymbol{\alpha})\,, \\
\end{align}
where $\Phi(\cdot)$ represents the c.d.f of a Gaussian distribution,
\begin{align}
    & \boldsymbol{\gamma} = \Phi(\boldsymbol{\beta})\,, \\
    & \zeta = 1-\exp\{\sum_{k=1}^{K} \log(\boldsymbol{\phi}_k)\}\,, \\
    & \log(\eta) = \sum_{j=1}^{J}\log(\boldsymbol{\gamma}_j) + \log(\zeta)\,, \\
    & \lambda = 1-\exp\{\sum_{j=1}^{J}\log(\boldsymbol{\gamma}_j)\}\,, \\
    & \tau = \max(\log(\eta),\log(\lambda))\,, \\
    & \log(Z_\Omega) = \log(\exp\{\log(\eta) - \tau\} + \exp\{\log(\lambda) - \tau\}) + \tau\,.
\end{align}
Having computed these terms, the log partial derivatives for the update of the factors that collaborate to the approximation of the objective variances $\hat{\boldsymbol{A}}_{ij}$ and the objective means $\hat{\boldsymbol{b}}_{ij}$ are given by the expressions:
\begin{align}
    &  \boldsymbol{\rho}_k = - \exp\{\log(\mathcal{N}(\boldsymbol{\alpha}_k))\} - \log(Z_\Omega) + \sum_{k=1}^{K}\{\log(\Phi(\boldsymbol{\alpha}_k))\} - \log(\Phi(\boldsymbol{\alpha}_k)) + \sum_{j=1}^{J}\{\log(\Phi(\boldsymbol{\beta}_j))\}\,, \\
    & \frac{\partial \log(Z_\Omega)}{\partial\boldsymbol{m}_{ijk}^{\setminus b}} = \frac{\boldsymbol{\rho}_k}{\sqrt{\boldsymbol{s}_k}}\big[1,-1\big] \,, \nonumber \\
    & \frac{\partial \log(Z_\Omega)}{\partial\boldsymbol{V}_{ijk}^{\setminus A}} = - \frac{\boldsymbol{\rho}_k\boldsymbol{\alpha}_k}{2\boldsymbol{s}_k}[[1,-1],[-1,\ 1]]\,.
\end{align}
Derivatives are computed for the two position vector mean and the 2x2 variance matrix, so they have the same structure, given by the [1, -1] and [[1, -1],[-1, 1]] expressions. The change in the sign appears due to the fact that the expression changes, whether it is the derivative of the mean of the observation point or the Pareto Set point or the derivative of the variance of one point or their correlation.

Alas, the derivative of the variance presents the same lack of robustness as in the constraint case shown in section 4.1.2. In order to ensure numerical robustness, we use the second partial derivative of the mean of the normalization constant instead of the first partial derivative of the variance for the further computation of the second moment. That is,
\begin{align}
    & \frac{\partial^2 \log(Z_\Omega)}{\partial\big[\boldsymbol{m}_{ijk}^{\setminus b}\big]^2} = - \frac{\boldsymbol{\rho}_k}{\boldsymbol{s}_k}(\boldsymbol{\alpha}_k+ \boldsymbol{\rho}_k) [[1,-1],[-1,\ 1]]\,.
\end{align}
For the log partial derivatives for the update of the factors that collaborate to the approximation of the constraint variances $\hat{ac}_j$ and the constraint means $\hat{bc}_j$, let $\boldsymbol{\omega}_j$ be defined as:
\begin{align}
\boldsymbol{\omega}_j = &  \exp\{\log(\mathcal{N}(\boldsymbol{\beta}_j))\} - \log(Z_\Omega) + \log(\zeta) + \sum_{j=1}^{J}(\log(\Phi(\boldsymbol{\beta}_j))) - \log(\Phi(\boldsymbol{\beta}_j)) - \exp\{\log(\mathcal{N}(\boldsymbol{\beta}_j))\}\,, \nonumber \\ & - 
\log(Z_\Omega) + \sum_{j=1}^{J}(\log(\Phi(\boldsymbol{\beta}_j))) - \log(\Phi(\boldsymbol{\beta}_j))\,.
\end{align}
Then, the robust log partial derivatives for the first and the second moments are given by the expressions:
\begin{align}
    & \frac{\partial \log(Z_\Omega)}{\partial m_{ij}^{\setminus \boldsymbol{b}_j}} = \frac{\boldsymbol{\omega}_j}{\sqrt{\boldsymbol{s}_j}}\,, \nonumber \\
    & \frac{\partial^2 \log(Z_\Omega)}{\partial [m_{ij}^{\setminus \boldsymbol{b}_j}]^2} = - \frac{\boldsymbol{\omega}_j}{\boldsymbol{s}_j}\ (\boldsymbol{\beta}_j + \boldsymbol{\omega}_j)\,.
\end{align}
The expressions for the log partial derivatives of $\hat{\boldsymbol{C}}_{ij}$, $\hat{\boldsymbol{d}}_{ij}$, $\hat{cc}_j$ and $\hat{dc}_j$ are similar to the presented expressions in this section, but taking into account pairs of points belonging to the set $\mathcal{X}^*$.

\subsubsection{Computation of the First and Second Moments for the Updates}
Giving the expressions computed in the previous section, the first and second moments of the different Gaussian Distributions that approximate the factor $\Omega(\cdot,\cdot)$ can now be computed.

The expressions for computing the factors $\hat{\boldsymbol{A}}_{ij}$, $\hat{\boldsymbol{b}}_{ij}$, $\hat{\boldsymbol{C}}_{ij}$, $\hat{\boldsymbol{d}}_{ij}$ for each of the $K$ objectives and the factors $\hat{ac}_j$, $\hat{bc}_j$, $\hat{cc}_j$ and $\hat{dc}_j$ for each of the $J$ constraints are the following ones:
\begin{align}
    & \hat{\boldsymbol{A}}_{ij}^{k} = \frac{\partial^2 \log(Z_\Omega)}{\partial\big[\boldsymbol{m}_{ijk}^{\setminus b}\big]^2}\ ((\boldsymbol{V}_{ijk}^{\setminus A}\frac{\partial^2 \log(Z_\Omega)}{\partial\big[\boldsymbol{m}_{ijk}^{\setminus b}\big]^2})^{-1}[[1, 0],[0, 1]])\,,\\
    & \hat{\boldsymbol{b}}_{ij}^{k} = ((\frac{\partial \log(Z_\Omega)}{\partial\boldsymbol{m}_{ijk}^{\setminus b}} - \boldsymbol{m}_{ijk}^{\setminus b})\frac{\partial^2 \log(Z_\Omega)}{\partial\big[\boldsymbol{m}_{ijk}^{\setminus b}\big]^2})\ ((\boldsymbol{V}_{ijk}^{\setminus A}\frac{\partial^2 \log(Z_\Omega)}{\partial\big[\boldsymbol{m}_{ijk}^{\setminus b}\big]^2})^{-1}+[[1, 0],[0, 1]])\,,\\
    & \hat{\boldsymbol{C}}_{ij}^{k} = \frac{\partial^2 \log(Z_\Omega)}{\partial\big[\boldsymbol{m}_{ijk}^{\setminus b}\big]^2}\ ((\boldsymbol{V}_{ijk}^{\setminus A}\frac{\partial^2 \log(Z_\Omega)}{\partial\big[\boldsymbol{m}_{ijk}^{\setminus b}\big]^2})^{-1}[[1, 0],[0, 1]])\,,\\
    & \hat{\boldsymbol{d}}_{ij}^{k} = ((\frac{\partial \log(Z_\Omega)}{\partial\boldsymbol{m}_{ijk}^{\setminus b}} - \boldsymbol{m}_{ijk}^{\setminus b})\frac{\partial^2 \log(Z_\Omega)}{\partial\big[\boldsymbol{m}_{ijk}^{\setminus b}\big]^2})\ ((\boldsymbol{V}_{ijk}^{\setminus A}\frac{\partial^2 \log(Z_\Omega)}{\partial\big[\boldsymbol{m}_{ijk}^{\setminus b}\big]^2})^{-1}+[[1, 0],[0, 1]])\,,
\end{align}
for the the rest of the factors, suppose that the index $h$ refers to the points of the Pareto Set $\mathcal{X}^*$:
\begin{align}
    & \hat{ac}_{h}^{j} = - \frac{\frac{\partial^2 \log(Z_\Omega)}{\partial [m_{ic}^{\setminus \boldsymbol{b}_j}]^2}}{1+\frac{\partial^2 \log(Z_\Omega)}{\partial [m_{ic}^{\setminus \boldsymbol{b}_j}]^2}v_{ic}^{\setminus a_j}}\,, \\
    & \hat{bc}_{h}^{j} = \frac{\frac{\partial \log(Z_\Omega)}{\partial m_{ic}^{\setminus \boldsymbol{b}_j}}-m_{ic}^{\setminus b_j}\frac{\partial^2 \log(Z_\Omega)}{\partial [m_{ic}^{\setminus \boldsymbol{b}_j}]^2}}{1+\frac{\partial^2 \log(Z_\Omega)}{\partial [m_{ic}^{\setminus \boldsymbol{b}_j}]^2}v_{ic}^{\setminus a_j}}\,, \\
    & \hat{cc}_{h}^{j} = - \frac{\frac{\partial^2 \log(Z_\Omega)}{\partial [m_{ic}^{\setminus \boldsymbol{b}_j}]^2}}{1+\frac{\partial^2 \log(Z_\Omega)}{\partial [m_{ic}^{\setminus \boldsymbol{b}_j}]^2}v_{ic}^{\setminus a_j}}\,, \\
    & \hat{dc}_{h}^{j} = \frac{\frac{\partial \log(Z_\Omega)}{\partial m_{ic}^{\setminus \boldsymbol{b}_j}}-m_{ic}^{\setminus b_j}\frac{\partial^2 \log(Z_\Omega)}{\partial [m_{ic}^{\setminus \boldsymbol{b}_j}]^2}}{1+\frac{\partial^2 \log(Z_\Omega)}{\partial [m_{ic}^{\setminus \boldsymbol{b}_j}]^2}v_{ic}^{\setminus a_j}}\,. 
\end{align}
All these factors are then used to rebuild the means and the variances of the Gaussian Processes that model the $K$ objectives and $C$ constraints of a constrained multi-objective optimization problem, as will be shown in the following section. That is, $C$ one-dimensional Gaussian Distributions for the constraint models and $C$ one-dimensional Gaussian Distributions and $K$ two-dimensional Gaussian Distributions for the objective models in each of the points in $\mathcal{X} = \{\mathcal{X}^* \cup \hat{\mathcal{X}} \cup \boldsymbol{x}\}$.

\subsection{Reconstruction of the Conditional Predictive Distribution}
In this section, we illustrate the way of obtaining a Conditional Predictive Distribution for every objective $f_k$ and every constraint $c_j$, given a sampled Pareto Set $\mathcal{X}^* = \{\boldsymbol{x}_1^*,...,\boldsymbol{x}_M^*\}$ of size $M$ and a set of $N$ input locations $\hat{\mathcal{X}} = \{\boldsymbol{x}_1,...,\boldsymbol{x}_N\}$ with corresponding observations of the $k$-th objective $\boldsymbol{y}_k$ and of the $j$-th constraint $\boldsymbol{y}_j$. For the following, it is assumed that we are given the EP approximate factors $\Phi(\cdot)$ and $\Omega(\cdot,\cdot)$, as an input for the next operations, which computation is explained in the previous section.

Recalling Eqs. 7, 8 and 9 of section 2, we want to obtain the $J$ Conditional Predictive Distributions in the products of constraints and the $K$ Conditional Predictive Distributions of the Gaussian Processes that model the objectives. The products presented in these factors are not a problem, due to the fact that the Gaussian Distributions are closed under the product operation, that is, the product of Gaussian Distributions is another Gaussian Distribution. These Conditional Predictive Distributions of the objectives and constraints are then used in Eq.(11) to build the final approximation.

Following the notation of section 4.1.1, let $\boldsymbol{\xi}_j$ and $\boldsymbol{\Xi}_j$ be the mean vector and variance matrix of the one-dimensional Gaussian Distributions of the $M+N$ points that generate the Gaussian Processes that model the constraints and let $\boldsymbol{m}_{k}$ and $\boldsymbol{V}_{k}$ be the mean vector and variance matrix of the two-dimensional Gaussian Distributions of the $M+N$ points that generate the Gaussian Processes that model the objectives. In order to update the constraint and objective distribution marginals, it is necessary to first follow the operations given by the equations 14 and 22, to obtain the natural parameters from the means and variances. Intuitively, as they are all natural parameters, these will be just sums taking into account that the matrices are formed first by the Pareto Set Points ,$M$, and then by the observations $N$. Univariate factors are added to the diagonal of these matrices, as they are not correlated with other points. Once the natural parameters are computed, the new means $\boldsymbol{\xi}_j$, $\boldsymbol{m}_{k}$ and variances $\boldsymbol{\Xi}_j$, $\boldsymbol{V}_{k}$ marginals are updated from the EP factors $\hat{\boldsymbol{A}}_{ij}$, $\hat{\boldsymbol{b}}_{ij}$, $\hat{\boldsymbol{C}}_{ij}$, $\hat{\boldsymbol{d}}_{ij}$, $\hat{e}_{j}$, $\hat{f}_{j}$, $\hat{ac}_j$, $\hat{bc}_j$, $\hat{cc}_j$ and $\hat{dc}_j$ by the following expressions:
\begin{align}
& \boldsymbol{\Xi}_{ii}^{j} = \boldsymbol{\Xi}_{ii(old)}^{j} + \sum_{m=1}^{M}\hat{cc}_{mi}^{j} + \hat{e}_{i}^{j} \ \ \ \ \ \ \ \ \ \ \ \ \ \ \ \ \ \ \ \ \ \ \ \ \ \ \ \ \ \ \ \ \ \ \ \ \ \ \ \ \textrm{for}\ \ i = 1,...,M\,, \nonumber \\
& \boldsymbol{\Xi}_{ii}^{c} = \boldsymbol{\Xi}_{ii(old)}^{j} + \sum_{m=1}^{M}\hat{ac}_{mi}^{j} \ \ \ \ \ \ \ \ \ \ \ \ \ \ \ \ \ \ \ \ \ \ \ \ \ \ \ \ \ \ \ \ \ \ \ \ \ \ \ \ \ \ \ \ \ \ \textrm{for}\ \ i = M+1,...,N+M\,, \nonumber \\
& \boldsymbol{\xi}_{i}^{c} = \ \boldsymbol{\xi}_{i(old)}^{j} + \sum_{m=1}^{M}\hat{dc}_{mi}^{j} + \hat{f}_{i}^{j} \ \ \ \ \ \ \ \ \ \ \ \ \ \ \ \ \ \ \ \ \ \ \ \ \ \ \ \ \ \ \ \ \ \ \ \ \ \ \ \ \ \textrm{for}\ \ i = 1,...,M\,, \nonumber \\
& \boldsymbol{\xi}_{i}^{c} = \ \boldsymbol{\xi}_{i(old)}^{j} +  \sum_{m=1}^{M}\hat{bc}_{mi}^{j} \ \ \ \ \ \ \ \ \ \ \ \ \ \ \ \ \ \  \ \ \ \ \ \ \ \ \ \ \ \ \ \ \ \ \ \ \ \ \ \ \ \ \ \ \ \ \ \  \textrm{for}\ \ i = M+1,...,N+M\,, \nonumber \\
& \boldsymbol{V}_{ii}^{k} = \boldsymbol{V}_{ii(old)}^{k} +  \sum_{j=M+1}^{N}\hat{\boldsymbol{A}}_{ji[1,1]}^{k} + \sum_{j=1}^{M}\hat{\boldsymbol{C}}_{ij[0,0]}^{k} + \sum_{j=1}^{M}\hat{\boldsymbol{C}}_{ji[1,1]}^{k} \ \ \  \textrm{for}\ \ i = 1,...,M\,, \nonumber \\
& \boldsymbol{V}_{ii}^{k} = \boldsymbol{V}_{ii(old)}^{k} +   \sum_{j=1}^{M}\hat{\boldsymbol{A}}_{ij[0,0]}^{k} \ \ \ \ \ \ \ \ \ \ \ \ \ \ \ \ \ \ \ \ \ \ \ \ \ \ \ \ \ \ \ \ \ \ \ \ \ \ \ \ \ \ \ \textrm{for}\ \ i = M+1,...,N+M\,, \nonumber \\
& \boldsymbol{V}_{ij}^{k} = \boldsymbol{V}_{ij(old)}^{k} + \boldsymbol{C}_{ij[0,1]}^{k} + {\boldsymbol{C}_{ij[1,0]}^{k}}^T \ \ \ \ \ \ \ \ \ \ \textrm{for}\ \ i = 1,...,M,\ \textrm{and}\ \textrm{for}\ \ j = 1,...,M\,, \nonumber \\
& \boldsymbol{V}_{ij}^{k} = \boldsymbol{V}_{ij(old)}^{k} + \boldsymbol{A}_{ij[0,1]}^{k} \ \ \ \ \ \ \ \ \ \ \ \  \ \ \ \ \ \ \ \ \ \ \ \ \ \textrm{for}\ \ i = M+1,...,N,\ \textrm{and}\ \textrm{for}\ \ j = 1,...,M\,, \nonumber \\
& \boldsymbol{V}_{ij}^{k} = \boldsymbol{V}_{ij(old)}^{k} + {\boldsymbol{A}_{ij[0,1]}^{k}}^T \ \ \ \ \ \ \ \ \ \ \ \ \ \ \ \ \ \ \ \ \ \ \ \textrm{for}\ \ i = 1,...,M,\ \textrm{and}\ \textrm{for}\ \ j = M+1,...,N\,, \nonumber \\
& \boldsymbol{m}_{i}^{k} = \boldsymbol{m}_{i(old)}^{k} +  \sum_{j=M+1}^{N+M}\hat{\boldsymbol{b}}_{ji[1]}^{k} + \sum_{j=1}^{M}\hat{\boldsymbol{d}}_{ij[0]}^{k} + \sum_{j=1}^{M}\hat{\boldsymbol{d}}_{ji[1]}^{k} \ \ \ \ \ \ \ \ \ \ \ \ \ \textrm{for}\ \ i = 1,...,M\,, \nonumber \\
& \boldsymbol{m}_{i}^{k} = \boldsymbol{m}_{i(old)}^{k} + \sum_{j=1}^{M}\hat{\boldsymbol{b}}_{ij[0]}^{k} \ \ \ \ \ \ \ \ \ \ \ \ \ \ \ \ \ \ \ \ \ \ \ \ \ \ \ \ \ \ \ \ \ \ \ \ \ \ \ \ \ \ \ \ \ \ \ \textrm{for}\ \ i = M+1,...,N+M\,. \nonumber \\
\end{align}
These natural parameters are then converted into Gaussian ones using the equations and 16 and 24. Once these operations are done the Gaussian Processes that model the objectives and constraints are updated from a full EP iteration.

\subsection{The Conditional Predictive Distribution at a New Batch}

After running EP until convergence one simply has to compute the covariance matrix of the posterior distribution for the process values of each objective and constraint at the candidate points $\mathbf{X}$. This implies computing the covariance matrix that results from the EP approximation to (\ref{eq:target}).  For this, one only has to replace the non-Gaussian factors with the corresponding approximation. The covariance matrices that are needed can be obtained using the fact that the Gaussian family is closed under the product operation. See Eq. (\ref{eq:ep_approx}).

\subsection{Initialization and convergence of EP}

When the EP algorithm computes the $\Phi(\cdot)$ and $\Omega(\cdot,\cdot)$ factors, it requires to set an initial value to all the factors that generates the Gaussians that approximate the $\Phi(\cdot)$ and $\Omega(\cdot,\cdot)$ factors. These factors, $\hat{\boldsymbol{A}}_{ij}$, $\hat{\boldsymbol{b}}_{ij}$, $\hat{\boldsymbol{C}}_{ij}$, $\hat{\boldsymbol{d}}_{ij}$, $\hat{e}_{j}$, $\hat{f}_{j}$, $\hat{ac}_j$, $\hat{bc}_j$, $\hat{cc}_j$ and $\hat{dc}_j$ are all set to be zero. The convergence criterion for stopping the EP algorithm updating the parameters is that the absolute change in all the cited parameters should be below $10^{-4}$. Other criteria may be used.

\subsection{Parallel EP Updates and Damping}

The updates of every approximate factor $\hat{\boldsymbol{A}}_{ij}$, $\hat{\boldsymbol{b}}_{ij}$, $\hat{\boldsymbol{C}}_{ij}$, $\hat{\boldsymbol{d}}_{ij}$, $\hat{e}_{j}$, $\hat{f}_{j}$, $\hat{ac}_j$, $\hat{bc}_j$, $\hat{cc}_j$ and $\hat{dc}_j$ are executed in parallel as it is described in the work by Gerven \cite{gerven2009bayesian}. The cavity distribution for each of the factors is computed and then the factors are updated afterwards. Once these operations are done the EP approximation is recomputed as it is described in the section 4.3.

In order to improve the convergence behavior of EP we use the damping technique described in Minka \& Lafferty \cite{minka2012expectation}. We use this technique for all the approximate factors. Damping simply reduces the quantity that the factor changes in every update as a linear combination between the old parameters and the new parameters. That is, if we define the old parameters of the factor to be updated as $u_{old}$, the new parameters as $u_{new}$ and the updated factor as $u$, then the update expression is:
\begin{align}
    u = \theta u_{new} + (1-\theta)u_{old}\,.
\end{align}
Where $\theta$ is the damping factor whose initial value is set to be 0.5, this factor controls the amount of damping, if this value is set to be one then no damping is employed. This factor is multiplied by 0.99 at each iteration, reducing the amount of change in the approximate factors in every iteration of the Bayesian Optimization. An issue that happens during the optimization process is that some covariance matrices become non positive definite due to a high large step size, that is, a high value of $\theta$. If this happens in any iteration, an inner loop executes again the update operation with $\theta_{new} = \theta_{old}\ /\ 2$ and the iteration is repeated. This inner loop is performed until the covariance matrices become non positive definite.

\section{Additional experiments information}
In this section, we include additional information about the experiments described on the main manuscript and their results.

\subsection{Benchmark Experiments}
We include tables with the analytical expressions of the benchmark of functions used for the benchmark experiments.

\begin{table}[H]
\centering
\caption{Summary of BNH, SRN, TNK and OSY problems used in the benchmark experiments.}
{\small
\begin{tabular}{| c | c | c |}
 \hline
 \multicolumn{3}{|c|}{Benchmark Experiments} \\
 \hline
 Problem Name & Input Space & Objectives $f_k(\mathbf{x})$ and Constraints $c_j(\mathbf{x})$ \\
 \hline
 \multirow{4}{4em}{BNH} & \multirow{4}{7em}{$x_1 \in [0,5]$\\$x_2 \in [0,3]$} & $f_1(\mathbf{x}) = 4x_1^2 + 4x_2^2$ \\
 & & $f_2(\mathbf{x}) = (x_1-5)^2 + (x_2-5)^2$ \\
 & & $c_1(\mathbf{x}) \equiv (x_1-5)^2 + x_2^2 \leq 25$ \\
 & & $c_2(\mathbf{x}) \equiv (x_1-8)^2 + (x_2+3)^2 \geq 7.7$ \\
 \hline
 \multirow{4}{4em}{SRN} & \multirow{4}{7em}{$x_1 \in [-20,20]$\\$x_2 \in [-20,20]$} & $f_1(\mathbf{x}) = 2+(x_1-2)^2+(x_2-2)^2$ \\
 & & $f_2(\mathbf{x}) = 9x_1 - (x_2-1)^2$ \\
 & & $c_1(\mathbf{x}) \equiv x_1^2 + x_2^2 \leq 225$ \\
 & & $c_2(\mathbf{x}) \equiv x_1 - 3x_2 + 10 \leq 0$ \\
 \hline
 \multirow{4}{4em}{TNK} & \multirow{4}{7em}{$x_1 \in [0,\pi]$\\$x_2 \in [0,\pi]$} & $f_1(\mathbf{x}) = x_1$ \\
 & & $f_2(\mathbf{x}) = x_2$ \\
 & & $c_1(\mathbf{x}) \equiv x_1^2 + x_2^2 - 1 - 0.1 cos(16arctan\frac{x_1}{x_2}) \geq 0$ \\
 & & $c_2(\mathbf{x}) \equiv (x_1-0.5)^2 + (x_2-0.5)^2 \leq 0.5$ \\
 \hline
\multirow{9}{4em}{OSY} & \multirow{9}{7em}{$x_1 \in [0,10]$\\$x_2 \in [0,10]$\\$x_3 \in [1,5]$\\$x_4 \in [0,6]$\\$x_5 \in [1,5]$\\$x_6 \in [0,10]$} &
$f_1(\mathbf{x}) = -[25(x_1-2)^2+(x_2-2)^2+(x_3-1)^2$ \\
 & &  $+(x_4-4)^2+(x_5-1)^2$ \\
 & & $f_2(\mathbf{x}) = x_1^2+x_2^2+x_3^2+x_4^2+x_5^2+x_6^2$ \\
 & & $c_1(\mathbf{x}) \equiv x_1+x_2-2 \geq 0$ \\
 & & $c_2(\mathbf{x}) \equiv 6-x_1-x_2 \geq 0$ \\
 & & $c_3(\mathbf{x}) \equiv 2-x_2+x_1 \geq 0$ \\
 & & $c_4(\mathbf{x}) \equiv 2-x_1+3x_2 \geq 0$ \\
 & & $c_5(\mathbf{x}) \equiv 4-(x_3-3)^2-x_4 \geq 0$ \\
 & & $c_6(\mathbf{x}) \equiv (x_5-3)^2+x_6-4 \geq 0$ \\
 \hline
\end{tabular}
}
\label{table:1}
\end{table}

\begin{table}[H]
\centering
\caption{Summary of CONSTR and Two-bar Truss problems used in the benchmark experiments.}
{\small
\begin{tabular}{| c | c | c |}
 \hline
 \multicolumn{3}{|c|}{Benchmark Experiments} \\
 \hline
 Problem Name & Input Space & Objectives $f_k(\mathbf{x})$ and Constraints $c_j(\mathbf{x})$ \\
 \hline
 \multirow{4}{4em}{CONSTR} & \multirow{4}{7em}{$x_1 \in [0.1,10]$\\$x_2 \in [0,5]$} & $f_1(\mathbf{x}) = x_1$ \\
 & & $f_2(\mathbf{x}) = \frac{(1+x_2)}{x_1}$ \\
 & & $c_1(\mathbf{x}) \equiv x_2+9x_1 \geq 6$ \\
 & & $c_2(\mathbf{x}) \equiv -x_2+9x_1 \geq 1$ \\
 \hline
 \multirow{3}{4em}{Two-bar\\Truss\\Design} & \multirow{3}{7em}{$x_1 \in [0,0.01]$\\$x_2 \in [0,0.01]$\\$x_3 \in [1,3]$} &
$f_1(\mathbf{x}) = x_1\sqrt{16+x_3^2} + x_2\sqrt{1+x_3^2}$ \\
 & & $f_2(\mathbf{x}) = \max(\frac{20\sqrt{16+x_3}}{x_1x_3},\frac{80\sqrt{1+x_3^2}}{x_2x_3})$ \\
 & & $c_1(\mathbf{x}) \equiv \max(\frac{20\sqrt{16+x_3}}{x_1x_3},\frac{80\sqrt{1+x_3^2}}{x_2x_3}) \leq 10^5$ \\
 \hline
\end{tabular}
}
\label{table:1_b}
\end{table}

\subsection{Real Experiments}

A summary of the parameters considered in the experiment of the hyper-parameter tuning of the deep neural network, their potential values, and their impact in each
black-box function (prediction error, time and chip area) is displayed in Table \ref{table:aladdin}.

\begin{table}[htb]
\centering
\caption{Parameter space of the deep neural network experiments. PE = Prediction error. T = Time. CA = Chip area.}
\begin{tabular}{lcccc}
 \hline
\textbf{Parameter} & \textbf{Min} & \textbf{Max} & \textbf{Step} & \textbf{Black-box} \\
 \hline
Hidden Layers & 1& 3& 1& PE/T/CA\\
Neurons per Layer & 5& 300& 1& PE/T/CA\\
Learning rate & $e^{-20}$& 1& $\epsilon$ & PE\\
Dropout rate & 0& 0.9& $\epsilon$ & PE\\
$\ell_1$ penalty & $e^{-20}$& 1& $\epsilon$ & PE\\
$\ell_2$ penalty & $e^{-20}$& 1& $\epsilon$ & PE\\
\hline
Memory partition & 1& 32& $2^{x}$& CA\\
Loop unrolling & 1& 32& $2^{x}$& CA\\
\hline
\end{tabular}
\label{table:aladdin}
\end{table}

\bibliographystyle{apalike}
\bibliography{supplemnetary_material}